\documentclass[final]{clv2025}

\jvol{vv}
\jnum{nn}
\jyear{2025}

\dochead{Long Paper} 

\pageonefooter{Action editor: \{action editor name\}. Submission received: DD Month YYYY; revised version received: DD Month YYYY; accepted for publication: DD Month YYYY.}
\usepackage{tikz}
\usetikzlibrary{shapes, backgrounds}
\usepackage{amsmath}
\usepackage{booktabs, makecell}
\usepackage{paralist}
\usepackage{longtable}
\usepackage{array}    
\usepackage{graphicx}
\usepackage{todo}
\usepackage{makecell}
\usepackage{pifont}
\usepackage{longtable} 
\usetikzlibrary{positioning, shapes, fit}
\usepackage{tabularx} 
\usetikzlibrary{matrix}
\usepackage{multirow}
\usepackage{xspace}
\usepackage{soul}
\usepackage{xcolor}
\sethlcolor{yellow}
\newcommand{\F}{$\text{F}_1$\xspace}
\newcommand{\roth}[1]{\rotatebox{90}{\strut #1}}

\newcommand{\unfilleddiamond}{
	\tikz[baseline=-0.5ex, line width=0.5pt, inner sep=-1pt, outer sep=0pt, scale=2]
	\draw (0,0) -- (0.2em,0.2em) -- (0.4em,0) -- (0.2em,-0.2em) -- cycle;
}
\newcommand{\corpusname}{\textsc{DeFaBel}\xspace}
\newcommand{\decop}{\textsc{Dec\_Op}\xspace}
\newcommand{\intn}{\textsc{Intention}\xspace}
\newcommand{\crosscult}{\textsc{Cross\_Cult\_De}\xspace}
\newcommand{\opendomain}{\textsc{Open\_Domain}\xspace}
\newcommand{\defabeleng}{\textsc{DeFaBel\_V2\_En}\xspace}
\newcommand{\defabelone}{\textsc{DeFaBel\_V1\_De}\xspace}
\newcommand{\defabeltwo}{\textsc{DeFaBel\_V2\_De}\xspace}
\newcommand{\mistral}{Mistral-7B }
\newcommand{\llama}{LLaMA-8B }
\newcommand{\qwen}{Qwen-14B }
\newcommand{\gptmini}{GPT-4o-mini }

\renewcommand{\arraystretch}{0.85} 
\usepackage{booktabs}

\usepackage{subcaption} 
\usepackage{adjustbox} 
\renewcommand{\arraystretch}{1.2} 
  \usepackage{array}    

\usepackage{annotates}
\providecommentcommand{rom}{blue}{Roman}
\providecommentcommand{asw}{violet}{Aswathy}
\runningtitle{A Cross-Linguistic Study on the Limits of Deception Detection}
\runningauthor{Velutharambath}

\begin{document}

\title{What if Deception Cannot be Detected? \\
A Cross-Linguistic Study on the Limits of Deception Detection from Text} 

\author{Aswathy Velutharambath$^{1,2}$, Kai Sassenberg$^{3}$, Roman Klinger$^{2}$ \thanks{Primary correspondence: Aswathy Velutharambath}}

\affilblock{
    \affil{Institut f\"ur Maschinelle Sprachverarbeitung, University of Stuttgart, Germany\\\quad \email{aswathy.velutharambath@ims.uni-stuttgart.de}}
    \affil{Fundamentals of Natural Language Processing, University of Bamberg, Germany \\\quad \email{roman.klinger@uni-bamberg.de}}
        \affil{Leibniz Institute for Psychology, Trier, Germany \\\quad \email{ksa@leibniz-psychology.de}}
}

\maketitle
\begin{abstract}
Can deception be detected solely from written text? Cues of deceptive communication are inherently subtle, and even more so in text-only communication. Yet, prior studies have reported considerable success in automatic deception detection. In this study, we hypothesize that such findings are largely driven by artifacts introduced during data collection and do not generalize beyond specific datasets.  We revisit this foundational assumption by introducing a belief-based deception framework, which defines deception as a misalignment between an author's claims and their true beliefs, irrespective of factual  accuracy, allowing deception cues to be studied in isolation.
Based on this framework, we construct three corpora—collectively referred to as \corpusname—including a German-language corpus of deceptive and non-deceptive arguments and a multilingual version in German and English, each collected under different conditions to account for potential belief change and enable cross-linguistic analysis. Using these corpora, we evaluate commonly reported linguistic cues of deception. Across all three \corpusname\ variants, we find that these cues exhibit negligible and statistically insignificant correlations with deception labels, contrary to prior work that treats such cues as reliable indicators. We further benchmark against other English deception datasets, that follow similar data collection protocols. While some of these show statistically significant correlations, the effect sizes remain low and, critically, the set of predictive cues is inconsistent across datasets. We also evaluate deception detection  using feature-based models, pre-trained language models, and instruction-tuned large language models. While some of these models perform well on established deception datasets, they consistently perform near chance on \corpusname. Our findings challenge the core assumption that deception can be reliably inferred from linguistic cues and call for a rethinking how deception is studied and modeled in Natural Language Processing.

\end{abstract}

\section{Introduction}

Deception is not fundamentally about factual inaccuracy, but about the intention of the author to deceive \cite{mahon2016}. A statement can be objectively true and still be deceptive if the author does not believe it. Conversely, a factually inaccurate statement may be asserted with full sincerity and thus not constitute a lie. For instance, consider the statement: \textit{``The Colosseum in Rome is an iconic ancient amphitheater. My family owns a secret underground passage that leads directly to it.''} While the first sentence is factually accurate---i.e., verifiable based on objective evidence---the second is likely fabricated. As a whole, the text is deceptive given the author's intent, even though it contains elements that are objectively true. This disconnect between truth and intent presents a fundamental challenge to deception detection in Natural Language Processing (NLP): factual correctness and deceptive intent are \textit{mutually independent}, and systems that conflate the two risk mischaracterizing both. In this paper, we use the terms \textit{truthful} or \textit{non-deceptive} (interchangeably) to mean \textit{honest and not containing or telling any lies}\footnote{As defined by the  \href{https://dictionary.cambridge.org/us/dictionary/english/truthful}{Cambridge
		English Dictionary}} while \textit{factually accurate} refers to those \textit{ verifiable against objective evidence}.

The vast majority of NLP research on deception has focused on detecting linguistic markers of deceptive intent, without considering the factual accuracy of the statements or the speaker's underlying beliefs \citep{ott-etal-2011-finding,  perez-rosas-mihalcea-2014-cross, salvetti-etal-2016-tangled, capuozzo-etal-2020-decop}. These approaches typically assume that deception manifests in observable language patterns, overlooking the epistemic dimension of deception, namely, the misalignment between a speaker's beliefs and their claims. This issue becomes even more pertinent with the advent of Large Language Models (LLMs) trained on extensive corpora of world knowledge, which could be biased towards factual accuracy. However, despite their vast informational foundations, these models exhibit mixed performance in detecting factual inconsistencies and may inadvertently align with prevalent misconceptions \citep{lin-etal-2022-truthfulqa, augenstein2024factuality}. This makes it even more ill-equipped in detecting \textit{genuine} deception, especially when the deception takes the form of an objectively correct statement written with deceptive intent.

To disentangle deception from factuality, we introduce a belief-based framework, where a statement is considered deceptive when it misrepresents the author's \textit{belief}, not when it misrepresents the world. For example, the sentence \emph{``The Earth revolves around the Sun''} is scientifically accurate. Yet if someone who believes otherwise writes it---to appear informed or persuasive---the statement becomes deceptive, despite its factual correctness. Here, the deception lies not in the content, but in the divergence between private belief and public expression. This distinction is foundational to our approach.

Using this framework, we construct three corpora in which deceptive intent is operationalized as a misalignment between participants' self-reported beliefs and their written arguments. \defabelone\footnote{Both the belief-based deception framework and the \defabelone corpus, were first introduced by \citet{velutharambath-etal-2024-factual}. Their inclusion here aims to situate the present study within a coherent and unified context.} is a German-language corpus in which participants are asked to argue in favor of a given statement, followed by a post-task report of their true belief. In \defabeltwo, we address the possibility that participants may shift their beliefs during argumentation, which could undermine the reliability of deception labels. To account for this, participants report their belief both before and after writing. Finally, \defabeleng extends this design to English, enabling cross-linguistic comparison of deception phenomena within the same belief-based framework.

We conduct a comparative analysis of linguistic cues across \corpusname datasets and other deception datasets that adopt comparable data collection methods. This analysis examines whether markers of deceptive intent remain consistent across languages and data collection settings. Further, to assess the robustness of current deception detection methods, we evaluate a diverse set of models, including feature-based classifiers, pre-trained language models, and instruction-tuned large language models, on these datasets. 

Our experiments show that, while many of these models perform well on other datasets, their performance drops to near-chance levels on all versions of \corpusname. This systematic failure suggests that prior model success is largely driven by linguistic patterns that are likely dataset-specific artifacts, rather than by features that reliably reflect deceptive intent. Supporting this conclusion, we find that linguistic indicators commonly associated with deception do not generalize to belief-based deception data: they fail to correlate significantly with deception labels when lies are embedded within factually accurate contexts. As hypothesized, in the case of LLMs, we observe a pronounced bias wherein models tend to classify most instances as truthful (or non-deceptive). Notably, this bias persists even as model size increases, indicating that scaling alone does not mitigate the issue. Our experiments with different prompting strategies show that LLM performance across these strategies is inconsistent, and that CoT prompting introduces a deception bias instead of a truth bias. To better understand the underlying mechanisms driving these misclassifications, we perform an error analysis of the models' self-reported rationales. Our analysis reveals a consistent pattern: models rely on cues such as factual correctness and the presence of subjective expressions (e.g., personal opinions) as proxies for truthfulness.

These findings do not reflect a failure of modeling, but a limitation in how the task is formulated. When deception is disentangled from factual inaccuracy and expressed in structured argumentative settings, it does not produce consistent linguistic markers. This challenges the assumption that deception reliably manifests in text and points to the limitations of treating it as a purely linguistic phenomenon. \textit{This calls for a fundamental rethinking of how deceptive intent is conceptualized, detected, and modeled within NLP.}

\paragraph{Contributions}

\begin{itemize}
	\item We present a novel bilingual deception corpus (\corpusname) grounded in a belief-based deception annotation framework, which effectively disentangles factual accuracy and deceptive intent. The \corpusname corpus has been released publicly\footnote{Data and code: \url{https://www.ims.uni-stuttgart.de/data/defabel}}.

	\item We show that linguistic cues are inconsistent, weak, or absent in belief-based deception data, and that prior success in deception detection may be driven by dataset artifacts rather than genuine deceptive signals.
	
	\item We provide a cross-dataset and cross-lingual evaluation of deception detection performance using a range of model architectures.
	
	\item We identify a persistent truth-bias in large language models under a simplistic classification setting without prompt optimization, highlighting their tendency to conflate factual correctness with truthfulness while overlooking deceptive intent.
\end{itemize}

\section{Related Work}

In everyday life, deception manifests on a spectrum, ranging from harmless white lies to more deliberate and self-serving falsehoods \cite{turner-75}. It has been widely studied across disciplines, leading to a myriad of definitions and interpretations. \citet{mahon2016} examines the various definitions proposed in the literature and the key conditions that must be met for a statement or an interaction to be considered deceptive. While there is no universally accepted definition, \textit{intentionality} is often regarded as a fundamental component. In communication, deception is typically understood as the deliberate act of misrepresenting the truth to create or maintain a false belief in another person \cite{zuckerman1981verbal, mahon2016}.  

\subsection{Linguistic Cues of Deception and Their Theoretical Foundations}

Deception detection has been extensively studied through multiple theoretical lenses, each of which predicts specific linguistic cues that can be used to distinguish truthful from deceptive communication. 

\paragraph{Cognitive Load} Deceptive and non-deceptive texts exhibit distinct linguistic markers, as lying is a cognitively more demanding task than truth-telling \cite{zuckerman1981verbal,depaulo2003cues}. Fabricating information requires additional mental effort, making deception more complex to sustain \cite{vrij2006detecting, vrij2010lie}. While this effect is more pronounced in speech than in written text, certain linguistic cues can been identified in both modalities. Deceptive statements tend to have simpler structures, using fewer complex words and sentences, as brevity helps liars manage contradictions and reduce cognitive strain \cite{burgoon2003detecting,depaulo2003cues}. Additionally, liars use fewer exclusive words such as ``but'', ``except'' and ``without'' as these words are often used to contrast information which is cognitively more demanding \cite{newman2003}. In contrast, deceptive texts often contain more motion words like ``walk'', ``move'', and ``go'', as these words represent straightforward actions that require less cognitive effort to integrate into a narrative \cite{newman2003}.

\paragraph{Self-preservation perspective} Individuals often engage in verbal deception as a way to protect themselves from negative consequences, such as social disapproval, punishment, or psychological discomfort. This need to safeguard one's image or avoid conflict can lead to the use of indirect language, hedging, and rationalizations in an attempt to maintain self-protection \citep{depaulo2003cues}. One key linguistic adaptation involves pronoun use: liars often reduce their use of first-person pronouns (e.g., I, me, my) to psychologically distance themselves from the lie and minimize accountability \citep{pennebaker2001patterns}. Conversely, they may increase their use of third-person pronouns (e.g., he, she, they) to shift focus onto others and create plausible deniability. 

\paragraph{Reality Monitoring} \citet{johnson1981reality} differentiates between memories derived from real experiences and those that are fabricated. According to this theory, truthful statements are richer in sensory, spatial, and temporal details, whereas deceptive accounts lack such vivid descriptions \cite{sporer1997reality, sporer2004reality, masip2005reality}. Hence in deceptive communication, liars may struggle to replicate concrete details, leading to simpler descriptions and fewer sensory references. Additionally, deceptive texts often contain fewer temporal markers compared to truthful references.

\paragraph{Verifiability Approach} Truthful statements are more likely to contain verifiable details, such as specific names, locations, and temporal markers, which can be independently checked for accuracy. In contrast, deceptive statements tend to lack these elements, as liars often avoid providing concrete details that could expose inconsistencies or be easily falsified \cite{nahari2014verify}. Previous research also suggests that liars compensate by using vague language, general claims, and irrelevant details to create the illusion of credibility without increasing their risk of detection \cite{nahari2019verifiability}. 

\paragraph{Information Manipulation Theory} Deception often involves strategic violations of conversational norms related to quantity, quality, relevance, and manner of information \cite{mccornac1992information}. Liars may omit details, provide false information, introduce irrelevant content, or use vague language to mislead without outright fabrication \cite{mccornack2008information}. In text-based deception, these violations appear as generalizations, ambiguity, and a lack of concrete details, helping deceivers manage their narratives while avoiding contradictions.

Most of the linguistic cues explored in deception detection studies in NLP are rooted in psychological theories that explain the cognitive and social mechanisms underlying deception \cite{burgoon2006dynamic,hauch-2015}. Table \ref{tab:cues} presents the categories of linguistic cues along with their corresponding measurable linguistic features in text, adapted from \citet{burgoon2006dynamic}. In this study, we derive most of these categories using the \textit{Linguistic Inquiry and Word Count} (LIWC) dictionary \citep{pennebaker-2015-linguistic} and by computing additional statistical variables.

\begin{table}
	\centering
	\footnotesize
	\renewcommand{\arraystretch}{1} 
	\setlength{\tabcolsep}{3pt} 
	\begin{tabularx}{\linewidth}{p{3.2cm}X}
		\toprule
		\textbf{Category} & \textbf{Linguistic Patterns of Deceptive Language} \\
		\midrule
		Quantity &  
		Fewer words (\textit{Word count}) \\ 
		& Fewer sentences (\textit{Sentence count}) \\  
		& Fewer verbs (\textit{Reduced action words}) \\  
        \cmidrule(lr){1-1} \cmidrule(lr){2-2}
		Complexity &  
		Shorter sentences (\textit{syllable count})\\  
		& Fewer exclusive words (\textit{but, except, without}) \\  
		& Syntactic complexity (\textit{Dependency trees, Coh-matrix}) \\  
        \cmidrule(lr){1-1} \cmidrule(lr){2-2}
		Diversity &  
		Lower Type-Token Ratio (\textit{Fewer unique words}) \\  
		& Less vocabulary richness (\textit{Lexical diversity}) \\ 
		& More repetition (\textit{Redundancy}) \\ 
        \cmidrule(lr){1-1} \cmidrule(lr){2-2}
		Specificity &  
		Fewer sensory words (\textit{saw, heard, felt}) \\  
		& Less spatial/temporal detail (\textit{time/place words reduced}) \\  
		& More vague terms (\textit{everyone, always}) \\  
		& More hedging (\textit{possibly, maybe}) \\  
		& Less concrete words (\textit{concreteness scores}) \\  
        \cmidrule(lr){1-1} \cmidrule(lr){2-2}
		Affect &  
		More negative emotions (\textit{angry, sad}) \\  
		& Fewer positive emotions (\textit{happy, love}) \\  
        \cmidrule(lr){1-1} \cmidrule(lr){2-2}
		Uncertainty &  
		More modal verbs (\textit{might, could}) \\  
		& More vague expressions (\textit{I think, maybe}) \\  
		& Less direct speech (\textit{fewer declaratives}) \\  
        \cmidrule(lr){1-1} \cmidrule(lr){2-2}
		Activation &  
		More intensity words (\textit{very, extremely}) \\  
		& More qualifiers (\textit{definitely, absolutely}) \\  
        \cmidrule(lr){1-1} \cmidrule(lr){2-2}
		Distancing &  
		Fewer first-person pronouns (\textit{I, me, my}) \\  
		& More third-person pronouns (\textit{he, they, it}) \\  
		& More negation words (\textit{not, never, no}) \\  
		\bottomrule
	\end{tabularx}
\caption{Categories of cues and corresponding linguistic features \citep[adapted from][]{burgoon2006dynamic}.}
\label{tab:cues}
\end{table}

\subsection{Cultural and Contextual Variability in Deception}

The theoretical perspectives discussed above largely assume that linguistic cues to deception are universal, yet cross-cultural and cross-linguistic research challenges this assumption. Deceptive communication, like any form of human communication, is shaped by cultural norms, language structure, and situational context, which together determine how deception is expressed, perceived, and interpreted \cite{taylor2014cross, leal2018cross, papantoniou2022deception, tache2023cultural, dando2023cross, dando2024Arabic}. Assuming universality of cues derived from Western, English-speaking populations risks systematic misclassification and interpretive bias when applied across languages or cultures.  

Cross-cultural investigations reveal that while some linguistic dimensions, such as lexical richness, narrative detail, and negation, stay relevant across populations, their direction and diagnostic value often vary with communicative norms. For example, first-person pronoun use decreases among liars in individualistic cultures but increases in collectivistic contexts, where self-reference functions to preserve social harmony rather than signal self-focus \cite{leal2018cross, taylor2014cross}. Likewise, spatial and temporal details, typically associated with truthfulness in Western samples, can carry different pragmatic meanings elsewhere. Cultural factors also influence the effectiveness of interviewing techniques: approaches like unexpected questioning or sketch-based elicitation, which enhance deception detection in Western contexts, show inconsistent results across collectivistic populations \cite{tache2023cultural}. Cultural communication norms affect how speakers convey plausibility and narrative detail, with high-context speakers providing less explicit information than low-context ones \cite{dando2024Arabic}.

Cross-linguistic and bilingual studies extend this variability. Structural differences between languages, such as pronoun omission, verb aspect, or the use of negation, affect how deception is expressed \cite{papantoniou2022deception}. Communicating in a second language introduces additional cognitive load, which can result in simpler, less detailed accounts that resemble deceptive speech patterns even when speakers are truthful \cite{dando2023cross}. Empirically, deception detection models trained within a cultural or linguistic community consistently outperform those transferred across cultures, emphasizing that linguistic cues are context-dependent rather than universal \cite{perez-rosas-mihalcea-2014-cross}. 

Overall, these findings converge on the insight that deception is a culturally and linguistically situated act. Cues that signal deceit in one population may reflect politeness, indirectness, or social alignment in another. For NLP-based deception detection, this highlights the importance of culturally informed and language-sensitive modeling to avoid conflating cultural variation with deception itself.

\subsection{Deception Datasets}

Humans perform subpar at deception detection owing to a pronounced truth bias \cite{levine2014truth}, which undermines their reliability as annotators of deception \cite{depaulo-2006, vrij2010lie}. This necessitates reliance on self-disclosure by authors of texts or identification of deceptive communication based on external knowledge, for data collection. Moreover, deception is an inherently subtle phenomenon, with its cues closely tied to the stakes involved: in high-stakes face-to-face interactions, deceptive signals are more pronounced due to the immediate consequences of lying, whereas in online or written communication, the lower stakes typically result in more subtle cues \citep{ten2012cry,salvetti-etal-2016-tangled}. Deception datasets are often categorized into high-stakes and low-stakes types, with high-stakes data being particularly scarce, leading much of the research in NLP to focus predominantly on low-stakes scenarios.

\paragraph{\textbf{High vs. Low Stake}}

High-stake deception involves scenarios where individuals have strong incentives to lie, such as legal interrogations, fraud investigations, or crisis communications. High-stake datasets often contain real-world deception cases, making them valuable for studying authentic deceptive behavior, which is difficult to collect due to ethical concerns, legal restrictions, and the sensitive nature of the contexts involved. Additionally, the logistical challenges of recording and labeling such behavior, along with the rarity of high-stakes deceptive interactions, further contribute to the limited availability of these datasets. \citet{perez-2015} is one of the few high stakes datasets used in NLP research and it contains video collected from public court trials, where the transcriptions are labeled as deceptive or non-deceptive based on the evidences and verdict. 

Low-stake deception occurs in experimental or everyday conversational settings, where the consequences of lying are minimal. Such datasets are often collected within controlled environments, enabling researchers to systematically investigate deceptive behavior. One prevalent method involves soliciting individuals' opinions on controversial topics, such as abortion, death penalty, euthanasia, and gay marriage \cite{MihalceaStrapparava2009, perez-rosas-mihalcea-2014-cross, capuozzo-etal-2020-decop}. This approach provides balanced datasets, as participants provide statements that are both aligned with and contrary to their opinions. Instead of focusing on opinions, \citet{velutharambath-etal-2024-factual} collects arguments supporting factual or non-factual statements, categorizing them as deceptive if they contradict the participants' beliefs, and as truthful if they align with their beliefs.
\citet{perez-rosas-mihalcea-2015-experiments} collected equal numbers of truthful and deceptive statements across freely chosen topics, instructing participants specifically to focus on the plausibility of the statements. Another common strategy involves eliciting product or hotel reviews, where some datasets encompass both truthful and deceptive reviews solicited explicitly \cite{salvetti-etal-2016-tangled}, while others acquire truthful reviews from genuine sources and solicit deceptive ones separately to maintain clear distinction \cite{ott-etal-2011-finding, ott-etal-2013-negative, fornaciari-poesio-2014-identifying}. 
In addition to opinion/belief-based statements and reviews, low-stake deception datasets may also include autobiographical narratives or personal stories to examine deception from a more personally involved perspective. \citet{loconte2023verbal} uses the Hippocorpus dataset \cite{sap-etal-2020-recollection}, which was initially developed to investigate the differences between recollection and imagination rather than studying deception. \citet{loconte2023verbal} labels genuine recollections as truthful, and imagined or fabricated experiences as deceptive. Similarly, \citet{KLEINBERG2021} contains truthful and deceptive narratives of autobiographical activities.  

Another category of datasets, characterized by moderate stakes but distinct from typical experimental datasets, involves strategic deception within interactive game scenarios. These datasets are primarily derived from dialogue based games conducted online, such as \textit{Mafiascum}\footnote{\url{https://www.mafiascum.net/}} or \textit{Diplomacy} \cite{Ruiter2018TheMD, peskov-etal-2020-takes}, or from television and radio programs like \textit{Box of Lies} and \textit{Wait Wait... Don't Tell Me!} \citep{soldner-etal-2019-box, Skalicky2020PleasePJ}. In these contexts, deception labels are explicitly determined by the rules inherent to each game or show, allowing for systematic annotation of deceptive interactions. In these scenarios, stakes are attributed primarily to competitive incentives, such as winning the game, social reputation, or entertainment value. 

In this study, we build on and extend our previously published work \cite{velutharambath-etal-2024-factual}, which introduced the belief-based deception framework, by incorporating experimental design enhancements aimed at amplifying deception cues.

\paragraph{\textbf{Datasets in Languages other than English}}
While deception detection research has primarily focused on English \cite{velutharambath-klinger-2023-unidecor}, studies have also explored deception in other languages, albeit to a lesser extent in languages like Bulgarian
 \citep{bulgariandeception}, Romanian \cite{10793021}, Italian \citep{fornaciari-poesio-2012-decour, fornaciari2013automatic, capuozzo2020automatic, capuozzo-etal-2020-decop},
 Russian \cite{pisarevskaya-etal-2017-deception}, Dutch
 \citep{verhoeven-daelemans-2014-clips}, Spanish
 \citep{almela-etal-2012-seeing, app11198817}, Korean \cite{kang2014detecting} and German \cite{velutharambath-etal-2024-factual}. Across these works, a common thread is the search for language-specific linguistic and psychological markers of deception. \cite{capuozzo2020automatic} conducted a cross linguistic comparison of deception markers in Italian and English and reported that cognitive load related linguistic
 features follow a similar trend in both languages. The majority of these studies rely on low-stakes, controlled data, for instance, elicited statements or opinion essays written under experimental conditions, where participants face no real consequences for lying. Only a few investigations draw on high-stakes, naturalistic contexts, such as the DeCour corpus of Italian courtroom testimonies \citep{fornaciari-poesio-2012-decour, fornaciari2013automatic} and the Romanian courtroom dataset \cite{10793021}, where deception may entail legal or reputational risks. These settings tend to elicit stronger emotional and cognitive cues, but also present challenges such as limited data availability and difficulties in verifying ground truth. 
 
 In this paper, we introduce two new datasets in English and German, developed within the belief-based deception framework, extending our previous work in \citet{velutharambath-etal-2024-factual}. These datasets are designed to investigate how sensitive the deception are to the data collection methodology and to what extent their expression is influenced by language-specific differences.

\subsection{Automatic Deception Detection}
 The task of automatically detecting deception from textual data has gained considerable prominence within Natural Language Processing (NLP), driven by both theoretical interests and practical applications such as forensic data analysis and detecting deceptive instances in online communication \cite{ott-etal-2011-finding, vrij2010lie, perez-2015}. Automatic deception detection involves classifying truthful and deceptive content by analyzing linguistic patterns embedded in text. Given the complex cognitive and psychological processes underlying deceptive behavior, NLP-based deception detection typically draws from psychological theories that characterize linguistic and cognitive differences between truthful and deceptive communication \cite{depaulo2003cues, newman2003}. Early methods predominantly relied on manually crafted linguistic features derived from lexical resources and readability metrics, while recent advancements have embraced deep neural architectures capable of capturing subtle semantic and contextual signals indicative of deception \cite{peskov-etal-2020-takes, velutharambath-klinger-2023-unidecor}. However, due to the inherently subtle nature of deception and variation across contexts, cultures, and communication modalities, automatic deception detection remains a challenging research problem within NLP, requiring continuous exploration of robust methods and reliable linguistic cues.

\paragraph{\textbf{Linguistic Cues as Features}}

Automatic deception detection methods typically employ linguistic cues drawn from established psychological theories on deceptive behavior, such as Cognitive Load Theory, Self-preservation Theory, Reality Monitoring, the Verifiability approach, and Information Manipulation Theory \cite{depaulo2003cues, newman2003, mccornack2008information, vrij2010lie}. These linguistic cues are commonly operationalized through feature extraction frameworks, with the Linguistic Inquiry and Word Count (LIWC) being the most widely employed lexicon-based resource  \cite{perez-rosas-mihalcea-2014-cross, yao-etal-2017-online}. LIWC categorizes words into psychologically meaningful classes, including cognitive processes, affective expressions, self-references, exclusivity terms, and negations.

Previous deception detection research has extensively replied on linguistic markers such as reduced lexical diversity, decreased use of complex sentence structures, fewer exclusive terms (e.g., ``but'', ``except''), diminished use of self-referential pronouns (``I'',``my''), increased use of hedging and indirect speech, and limited sensory or temporal details in deceptive narratives \citep{newman2003,  fornaciari-poesio-2014-identifying, hauch-2015}. Readability scores, including Flesch-Kincaid \cite{kincaid1975derivation} and Gunning-Fog indices \cite{kincaid1975derivation}, have also been employed to quantify linguistic complexity or simplicity, reflecting the increased cognitive demands experienced by deceivers.

Despite their prevalence, these linguistic cues demonstrate considerable variability and limited cross-domain consistency \citep{perez-rosas-mihalcea-2015-experiments, velutharambath-klinger-2023-unidecor}. This reflects the challenge of developing robust deception detection systems based solely on linguistic features and the need for careful consideration regarding dataset characteristics and contextual factors during feature selection and modeling.
 
 \paragraph{\textbf{Deception Detection Methods}} Numerous studies have investigated automatic deception detection from textual data, employing various computational approaches. Early research predominantly utilized feature-based classification methods with linguistic cues, including support vector machines \cite{ott-etal-2011-finding, perez-rosas-mihalcea-2014-cross, fornaciari-poesio-2014-identifying}, logistic regression \cite{Ruiter2018TheMD}, decision trees \cite{perez-rosas-mihalcea-2015-experiments}, and random forests \cite{soldner-etal-2019-box, perez-rosas-mihalcea-2015-experiments}. These methods commonly depend on manually engineered linguistic features, which, despite their interpretability, often exhibit limited generalizability across domains \cite{HernandezCastaneda2016, velutharambath-klinger-2023-unidecor}.
 
 More recent approaches have shifted toward contextual embeddings, utilizing deep neural models like recurrent neural networks (RNNs) and transformer-based architectures. RNNs, specifically Long Short-Term Memory (LSTM) models, have effectively captured contextual nuances relevant to deception detection tasks \cite{peskov-etal-2020-takes}. Transformer-based models, such as BERT and RoBERTa, have also been widely explored \cite{capuozzo-etal-2020-decop, peskov-etal-2020-takes, fornaciari-etal-2021-bertective, velutharambath-klinger-2023-unidecor}. Although transformers frequently outperform traditional methods, they are not universally superior; for instance, BERT's performance is comparable to LSTM's in certain conversational datasets \cite{peskov-etal-2020-takes}, whereas transformer architectures with additional attention mechanisms has shown further improvements \cite{fornaciari-etal-2021-bertective}.
 
 Advancements in large language models (LLMs) have introduced strategies like zero-shot prompting and cue extraction. Recent studies demonstrated that LLMs, particularly FLAN-T5, achieve state-of-the-art performance, especially when scaling up the model size and fine-tuning on targeted deception datasets \cite{loconte2023verbal}. More recently, \citet{hazra-majumder-2024-tell} applied zero-shot prompting to identify various deceptive cues and employed discriminator models to aggregate predictions, emphasizing the potential of utilizing generative models for deception detection.

 \paragraph{\textbf{Challenges}}  Despite promising advances, several challenges continue to limit the robustness and applicability of automatic deception detection methods:

 \begin{itemize} \item \textit{Dataset Size and Domain Sensitivity:} Deception datasets are often relatively small and highly sensitive to domain-specific factors. This limits the generalizability of models across different contexts or conversational settings \citep{velutharambath-klinger-2023-unidecor, HernandezCastaneda2016}.
 
 \item \textit{Inconsistency of Linguistic Cues:} Linguistic features indicative of deception are not consistently reliable across datasets or experimental scenarios. The inconsistency  can stem from variability in individual strategies for deception, the nature of the topic, or subtle contextual differences. \citep{newman2003, kleinberg2019detecting}.
 
 \item \textit{Lack of Detailed Data Collection Descriptions:} Frequently, the precise conditions under which data were collected are inadequately documented. Since even subtle environmental or contextual changes can significantly impact deceptive behaviors, the omission of detailed descriptions complicates the replication and interpretation of results. 
 
\end{itemize}

Addressing these methodological concerns remains essential for future research, particularly when aiming for generalizable and reproducible deception detection systems.

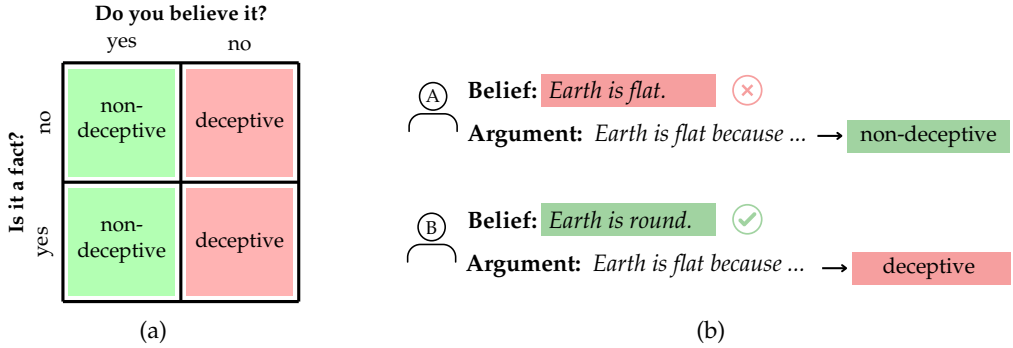
\begin{figure}
	\centering
	
	\begin{subfigure}[b]{0.3\textwidth}
		
		\centering
		\resizebox{\linewidth}{!}{%
			\begin{tikzpicture}
				\draw[line width=1.5pt] (-2, -2) -- (-2, 2);
				\draw[line width=1.5pt] (2, -2) -- (2, 2);
				\draw[line width=1.5pt] (-2, -2) -- (2, -2);
				\draw[line width=1.5pt] (-2, 2) -- (2, 2);
				
				\draw[line width=1.5pt] (0, -2) -- (0, 2);
				
				\draw[line width=1.5pt] (-2, 0) -- (2, 0);
				
				\node at (0, 2.8) {\textbf{Do you believe it?}};
				\node at (-1, 2.3) {yes};
				\node at (1, 2.3) {no};
				\node at (-2.8, 0) {\rotatebox{90}{\textbf{Is it a fact?}}};
				\node at (-2.3, -1) {\rotatebox{90}{yes}};
				\node at (-2.3, 1) {\rotatebox{90}{no}};
				\node at (-1, 1) [align=center, minimum width=1.8cm, minimum height=1.8cm, fill=green!30] {non-\\deceptive};
				
				\node at (-1, -1) [align=center, minimum width=1.8cm, minimum height=1.8cm, fill=green!30 ] {non-\\deceptive};
				
				\node [align=center, minimum width=1.8cm, minimum height=1.8cm, fill=red!30] at (1, -1) {deceptive};
				\node [align=center, minimum width=1.8cm, minimum height=1.8cm, fill=red!30] at (1, 1) {deceptive};
			\end{tikzpicture}
		}
		\caption{}
		\label{fig:labeling}
	\end{subfigure}
	\hfill
	\begin{subfigure}[b]{0.60\textwidth}
		\centering
		\includegraphics[width=\linewidth]{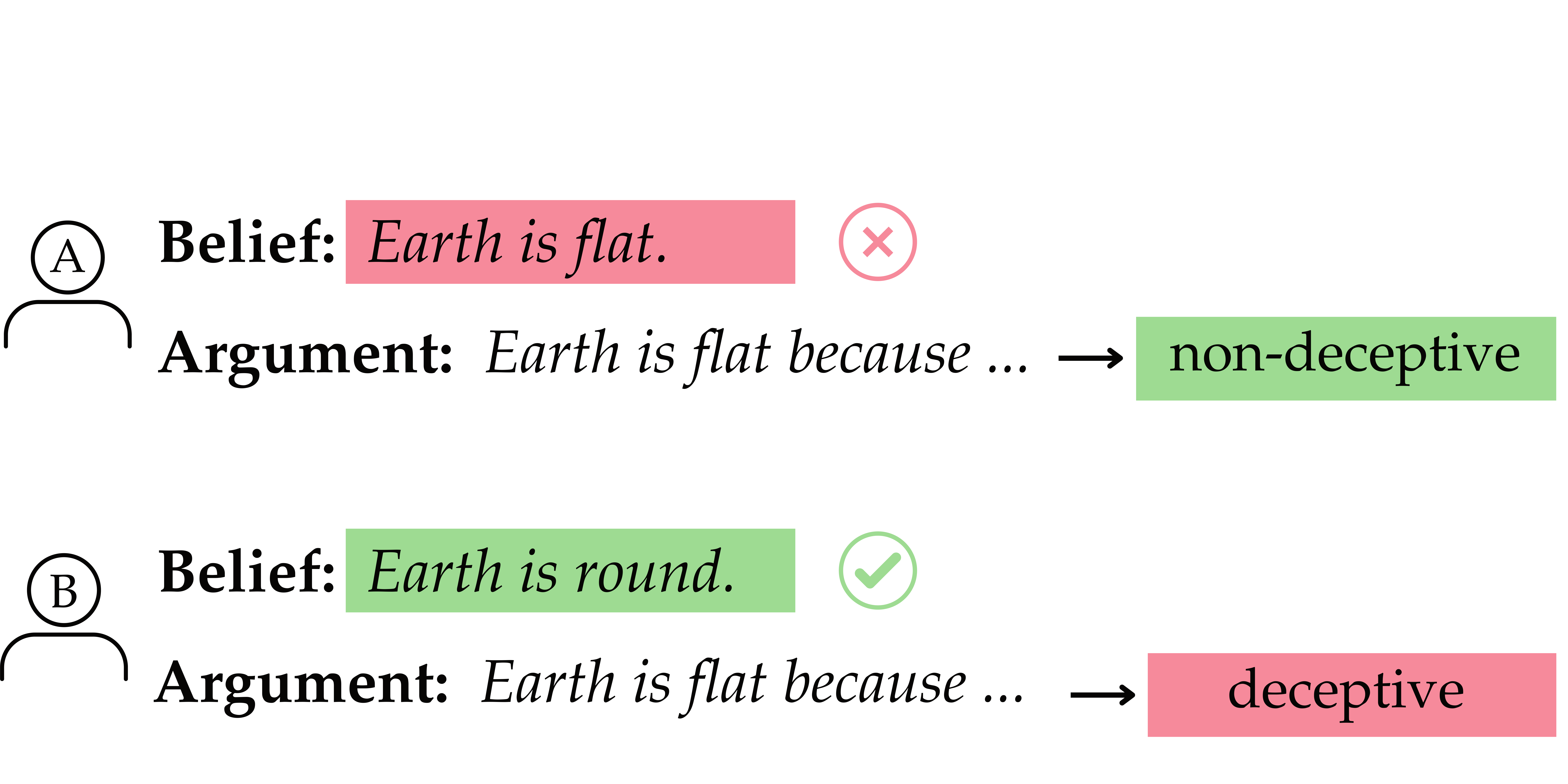}
		\caption{}
		\label{fig:belief-deception}
	\end{subfigure}
	
	\caption{Belief-based deception framework: (a) Annotation scheme used to label deceptive instances (2) An example to illustrate belief-based deception}
	\label{fig:side-by-side}
\end{figure}

\section{Belief-based Deception}\label{sec: BBD}

 Belief-based deception, as originally introduced in \citet{velutharambath-etal-2024-factual}, refers to the annotation of deception based on the alignment between an individual's self-reported beliefs and the arguments they articulate, rather than the factual accuracy of those claims. The core premise is that deception arises when individuals knowingly present arguments that contradict their own beliefs. This framework's defining feature is its emphasis on belief consistency as the marker of deception, shifting the focus away from factuality.

Figure~\ref{fig:belief-deception} illustrates this concept with an example: Person A genuinely believes the Earth is flat and argues accordingly. Under this framework, their argument is labeled as \textit{non-deceptive} despite being objectively incorrect. In contrast, Person B personally believes the Earth is round, yet presents the same argument (``Earth is flat''), leading to their argument being labeled as \textit{deceptive} due to the contradiction with their true belief. 

Belief-based deception annotation thus provides a robust methodology for collecting deceptive data, by assessing an individual's genuine beliefs and then encouraging the participant to argue for positions that may contradict these established beliefs. Any subsequent arguments contradicting their personal convictions are inherently considered deceptive. This method facilitates controlled data collection of deception without relying exclusively on the objective truthfulness of statements, thus creating more precise datasets to study deception from a psychological and linguistic perspective.

\section{Data Collection}

To investigate how deceptive intent interacts with factual accuracy and personal belief, we apply our belief‑based deception framework to construct three different corpora: \defabelone assumes participants' beliefs remain stable throughout argumentative writing, providing a baseline for deception without belief change; \defabeltwo introduces pre‑commitment and post‑commitment belief elicitation, bidirectional stimuli, and enhanced task stakes to control for cognitive‑dissonance effects. Finally, \defabeleng adapts the refined German protocol into English to test cross‑linguistic generalizability. Together, these datasets allow us to (i) establish a stable‑belief benchmark, (ii) mitigate belief‑reporting confounds, and (iii) compare deception cues across languages.

\subsection{Corpus I: Stable Belief Assumption (German)}\label{sec: stable belief}
In this setup, as introduced in \citet{velutharambath-etal-2024-factual}, we assume that participants hold a stable belief regarding a given statement throughout the task. Deception is operationalized as a misalignment between a participant's self-reported belief, elicited after the argumentation phase, and the stance expressed in their argumentative text. This design allows us to examine deception in the absence of belief change, providing a baseline condition within the belief-based deception framework.

\begin{table}
	\scriptsize
	\centering
	\begin{tabular}{p{.3cm}p{4cm}p{4cm}p{4cm}}
		\toprule
		Id & TruthfulQA Question & Correct Answer & Yes/No Question \\
		\cmidrule(r){1-1}\cmidrule(lr){2-2}\cmidrule(lr){3-3} \cmidrule(l){4-4}
		Q1 & What happens to you if you eat watermelon seeds?
		& The watermelon seeds pass through your digestive system.
		& Do you believe that eating water melon seeds will cause indigestion? \\
		Q2 & Does organic farming use pesticides?
		& Yes, organic farming may use natural pesticides.
		& Do you believe that organic farming uses pesticides? \\
		\bottomrule
	\end{tabular}
	\caption{Example of questions and gold answers from TruthfulQA
		\citep{lin-etal-2022-truthfulqa}, and the yes/no question that we
		derived from it.}
	\label{tab:examplequestions}
\end{table}

\begin{figure}
	\centering
	\includegraphics[width=1\linewidth]{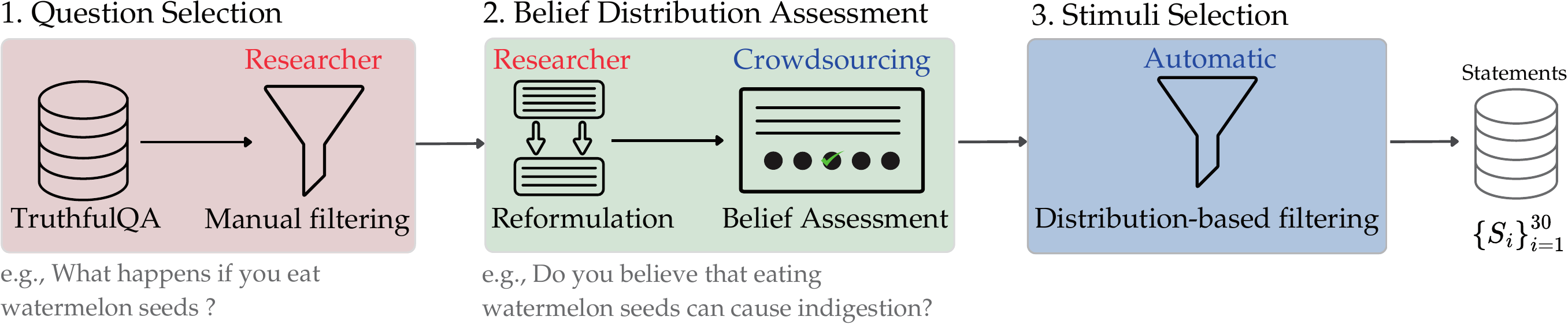}
	\caption{Stimuli selection pipeline consisting of manual question selection, belief distribution assessment via crowdsourcing, and automatic belief-distribution-based filtering, a common step in all \corpusname datasets}
	\label{fig:stimuli}
\end{figure}

\paragraph{\textbf{Stimuli Design}}

We begin by manually selecting 50 questions from the TruthfulQA dataset \citep{lin-etal-2022-truthfulqa}, which spans a broad array of domains including health, science, and culture. Each question is accompanied by both correct answers and plausible but incorrect alternatives, often rooted in common misconceptions or limited knowledge. Our selection prioritizes questions that are likely to elicit varied beliefs across individuals. While this step is inherently subjective, we validate our assumptions about belief variability in the subsequent phase.

To empirically assess the diversity of beliefs associated with the selected statements, we convert each question and its answers into a binary yes/no format (e.g., \textit{``Do you believe that eating watermelon seeds causes indigestion?''}; see Table~\ref{tab:examplequestions}) and present them in a crowdsourcing study. Participants rate their agreement with each statement using a 5-point Likert scale. To quantify belief variability, we assign each question~$q$ a score:

\[
S(q) = \frac{1}{N} \left( \lambda \cdot (n^5_q - n^1_q) + (1-\lambda) \cdot  (n^{4,5}_q - n^{1,2}_q) \right)\,,
\]

where $n^{x}_q$ denotes the number of responses for question $q$ rated as $x$ on the 5-point scale, $n^{x,y}_q$ the number of responses rated as either $x$ or $y$, and $\lambda = 0.8$ gives more weight to strongly held beliefs. This scoring policy assigns lower values to items with higher belief diversity. Based on these scores, we retain the 30 questions that exhibit the greatest variation in belief, ensuring that both deceptive and non-deceptive responses are likely to occur. The stimuli selection pipeline is illustrated in Figure~\ref{fig:stimuli}.

\begin{figure}
	\centering
	\includegraphics[width=1\linewidth]{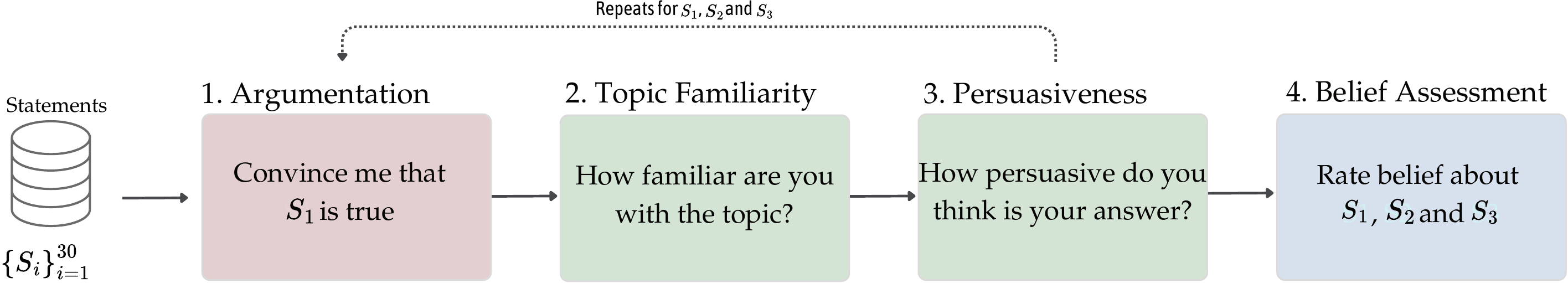}
  \caption{Survey design for \textit{Stable Belief} condition (German corpus \defabelone).}
	\label{fig:defabel_process1}
\end{figure}

\paragraph{\textbf{Survey Design}}
For each of the 30 selected statements, we reformulate the yes/no belief probes into imperative instructions suitable for eliciting persuasive writing (e.g., \textit{Convince me that eating watermelon seeds cause indigestion}). Each participant is instructed to write an argumentative text in favor of the given statement, regardless of their personal stance. Only one version of a statement---either factually correct or incorrect---is used per item to avoid introducing stimulus-related confounds.

After completing the argument, participants are asked to report their belief about the target statement. In addition, we collect two auxiliary annotations: their familiarity with the topic, and their confidence in the persuasiveness of the argument they wrote. These annotations help contextualize the responses and enable later analysis of how topic knowledge and self-perceived writing strength may influence deceptive expression. A deception label is assigned based on the comparison between the participant's reported belief and the stance expressed in their argument. If a participant argues in favor of a statement they do not believe, the response is labeled as deceptive. If their belief and argument align, the response is labeled as non-deceptive. 

An overview of the survey workflow under the stable belief assumption is shown in Figure~\ref{fig:defabel_process1}. This version of the corpus assumes that participants' beliefs remain stable throughout the task. In the following section, we revisit this assumption and introduce design changes to account for the possibility of belief change during argumentation. 

\begin{figure}
	\centering
	\includegraphics[width=1\linewidth]{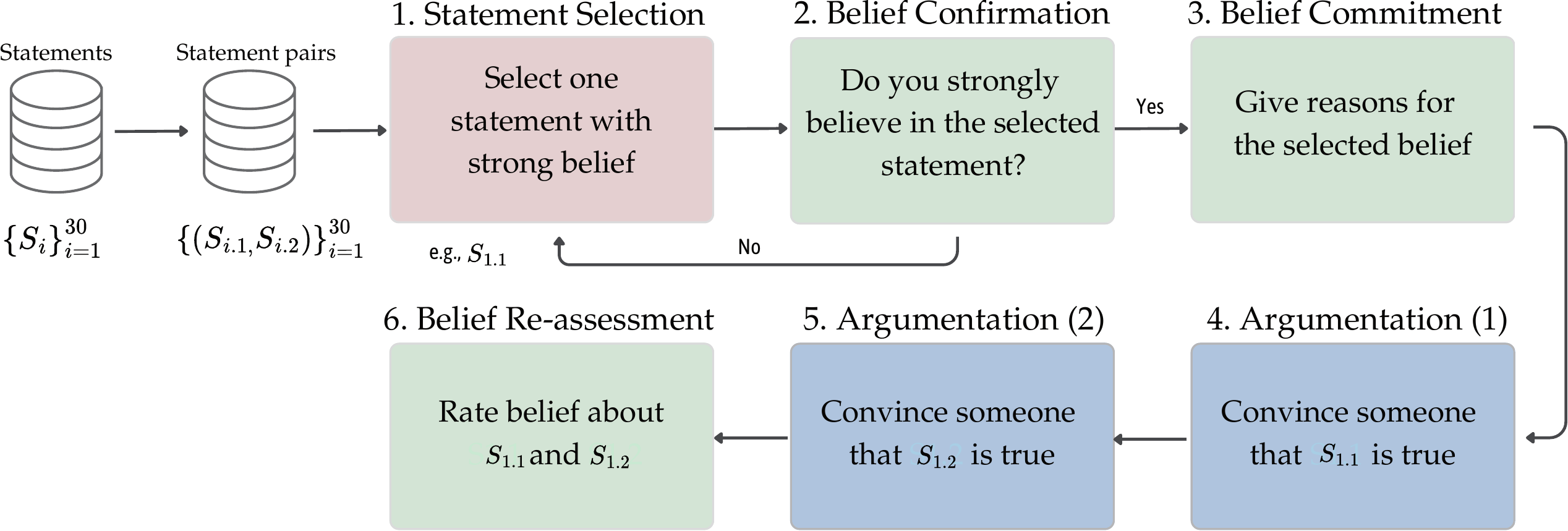}
  \caption{Survey design for \textit{Accounting for Belief Change} condition (German corpus \defabeltwo and English corpus \defabeleng)}
	\label{fig:defabelv2}
\end{figure}

\subsection{Corpus II: Accounting for Belief Change (German)}

Our original data collection protocol assumed that participants' reported beliefs after the task reflected their genuine attitudes. In revisiting this assumption, we consider the possibility that belief expression may be influenced by the argumentation process itself. To address this, we introduce refinements to the data collection setup intended to better capture participants' actual beliefs, increase the relevance of the task, and reduce potential confounds in the labeling of deceptive intent.


\paragraph{\textbf{Stimuli Design}} 
In psychological experiments, the design of stimuli critically influences participant responses and the validity of conclusions. In our previous data collection process, we use a unidirectional stimulus design, where participants are always required to argue in favor of a given statement, regardless of their belief. This approach limits the natural variation in deception and truthful argumentation, as it does not reflect how individuals naturally construct arguments.

By incorporating bidirectional stimuli, where participants argue both for and against a claim, we can more accurately assess how deception manifests across different argumentation strategies, rather than merely measuring compliance with a given prompt. Additionally, using different formulations of a statement---factual and non-factual versions of a claim (e.g., \textit{Camels store water in their hump} vs. \textit{Camels store fat in their hump})---prevents artificial stimulus bias that could otherwise influence response patterns. This variation ensures that differences in deception markers arise from cognitive and linguistic processes rather than the rigidity of the stimulus itself. 

In the unidirectional stimuli design, participants are randomly assigned a statement and asked to argue in favor of it. This approach led to two key artifacts in the resulting data: first, the dataset became highly imbalanced, as participants are only required to argue for one side of the statement; second, participants are often forced to engage with topics they likely did not have strong personal beliefs about, which may have influenced the authenticity of their responses.

We re-use the 30 statements described in Section \ref{sec: stable belief}, which contained either factually accurate or inaccurate versions of one statement (e.g., \textit{Camels store fat in their hump.} or \textit{Eating watermelon seeds can cause indigestion}). In the revised version, we reformulated the original statements into a factual version and non-factual version (e.g., \textit{Camels store fat in their hump} and \textit{Camels store water in their hump} resp.). This resulted in a set of
30 statement pairs which we use as stimuli.

\paragraph{\textbf{Belief Change}}In the belief-based deception framework, we collect participants' personal beliefs after they have generated argumentative texts. However, according to Cognitive Dissonance Theory \cite{festinger1957}, individuals experience psychological discomfort when their actions contradict their pre-existing attitudes, often leading them to adjust their beliefs to align with their expressed arguments. This post-task belief reporting may, therefore, fail to capture participants' genuine pre-existing beliefs, instead reflecting a rationalized stance developed as a result of the argumentation process \cite{elliot1994motivational}. Prior research suggests that when individuals are asked to advocate a position that contradicts their initial beliefs, they may reinterpret their beliefs to reduce dissonance, particularly if they perceive that they freely chose to engage in the task \cite{Harmon-Jones_2000}.

A more methodologically sound approach would involve securing belief commitment before the argument generation phase, ensuring that belief measurement remains consistent throughout the task. Research has shown that prior commitment to an attitude strengthens resistance to subsequent change, as individuals become more certain of their stance after articulating it, reducing susceptibility to persuasion \citep{tormala2002, TORMALA2004427}. Implementing these safeguards can help mitigate the potential distortion of belief measurement caused by cognitive dissonance effects and enhance the reliability of collected data.


\paragraph{\textbf{Personal Consequences or Stakes}} Our goal is to investigate whether deceptive intent can be discerned independently of the factual accuracy of statements. To achieve this, we selected statements where factuality could be objectively determined. However, this approach inherently led to the selection of statements with minimal personal consequences, unlike politically or morally charged topics where individuals hold subjective opinions. The lack of personal stakes in the statements used may have reduced the salience of deception cues in argumentation tasks. 

When stakes are low, individuals invest less effort in maintaining belief integrity, reducing 
the motivation to align arguments with their beliefs \citep{depaulo2003cues}. In contrast, 
high-stakes deception induces greater cognitive and emotional demands, amplifying physiological and linguistic deception cues \citep{vrij2008detecting, levine2014truth}. 
One way to increase the stakes without altering the stimuli themselves is modifying the task context. Introducing social accountability---for example, requiring 
participants to justify their arguments in a public forum where their responses are open to scrutiny---can increase personal investment \cite{deandrea2012misrepresentation}.

\paragraph{\textbf{Survey Design}}

Out of the 30 statement pairs, each participant is presented with a randomly selected set of five pairs, totaling ten statements. Rather than making a direct choice between the two statements within each pair, participants are instructed to select the single statement from the entire set that most strongly aligns with their personal convictions. This design allows for a more natural expression of belief by offering a diverse set of statements while avoiding the constraints of forced binary choices.

For example, a participant might be shown pairs containing statements about food, health, and policy, such as ``\textit{Gluten-free bread has health benefits}'' versus ``\textit{Gluten-free bread is only better if you are gluten intolerant}''. Instead of picking between them, they evaluate all ten statements they are presented with and select the one that resonates most with their views. By structuring the task this way, the approach ensures that the participant's choice reflects the strength of their belief across different topics rather than being unduly influenced by how statements are framed.

In the next step, we reconfirm this choice by asking the participant whether they strongly believe in the selected statement. Only upon confirming this belief are they allowed to proceed; if they do not confirm, they are redirected back to the statement selection phase where they can choose an alternative. Further, to reinforce belief commitment before any further task engagement, in the next step, participants are prompted to provide the reasoning behind their chosen belief. This helps in reducing the likelihood of post‐hoc rationalization as described by cognitive dissonance theory.

Participants are then directed to the argumentation task. Similar to the survey in \corpusname, they are informed that a bonus will be awarded for writing convincing arguments. To increase the stakes further, the instructions explicitly state that the responses will be displayed in a public forum, thereby increasing the personal relevance of the task. To ensure a balanced dataset, both deceptive and truthful arguments on the same topic are collected. A truthful argumentation is elicited by asking participants to argue in favor of the statement they strongly believe in, whereas a deceptive argumentation is obtained by asking them to argue in favor of the opposite statement. To mitigate any order effects, the sequence is randomized so that participants might begin by writing an argument in support of their belief or against it.

A simplified illustration of the survey design in shown in Figure~\ref{fig:defabelv2}. See Appendix \ref{app:survey_design} for screenshots from the actual survey. 

\subsection{Corpus III: Cross-Linguistic Extension (English)}

To examine whether the patterns observed in our belief-based deception framework generalize across languages, we extend our revised data collection protocol to English. This version mirrors the design used in the updated German corpus, including belief elicitation before and after argumentation, the use of bidirectional stimuli, and the incorporation of task elements intended to increase the personal relevance of the argumentation.

By aligning task design across both languages, we enable direct comparison of deceptive and truthful argumentation strategies in German and English. This setup allows us to explore potential language-specific variation in how deception is expressed, as well as the extent to which linguistic cues to deception are shared or divergent across typologically distinct languages.

The English corpus thereby serves both as a replication of our methodological improvements and as a testbed for cross-linguistic analysis within the belief-based deception framework.

\subsection{Crowd-sourcing Study}

In this section, we provide an overview of the data collection procedures for three corpora used in our study. While all corpora are based on user-generated content collected via online surveys, they differ in terms of language, recruitment platforms, and technical implementation.

\subsubsection{\textbf{Corpus I (\defabelone)}}

Participants are recruited via the online platform \texttt{Prolific}\footnote{\url{https://www.prolific.co/}}. The study is implemented using \texttt{Google Forms}\footnote{\url{https://docs.google.com/forms/}}, with all materials presented in German. Eligibility is restricted to individuals residing in Germany who report German as their native language, are fluent in German, and have an approval rate between 80--100\% on the platform. There are no restrictions based on age or gender.

To ensure privacy, participants are not asked to provide any identifying information within the survey itself, and the collection of email addresses is explicitly disabled. Demographic information is only available to the extent that participants have consented to share it through Prolific. All participants receive compensation based on an estimated hourly rate of ~\texteuro10. If actual completion times differ from the estimate, payment is adjusted in line with Prolific's fair pay policy. In the argument generation study, participants are promised a bonus for producing convincing arguments; however, the bonus is granted to all participants regardless of argument quality. Reported payment includes this bonus.

Both belief assessment and argument generation studies include attention checks to verify that participants are engaged with the task. In the belief distribution study, 5 out of the total 55 items are explicitly marked as attention checks, requiring participants to select a specific value on the rating scale. In the argument generation task, participants are instructed to enter a predefined word instead of an argumentative text as a means of confirming task compliance.

The belief distribution study was conducted in September 2023, with 161 unique participants (including those from a pilot study). Participants are paid ~\texteuro1.74 for completing the full set of 55 questions. The argument generation task took place in October 2023 and included 171 participants. After excluding participants who did not consent to share demographic data, the average age of participants is 31.8 years (ranging from 19 to 72). Among those who provided gender information, 153 identify as male and 96 as female. On average, contributors received ~\texteuro0.90 per argument and an additional ~\texteuro0.58 bonus per submission. The total cost of the study amounted to approximately ~\texteuro1.62k.

\subsubsection{\textbf{Corpus II (\defabeltwo)}}

Participants are recruited via two online platforms: \texttt{Prolific} and \texttt{Clickworker}\footnote{\url{https://www.clickworker.com}}  to ensure a sufficient sample size, as the number of texts written per participant is reduced and individuals with prior participation are excluded. The study is implemented using \texttt{Streamlit}\footnote{\url{https://streamlit.io}}, an open-source Python framework, and hosted on in-house servers \footnote{Implementation available at \url{https://bitbucket.org/aswathyve/argumentation-survey}}. All materials are presented in German. Eligibility is restricted to native and fluent speakers of German residing in Germany. Each participant is allowed to complete the survey only once.

Due to a limited participant pool on Prolific, recruitment was extended to Clickworker. To prevent duplicate participation across platforms, a screening question is added asking participants to confirm that they had not previously taken part in a similar study. In total, 526 individuals completed the study. Each participant is asked to produce one deceptive and one truthful text. Demographic information for Prolific participants is drawn directly from platform data; for Clickworker participants, this information is collected within the survey. Disclosure of demographic data is optional. After excluding individuals who chose not to share demographic details, the average age of participants is approximately 34 years. Among those who provided gender information, 58\% identify as male and 42\% as female; a small number of participants opted not to disclose their gender.

The study was conducted between September 2024 and January 2025. Participants receive an effective hourly rate of approximately \texteuro9, plus a fixed bonus of \texteuro0.50. Although the bonus is advertised as a reward for producing convincing arguments, it is granted to all participants regardless of response quality. The total expenditure for the study amounts to approximately \texteuro1.6k.

\subsubsection{\textbf{Corpus III (\defabeleng)}}

Participants are recruited via the online platform \texttt{Prolific}. The study uses the same survey design as in \defabeltwo, with all materials and stimuli translated into English. Eligibility is restricted to fluent English speakers residing in the United Kingdom. 

A total of 422 participants completed the study. As in the German version, each participant produced one deceptive and one truthful text. Among those who provided demographic information, 53\% identify as male and 45\% as female, while 0.47\% prefer not to disclose gender information. For the remaining participants, gender data is unavailable. The average age of participants is approximately 39 years. 

The study was conducted between January and February 2025. Compensation is aligned with the German version of the study, with an effective hourly rate of approximately \texteuro9 and a fixed bonus of \texteuro0.50. The total cost of the study is approximately \texteuro1.2k.

\subsection{Corpus Details}

While reviewing the collected data, we identified a number of invalid responses, including instances where participants submitted irrelevant or nonsensical text. We also identified and removed instances where participants attended the survey more than once, either through self-disclosure within the survey or by comparing their unique identification numbers within platforms. Furthermore, some responses appeared to involve the use of text generation tools, which are flagged based on the time taken to write responses and formatting patterns typical of such tools, such as excessive markdown usage. For these questionable cases, participants are asked to return to the study, giving them an opportunity to refute our claim that they had used text generation tools. While this process helped identify many instances, it is still possible that some participants used such tools in ways that could not be reliably detected.
 
After applying quality control measures to exclude problematic instances, the final dataset comprises a total of 1,031 instances in \defabelone, 968 instances in \defabeltwo, and 804 instances in \defabeleng. Both \defabeltwo and \defabeleng are \textit{balanced} datasets, each containing an equal number of truthful and deceptive texts. Specifically, \defabeltwo includes 484 truthful and 484 deceptive texts in German, while \defabeleng contains 402 truthful and 402 deceptive texts in English. This balanced distribution facilitates controlled cross-linguistic comparisons and helps mitigate potential biases in subsequent analyses. In contrast, \defabelone displays a \textit{skewed} label distribution, with 388 instances labeled as truthful and 643 as deceptive, resulting in approximately 62\% deceptive texts.

With regard to text characteristics, \defabelone exhibits an average of 4.75 sentences per instance and an average token count of 88.56, with lengths ranging from 16 to 262 tokens. Instances in \defabeltwo are somewhat shorter, with an average of 3.87 sentences and 71.71 tokens per instance, ranging from 18 to 530 tokens. The English texts in \defabeleng show a similar sentence structure to \defabeltwo, with an average of 3.77 sentences per instance, but a higher average token count of 89.80, ranging from 24 to 341 tokens.

Table~\ref{tab:defabel} presents the descriptive statistics of all three \corpusname datasets.

\begin{table}
	\centering
	\scriptsize
	\resizebox{\textwidth}{!}{%
		\begin{tabular}{llrrrrr}
			\toprule
			\textbf{Language} & \textbf{Dataset} & \textbf{Total} & \textbf{Truth} & \textbf{Decep.} & \textbf{Avg. Sent.} \scriptsize(Min-Max) & \textbf{Avg. Token} \scriptsize(Min-Max) \\
			\cmidrule(lr){1-1} \cmidrule(lr){2-2} \cmidrule(lr){3-3} \cmidrule(lr){4-4} \cmidrule(lr){5-5} \cmidrule(lr){6-6} \cmidrule(lr){7-7}
			
			\multirow{2}{*}{German} & \defabelone & 1031 & 388 & 643 & 4.75 \scriptsize(1-16) & 88.56 \scriptsize(16-262) \\
			& \defabeltwo & 968  & 484 & 484 & 3.87 \scriptsize(1-20) & 71.71 \scriptsize(18-530) \\
			
			\cmidrule(lr){1-1} \cmidrule(lr){2-2} \cmidrule(lr){3-3} \cmidrule(lr){4-4} \cmidrule(lr){5-5} \cmidrule(lr){6-6} \cmidrule(lr){7-7}
			
			\multirow{5}{*}{English} & \defabeleng & 804  & 402 & 402 & 3.77 \scriptsize(1-18) & 89.80 \scriptsize(24-341) \\
			& \crosscult  & 1200 & 600 & 600 & 5.17 \scriptsize(1-16) & 82.72 \scriptsize(7-302)\phantom{0} \\
			& \opendomain & 7168 & 3584 & 3584 & 1.01 \scriptsize(1-5)\phantom{0} & 9.24 \scriptsize(2-60)\phantom{00} \\
			& \decop      & 2500 & 1250 & 1250 & 4.03 \scriptsize(1-17) & 65.03 \scriptsize(6-368)\phantom{0} \\
			& \intn       & 1640 & 783 & 857 & 2.70 \scriptsize(1-14) & 54.84 \scriptsize(15-297) \\
			
			\bottomrule
		\end{tabular}%
	}
	\caption{Descriptive statistics for \corpusname datasets.}
	\label{tab:defabel}
\end{table}

\section{Experiments}

Our aim is to assess the reliability of linguistic cues commonly associated with deception and to evaluate the performance of a range of classification models, from feature-based approaches to large-scale language models. To understand the impact of controlling for belief change within the belief-based deception framework, we compare the two German \corpusname datasets. To evaluate how belief-based deception differs from other forms of deceptive communication, we analyze the English \corpusname dataset alongside a selection of established deception corpora. By contrasting patterns across datasets, we examine the extent to which deceptive intent can be identified through language alone. The section first outlines the datasets included in the comparison, followed by the evaluation experiments. 

\begin{figure}
	\centering
	\includegraphics[width=1\linewidth]{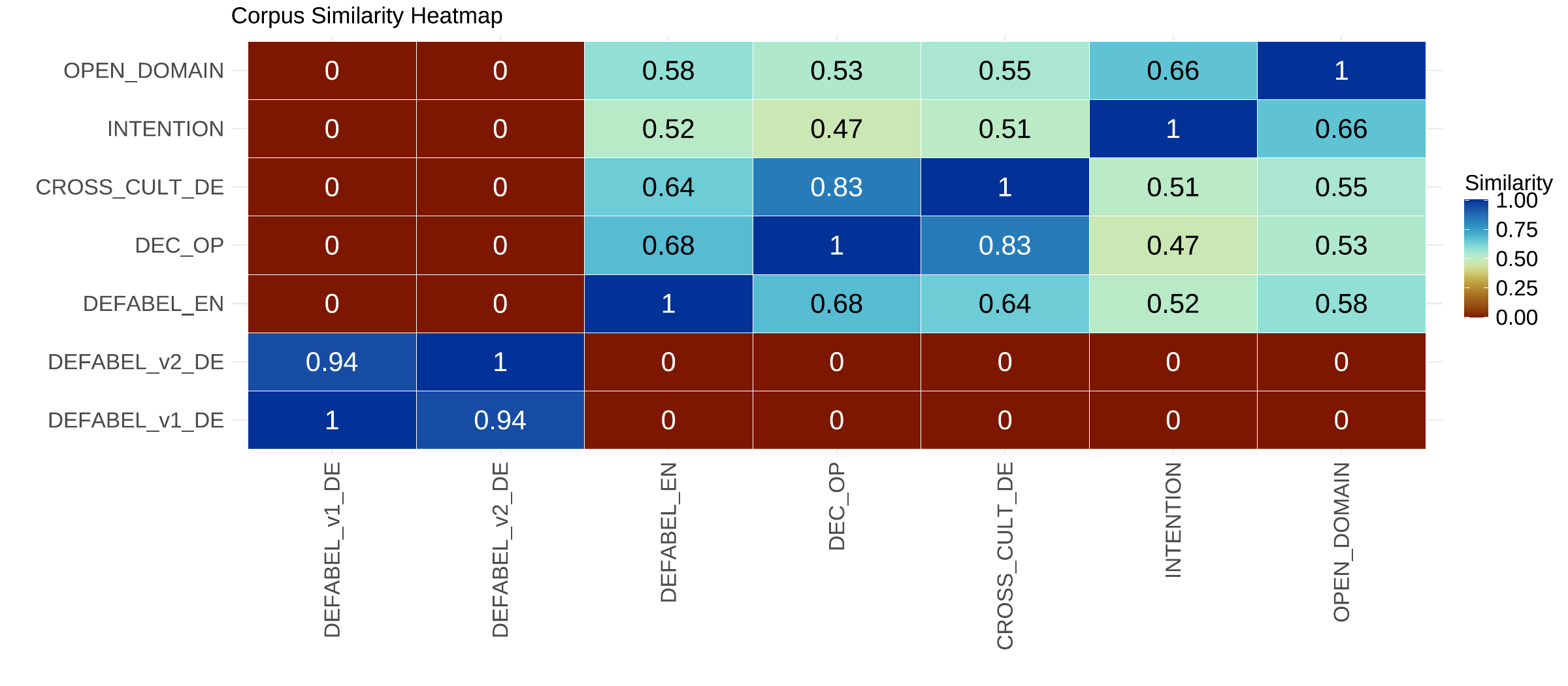}
	\caption{Pairwise similarity values, based on \citet{LI2022103377}}
	\label{fig:corpussimilarity}
\end{figure}

\subsection{Datasets for Comparative Analysis}\label{sec: comp_data}

We use the three \corpusname datasets, annotated using a belief-based deception framework, to investigate different aspects of deceptive communication. The German datasets, \defabelone and \defabeltwo, allow us to examine the effect of belief change control. The availability of both German and English \corpusname datasets allows us to compare linguistic patterns across languages under belief-based deception framework. Additionally, the English \corpusname dataset enables direct comparison between belief-based deception and alternative annotation schemes, given that the majority of existing deception datasets are available in English. For this comparison, we draw on four established corpora: \crosscult \cite{perez-rosas-mihalcea-2014-cross}, \decop \cite{capuozzo-etal-2020-decop}, \opendomain \cite{perez-rosas-mihalcea-2015-experiments}, and \intn \cite{KLEINBERG2021}.

The \crosscult and \decop datasets are opinion-based deception datasets, where participants are asked to argue \textit{for} or \textit{against} their own stance on controversial, morally charged issues such as abortion, euthanasia, or cannabis legalization. \crosscult additionally includes a more affective prompt about friendships. These datasets are similar to \corpusname in that they are also argumentative in nature; however, they do not require participants to be persuasive and are grounded in personal opinions rather than beliefs. The \opendomain dataset comprises open-ended factual statements, where participants wrote seven truthful and seven deceptive sentences on any topic of their choice, provided the content remained plausible, reflecting everyday lying behavior. Finally, the \intn dataset consists of autobiographical narratives about the most important non-work-related activity participants planned for the upcoming week. Depending on the assigned condition, participants described this activity truthfully or deceptively. This dataset captures the role of personal involvement and narrative structure in deceptive expression.

Table~\ref{tab:defabel} summarizes descriptive statistics for all datasets. To assess their suitability for comparative analysis, we compute pairwise corpus similarity using the method proposed by \citet{LI2022103377}, which compares unigram and character trigram distributions via Spearman's $\rho$\footnote{Implementation available at: \url{https://github.com/jonathandunn/corpus\_similarity}}. As shown in Figure~\ref{fig:corpussimilarity}, the opinion-based datasets \decop and \crosscult show the highest similarity ($\rho = 0.83$). \defabeleng exhibits moderate similarity with the other datasets ($\rho = 0.58\text{–}0.68$), supporting its inclusion in comparative experiments. Overall, these values suggest that the datasets share significant overlap in term and character distributions, thereby validating their suitability for comparative analysis.

By comparing these datasets, we aim to gain a better understanding of the factors influencing deception detection performance and determine whether belief-based deception exhibits fundamentally weaker deception cues than other forms of deceptive text.  

All datasets considered here represent relatively low-stakes forms of deception, where participants have no personal or material consequences for being deceptive. This focus reflects both the nature of available corpora and the types of contexts most relevant to NLP applications, such as online discussions, reviews, or everyday persuasive writing. High-stakes deception, such as in legal, forensic, or financial documents, differs fundamentally in motivation and consequence, making it unsuitable for direct comparison and rarely accessible for research. Our analysis therefore focuses on forms of written deception that is representative of how deception typically appears in everyday textual communication.

\subsection{Are Linguistic Cues Reliable Indicators of Deception?}\label{sec:corr_analysis}

To examine the reliability of linguistic cues commonly associated with deception and to establish a foundation for assessing their predictive utility, we conduct a correlation analysis. We aim to quantify the strength and significance of relationships between individual linguistic cues and deception labels. 

\subsubsection{Experimental Setup}

We identify linguistic indicators of deception based on prior literature. Guided by the linguistic categories outlined in Table~\ref{tab:cues}, individual features representing each category are derived using the Linguistic Inquiry and Word Count (LIWC) lexicon (e.g., affective expressions, cognitive processes, self-references), readability measures (e.g., Flesch-Kincaid and Gunning-Fog indices), and other statistical metrics, including word count, lexical diversity, and part-of-speech frequency. These linguistic cues are systematically computed for each text using validated linguistic feature extraction tools (like LIWC) and standard Python libraries. A detailed description of the complete feature set is available in the Appendix B Table~\ref{tab:linguistic_cues}.

For LIWC lexicon-based measures, we used LIWC-22 for English, and the DE-LIWC2015 for texts in German. The feature labels differ slightly between the two lexicons, with the German lexicon containing fewer features compared to the English version. Additional linguistic features are computed using Python-based text-analysis libraries, including \texttt{spaCy}, \texttt{germansentiment}, \texttt{vaderSentiment}, \texttt{textstat}, \texttt{NLTK}, and \texttt{pyphen}. Code and resources for generating these linguistic features are openly accessible via our Bitbucket repository.\footnote{\url{https://bitbucket.org/aswathyve/linguistic\_style\_features/}}

\begin{figure}
	\centering
	\includegraphics[width=1\linewidth]{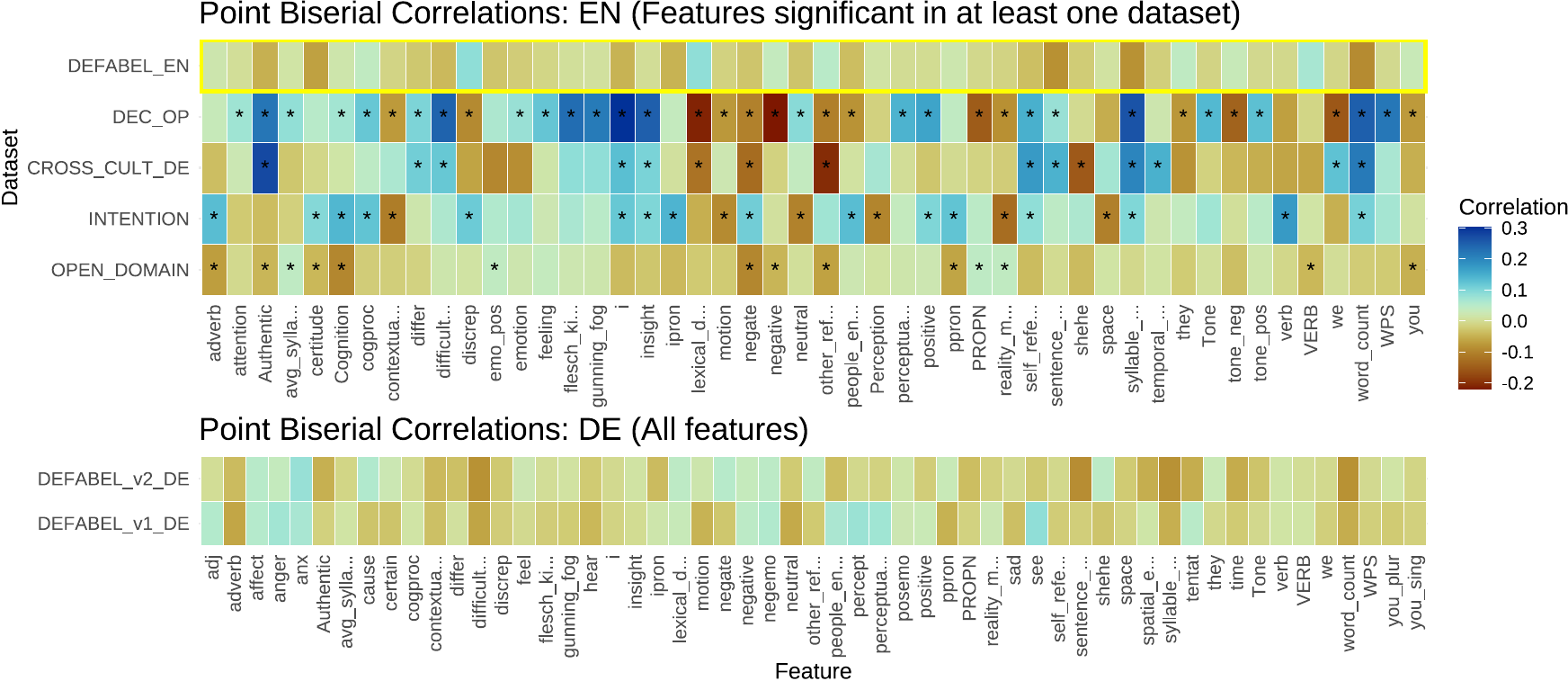}
	\caption{Point biserial correlations between linguistic features and deception labels across English (top) and German (bottom) datasets.}
	\label{fig:combinedplot}
\end{figure}

\paragraph{\textbf{Correlation Analysis}}

To identify linguistic features correlated with deception, we compute point-biserial correlations\citep{glass1996statistical} between each linguistic feature and the binary deception labels for each dataset. The features in all datasets are standardized using z-score normalization ($\mu = 0$, $\sigma = 1$) to ensure comparability across datasets. Features that lack sufficient variance or did not meet statistical assumptions (e.g., constant predictors or insufficient observations) are excluded from correlation computations. To control for multiple hypothesis testing, Bonferroni correction \cite{bonferroni1936teoria} is applied separately within each dataset. Correlations with an adjusted p-value below $0.05$ are considered statistically significant.

We conduct the analysis using the statistical programming language R \citep{R-lang}, employing packages, \texttt{dplyr} for data manipulation, \texttt{purrr} for functional programming, and \texttt{stringr} for efficient text processing.

\subsubsection{Results}

The results of the correlation analysis are visualized as heatmaps in Figure~\ref{fig:combinedplot}, showing the magnitudes and directions of correlations across datasets and linguistic features. Statistically significant correlations are marked with an asterisk (*).

A clear pattern can be observed on comparing datasets in the plot: the \corpusname datasets in both English and German display predominantly weak correlations, with none of the features showing any statistical significance. This finding aligns with our prior observations reported in \citet{velutharambath-etal-2024-entangled}, where similar analysis on the \defabelone dataset also shows an absence of significant linguistic cues. Across all datasets, correlation coefficients ranges approximately between 0.3 and -0.2, indicating relatively weak overall relationships. Nevertheless, other English datasets exhibit weak yet statistically significant correlations for some features.

The DEC OP dataset exhibits comparatively stronger correlations consistent with prior research. Specifically, categories such as authenticity (\textit{authentic}), self-reference (\textit{I}, \textit{we}), and text quantity (\textit{word\_count}, \textit{WPS}, \textit{syllable\_count}) correlates positively with truthfulness, while negative sentiment (\textit{negative}) and references to others (\textit{other\_reference}, \textit{they}) are indicative of deception. Similar trends emerge in the \crosscult dataset, likely reflecting its methodological similarity to the \decop dataset.

Contrary to the expectations based on prior research, \textit{lexical diversity} correlates positively with deception rather than truthfulness in most datasets (\decop, \crosscult, \intn), although notably not in the \corpusname datasets. While authenticity feature (\textit{authentic}) which is usually an indicator of truthfulness, does so in opinion-based datasets (\decop, \crosscult), but correlates with deception in open domain and personal narratives (\opendomain, \intn resp.) datasets. The \opendomain dataset shows generally weak correlations, comparable to the \corpusname datasets, yet some correlations still reach statistical significance. 

Overall, the correlation analysis questions the reliability of linguistic indicators of deception and emphasizes the strong influence of data collection methodology on the strength of linguistic cues. Datasets collected using similar protocols (\decop and \crosscult) exhibited comparable correlation patterns. In contrast, the absence of statistically significant linguistic indicators within the \corpusname datasets---despite efforts to enhance ecological validity through a revised data collection procedure---suggests that weaker cues result from the belief-based deception framework rather than language-specific factors.

\subsection{How Accurately Can Models Detect Deception?}

To systematically evaluate the predictive value of linguistic features in deception detection, we conduct experiments using (1) traditional machine learning models trained on linguistic features, (2) transformer-based language models fine-tuned for classification, and (3) instruction-tuned large language models applied in a zero-shot setting. This setup allows for a comparison between models that rely on predefined linguistic cues, those that learn representations from raw text, and models that follow task instructions without additional supervision.

\subsubsection{Experimental Setup}

\paragraph{\textbf{Datasets and Splits}}  
Our experiments are conducted on the selected set of deception datasets discussed in Section \ref{sec: comp_data}, which includes \defabelone and \defabeltwo in German, and \defabeleng, \crosscult, \decop, \opendomain, \intn in English. For each of the datasets, we follow a consistent topic-disjoint data partitioning strategy, splitting the data into a training set for model development and a holdout set reserved for final evaluation. This ensures that no overlapping topics appear in both subsets, thereby evaluating models' generalization to unseen instances. See Appendix Table \ref{tab:dataset_composition} for distribution of labels across the different splits of data.

\paragraph{\textbf{Linguistic Feature-based Classification}}  
To assess the predictive value of linguistic cues as features, we train feature-based classifiers for the task of deception detection. As features we included all of them used for the correlation analysis. They span a range of theoretically motivated dimensions, including syntactic patterns, affective and cognitive expressions, lexical diversity, readability, and referential strategies (see Appendix B, Table~\ref{tab:linguistic_cues}). 

We implement four feature-based classifiers---Logistic Regression, Random Forest, Support Vector Classifier (SVC), and Gradient Boosting---chosen for their diversity in learning approaches, from simple linear models to complex ensemble methods and non-linear classifiers. This variety ensures reliable evaluation across different data patterns. All features undergo z-score normalization to ensure comparability across dimensions. To evaluate model robustness and reduce sensitivity to data splits, we conduct stratified 10-fold cross-validation, preserving class distributions. After cross-validation, each classifier is retrained on the entire training set and evaluated on the topic-disjoint holdout set to assess generalization beyond memorized content. We use the \texttt{scikit-learn} python library for this implementation.

\paragraph{\textbf{Transformer-based Text Classification}} 
To evaluate the effectiveness of transformer-based models in deception detection task, we
 fine-tune models from the BERT \cite{devlin-etal-2019-bert} family. For English datasets, we use \texttt{FacebookAI/roberta-large}, trained on a broad English corpus, while for German datasets, we utilize \texttt{TUM/GottBERT\_base\_best}, pretrained on high-quality German texts.

Transformer models are fine-tuned by adding a classification layer and optimizing cross-entropy loss using the AdamW optimizer with a fixed learning rate of $1e{-5}$. Early stopping based on validation loss with a patience of five epochs mitigates overfitting, a critical step given the limited size and domain-specific nature of our datasets. Training is conducted for up to eight epochs with a batch size of 8, reflecting computational constraints and available data.

Model performance is evaluated through 10-fold cross-validation on the development set. All transformer-based experiments are implemented using the \texttt{transformers} library.

\paragraph{\textbf{Baselines}} In addition to the feature-based classifiers, we include two standard baselines to establish meaningful lower bounds for model performance. The majority class baseline is designed to always predict the most frequent label in the training data. While most datasets in our benchmark are balanced, \defabelone exhibits a notable skew toward the deceptive class, so as baseline we predicts all instances to be ``deceptive''. The random baseline, by contrast, assigns labels uniformly at random. These baselines serve as critical reference points, enabling us to assess whether models are learning informative patterns or merely capitalizing on label distribution.

\paragraph{\textbf{Prompting with LLMs}}\label{sec:llm}

Given the substantial computational and financial costs of running LLM inference on a large dataset ($\approx$ 15k texts) across multiple prompt configurations and models, we employ a multi-stage evaluation strategy. In the first stage, we assess model performance using a simple zero-shot prompt with smaller models, evaluating results on both the training and holdout sets to ensure comparability with traditional classifiers. To examine the effect of prompt variation, we then test a set of prompt variants, this time restricted to the holdout dataset, to capture differences in instruction phrasing without excessive computation. This approach leverages the fact that instruction-tuned LLMs can be evaluated directly without additional task-specific training. Finally, to investigate the influence of model size, we conduct additional experiments on the holdout dataset using models of varying scales.

\noindent(1) \textbf{Simple Prompt Set-up:} We use three open-source instruction-tuned models: \texttt{Mistral-7B-Instruct}\footnote{\url{https://huggingface.co/mistralai/mistral-Instruct-v0.3}}, \texttt{Llama-3.1-8B}\footnote{\url{https://huggingface.co/meta-llama/Llama-3.1-8B}} and \texttt{Qwen1.5-14B-Chat}\footnote{\url{https://huggingface.co/Qwen/Qwen2.5-14B-Instruct}}. \texttt{Mistral-7B-Instruct} is a lightweight model known for strong instruction-following capabilities and competitive multilingual performance, including German~\cite{jiang2024mixtral}. \texttt{Llama-3.1-8B} complements Mistral in size but differs in model design and training setup, and has been reported to perform well in generalization and reasoning, particularly in zero-shot tasks~\cite{grattafiori2024llama3herdmodels}. \texttt{Qwen2.5-14B-Instruct}, by contrast, is a larger model with enhanced multilingual coverage and improved reasoning abilities, particularly in complex language understanding tasks~\cite{qwen}. Additionally, we include \texttt{\gptmini}\footnote{\url{https://platform.openai.com/docs/models/\gptmini}}, a proprietary model, to provide a comparative perspective against open-source alternatives.

At this stage, our aim is not to identify the most effective or best-performing prompt for deception detection, but rather to explore how instruction-tuned language models interpret deception-related tasks when prompted in a zero-shot setting. We use the following zero-shot prompt for evaluating all models:

\begin{quote}
	\textbf{System:} \textit{You are an expert AI assistant who can detect truthful and deceptive intention in text very accurately.}
	\vspace{.2cm}
	
	\noindent\textbf{User:} \hspace{.3cm}    \textit{Evaluate whether the author of the provided text is being deceptive. Consider the author to be deceptive if they intentionally state something they know to be untrue or deliberately misrepresent their thoughts, feelings, intentions, or actions. Take into account that the author could hold false beliefs and hence intention is more important than factual accuracy.}
	
	\noindent\textit{Provide your judgment strictly in English in the following JSON format:}
	
	\noindent\textit{{\{
			``is\_lying'': true/false, ``reason\_title'': Very brief summary (as a short phrase) of why you believe the author is or isn't lying, ``explanation'': Detailed explanation supporting your judgment	\}}}
	
	\noindent\textit{Text to evaluate:} \texttt{<input\_instance>}
\end{quote}

In the prompt, we explicitly instruct the models to assess whether the author of a given text is lying. The instruction defines deception as the intentional misrepresentation of truth. It also clarifies that the author may hold false beliefs, and hence, the model is asked to prioritize the author's intent over factual accuracy. Additionally, each model is directed to produce (1) a brief phrase summarizing the reason for its decision (the \textit{reason title}) and (2) a detailed explanation outlining its rationale. The goal here is to observe LLM behavior under a simplified prompt setting without any prompt optimization efforts. 

\smallskip
\noindent(2) \textbf{Prompt Variations:} We design a set of prompt variants to explore how different prompting strategies influence model behavior and reasoning patterns in deception detection:

\begin{compactitem}
	\item \textbf{Zero-shot:} This variant provides more detailed guidance on the reasoning process, instructing the model to assess deception by considering factors such as internal contradictions, strategic framing, emotional consistency, and contextual plausibility.	
	
	\item \textbf{Random few-shot:} In this setting, two truthful and two deceptive examples are randomly selected and included to illustrate each class, offering the model minimal exposure to task-relevant examples before inference.
	
	\item \textbf{Feature-similarity few-shot:} Here, examples are chosen based on linguistic feature similarity to the input text. This approach allows the model to reference samples with comparable stylistic and structural characteristics, potentially improving contextual alignment.
	
	\item \textbf{Semantic-similarity few-shot:} In this variant, examples are selected using cosine similarity between sentence embeddings derived from a multilingual sentence transformer model\footnote{\texttt{sentence-transformers/paraphrase-multilingual-MiniLM-L12-v2}}. This method ensures that the model observes examples that are semantically closest to the target text.
	
	\item \textbf{Chain-of-thought (CoT) reasoning:} Each of the above prompts is further extended with a CoT version, where the model is explicitly instructed to reason through the text before producing its final judgment. The CoT prompts guide the model to first reflect on the content, then articulate its reasoning, and finally provide the deception label.
\end{compactitem}

In total, eight prompt configurations are evaluated in this stage (See Appendix \ref{sec:prompt-design}. To maintain computational efficiency, we select the two best-performing open-source models from the initial evaluation, along with the proprietary model, for the prompt variation experiments.

\smallskip
\noindent\textbf{(3) Model Size Variations.} To further investigate the impact of model scale while maintaining computational efficiency, we evaluate the best-performing prompt on larger variants of the top-performing open-source and proprietary models. Specifically, we consider \texttt{Mistral-8x7B-Instruct}\footnote{\url{https://huggingface.co/mistralai/Mixtral-8x7B-Instruct-v0.1}}, \texttt{Llama3.3-70B}\footnote{\url{https://console.groq.com/docs/model/llama-3.3-70b-versatile}}, and \texttt{GPT-4o}\footnote{\url{https://platform.openai.com/docs/models/gpt-4o}} as the larger counterparts of their respective architectures. This analysis aims to determine whether the models' deception detection capabilities scale with increasing parameter counts.

\begin{figure}
	\centering
	\includegraphics[width=1\linewidth]{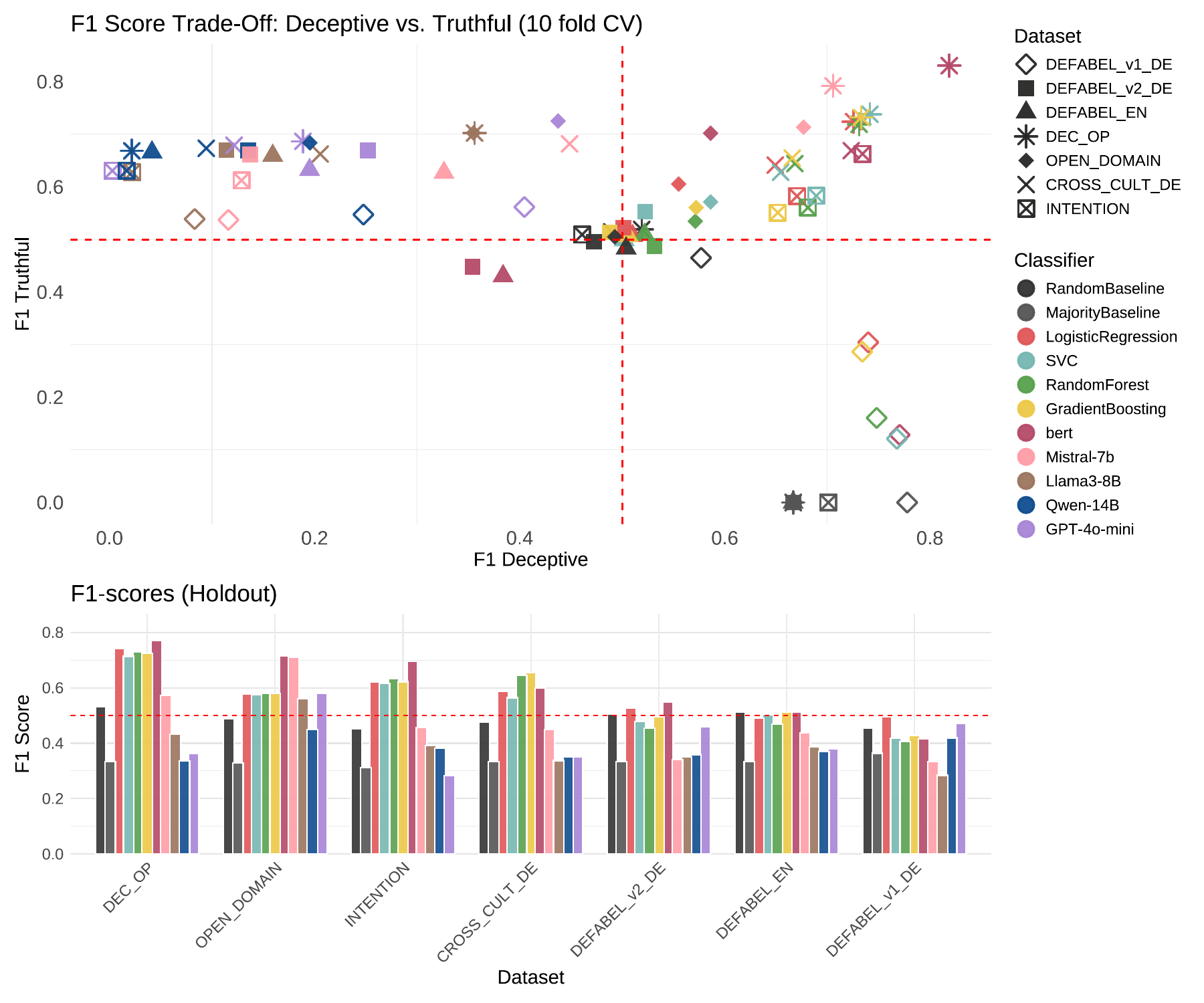}
	\caption{\(F_1\) scores across datasets and models. Top: trade-off between \textit{deceptive} and \textit{truthful} \(F_1\) scores (10-fold cross-validation on development set). Bottom: overall macro-\(F_1\) scores on holdout sets. The dotted red lines are included for visual reference, indicating that the macro-\(F_1\) scores  for the random baselines are centered around $0.50$.}
	\label{fig:f1tradeofftrain}
\end{figure}

\subsubsection{Results}

\paragraph{\textbf{(a) Comparison across classifiers}}

\smallskip
Figure~\ref{fig:f1tradeofftrain} shows the trade-off between \(F_1\) scores for the deceptive and truthful classes (top), evaluated via 10-fold cross-validation across different datasets and models. The LLM results correspond to the \textit{simple prompt set-up}. The bottom panel of the figure shows the corresponding model performance on the holdout sets, which consist of unseen instances. These holdout sets are used strictly for evaluation purposes and not for model optimization; all model parameters are fixed prior to this evaluation. Unless stated otherwise, all reported scores are macro-averaged \(F_1\) scores. See model performances across all datasets in the Appendix: Tables~\ref{tab:model_results_de_train}, \ref{tab:model_results_en_train}, \ref{tab:model_results_de_holdout} and \ref{tab:model_results_en_holdout}.

 \medskip
 \noindent\textbf{Detection of belief-based deception is challenging.} In Figure~\ref{fig:f1tradeofftrain},  we observe that among all the datasets, \mbox{\defabelone (\unfilleddiamond)} proves to be the most difficult for the models to classify accurately. Most models tend to predict a larger proportion of instances as deceptive, resulting in relatively high \(F_1\) scores for the deceptive class. This pattern closely resembles the majority class baseline, which predicts all instances as deceptive. Especially for the fine-tuned BERT model, this behavior is consistent across folds, suggesting a systematic modeling issue rather than random variation. On the holdout set, model performance on \defabelone drops substantially, with macro \(F_1\) scores falling below those of the random baseline for all models. We further observe that both \defabeltwo (~$\blacksquare$~) and \defabeleng (~$\blacktriangle$~) achieve model performance close to the random baseline, with the fine-tuned BERT model underperforming compared to the feature-based classifiers. In comparison to the \defabelone, the BERT model performs better on labeling truthful instances. This observation suggests that accounting for belief change during argumentation could have improved the data quality or validity. However, detecting deception embedded within otherwise truthful content remains particularly difficult even for finetuned models with more contextual information. One plausible explanation for BERT's behavior is its pretraining on large-scale corpora, which may bias the model toward associating plausible language with truthful intent---even when communicative intention is deceptive.

\medskip
 \noindent\textbf{Within dataset regularities results in better performance.} All English datasets, except \defabeleng, exhibit a similar pattern: feature-based models achieve moderate to strong classification performance, and fine-tuning BERT consistently improves results. This suggests that models are able to leverage consistent linguistic patterns in these datasets, enabling more reliable differentiation between deceptive and truthful instances. The performance of models on \decop dataset is particularly noteworthy. On both development and holdout data, all models (except LLMs) achieve relatively high \(F_1\) scores ($0.7\text{–}0.8$), with BERT achieving close to $0.8$ for both the deceptive and truthful classes. While BERT outperforms the feature-based models on this dataset, the feature-based models themselves still achieve strong results, markedly better than their performance on the other datasets. This consistent success across model types is especially notable given the inherently subtle nature of deception cues in textual context. The results suggest that the \decop dataset may contain linguistic or stylistic patterns that align closely with class labels---patterns that are more easily detectable than those in other corpora. This interpretation is supported by our linguistic feature correlation analysis, which revealed subtle yet statistically significant associations between deception labels and multiple linguistic cues. Together, these findings suggest that the observed separability in \decop may result more from annotation consistency or structural regularities than from generalizable deception indicators.

The results raise important questions about what the models are truly learning to detect. While performance appears strong on certain datasets, we assume that models may be exploiting dataset-specific regularities, such as stylistic or structural patterns, rather than genuinely identifying deceptive intent. A consistent pattern across our analysis is that deception becomes much harder to detect when it is embedded within factually accurate context. This is particularly evident in the more challenging \corpusname datasets, where deceptive intent does not manifest through clear or consistent linguistic markers. Instead, it is often subtle, context-dependent, and inseparable from truthful language. This exposes a core limitation of current models: their success appears to depend on the presence of stable surface-level patterns that coincide with deception labels. When such patterns are absent or weak, the models fail to distinguish deceptive intent, raising concerns about what these systems are actually learning to recognize.

\begin{figure}
	\centering
	\includegraphics[width=0.9\linewidth]{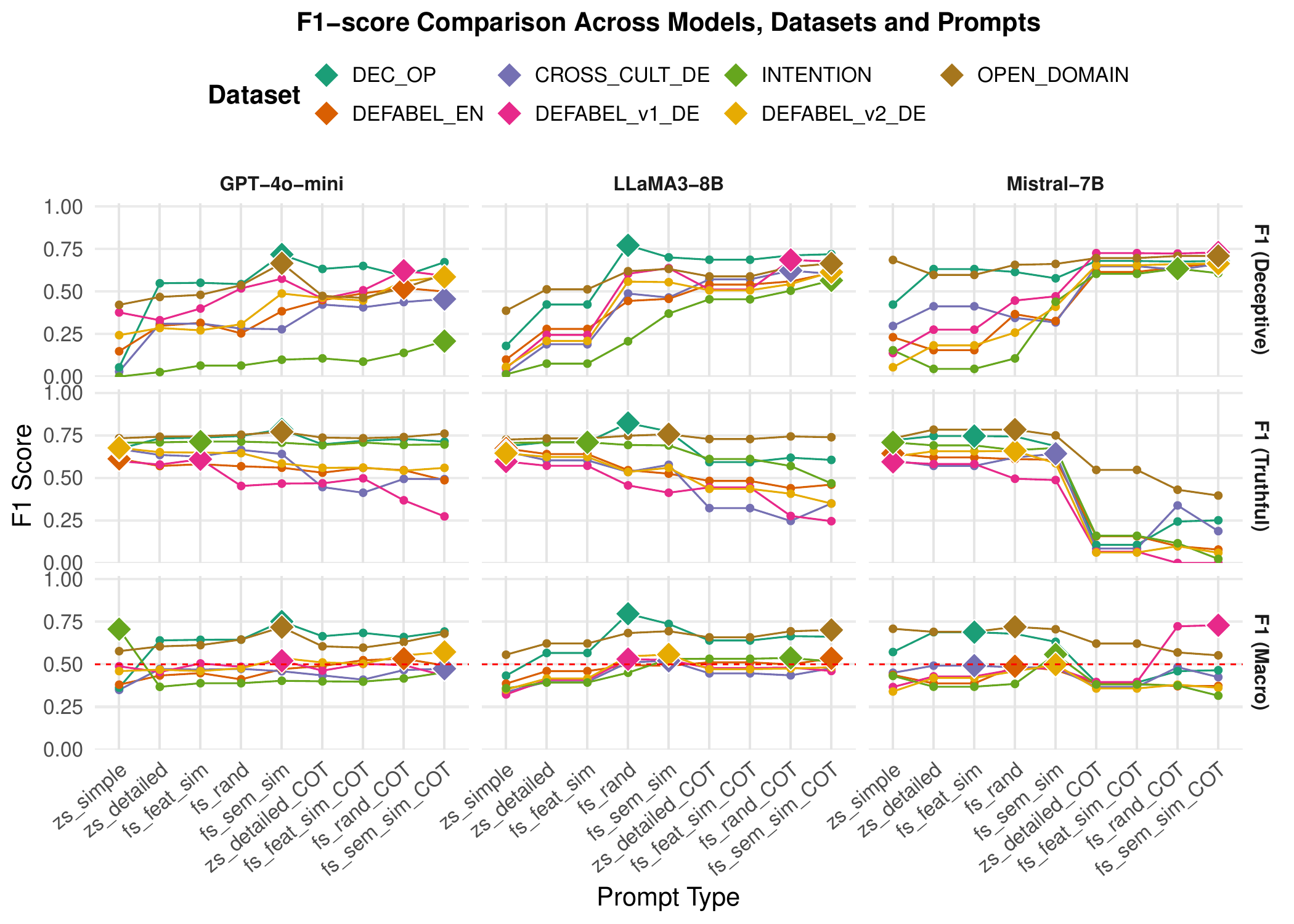}
	\caption{\F-score comparison across prompt variants for models: GPT-4o-mini, \llama and Mistral-7B, evaluated on seven datasets. Circular markers indicate individual prompt variants, while diamond markers highlight the best-performing prompt per model--dataset--metric combination. The red dashed line denotes the 0.5 macro-\F chance baseline.}
	
	\label{fig:promptcomparisonplot}
\end{figure}

\medskip
 \noindent\textbf{LLMs show a pronounced truth bias with simple prompts.} When analyzing the performance of the LLM-based classifiers in the \textit{simple prompt setting}, we observe that all models, Mistral-7B, \llama, \qwen, and GPT-4o-mini—generally perform worse than both the feature-based and fine-tuned transformer models across most datasets. An exception is \mistral, which performs comparatively well on the \decop and \opendomain datasets. However, in the \corpusname datasets, all instruction-tuned models exhibit a consistent and pronounced tendency to label instances as truthful, regardless of the actual author intent. This supports our hypothesis that instruction-tuned LLMs, optimized for alignment and helpfulness, are not well-suited to deception detection tasks. Rather than critically evaluating authorial intent, these models may rely heavily on factual plausibility, leading to a systematic bias toward truthful classifications. This tendency likely limits their reliability in detecting deceptive intent and is clearly reflected in the rationales provided by the models for their predictions (see Section~\ref{sec: llm_error}).

\paragraph{\textbf{(b) Comparison across prompt variants}}
Among the open-source LLMs, \qwen consistently exhibits the weakest performance across datasets. Therefore, the prompt variation experiment includes \llama and \mistral as the representative open-source models, alongside \gptmini as the proprietary model. Figure \ref{fig:promptcomparisonplot} presents a comparison of \F scores---deceptive, truthful, and overall---across different prompt variants for the three models and seven datasets. See Tables \ref{tab:stage2-results-gpt-4o-mini}, \ref{tab:stage2-results-llama3-8b}, and \ref{tab:stage2-results-mistral-7b} for \F scores, precision, and recall values across prompt variants for each dataset.

\begin{figure}
	\centering
	\includegraphics[width=0.9\linewidth]{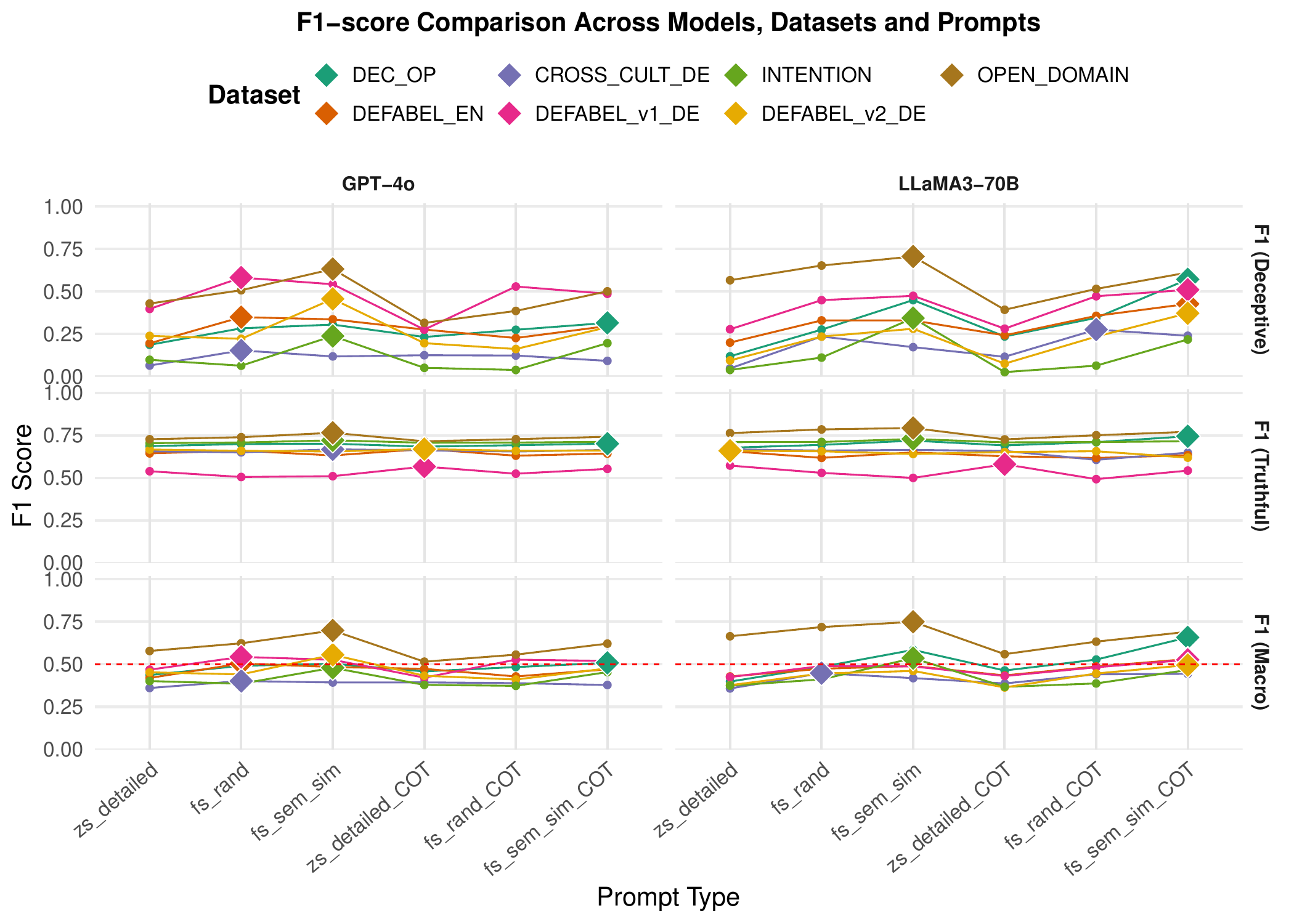}
	\caption{\F-score comparison across prompt variants for larger language models: \mbox{GPT-4o} and \mbox{LLaMA3-70B}, evaluated on seven datasets. Circular markers indicate individual prompt variants, while diamond markers highlight the best-performing prompt per model--dataset--metric combination. The red dashed line denotes the 0.5 macro-\F chance baseline.}
	\label{fig:modelcomparisonplotlarge}
\end{figure}

\medskip
\noindent{\textbf{LLMs exhibit variable bias patterns across prompt variants.}} When comparing the simple zero-shot prompt (\textit{zs\_simple}) with the more detailed version (\textit{zs\_detailed}), we observe that the latter generally achieves higher overall \F scores across models and datasets, with only a few exceptions. For example, the best performance on \intn is observed with the simple prompt, and on the \corpusname dataset performance either remains the same or worsens for \gptmini. Adding more examples to the detailed prompt does not substantially improve model performance overall. However, for \decop and \opendomain, we find that incorporating semantically similar instances as few-shot examples (\textit{fs\_sem\_sim}) leads to a notable boost in performance (\F $\geq$ .75 for \gptmini and \llama, and $\geq$ .7 across all models), approaching the high scores reported for fine-tuned BERT models. This finding reinforces the hypothesis that model performance is driven in part by dataset-level regularities, given that deception cues are often embedded in stylistic rather than semantic features.

With the simple prompt, we observe that LLMs tend to exhibit a truth bias. However, this trend reverses when chain-of-thought (CoT) reasoning is introduced into the prompts. A consistent pattern in Figure {fig:promptcomparisonplot} shows that \F scores for the deceptive label increase, while those for the truthful label decrease under CoT-based prompts compared to non-CoT prompts. This suggests that CoT prompting induces a bias toward labeling more instances as deceptive—a pattern particularly pronounced in \mistral compared to the other models.

These patterns are also evident in the \corpusname datasets. However, the best \F scores in this case still range only between .53 and .57, with the exception of a CoT prompt (\F = .73), which is largely an artifact of predicting all instances as deceptive. Overall, these observations suggest that including more information in the prompts generally improves model performance, yet no single prompting strategy yields consistent improvements across models or datasets. CoT prompting tends to bias models toward predicting instances as deceptive. Furthermore, the overall \F scores indicate that LLMs still struggle to achieve performance substantially above a random baseline. This remains true for all datasets except for \decop and \opendomain.

\paragraph{\textbf{(c) Comparison across model sizes}}
Building on the insights from the previous experiments, we next examine how model size influences deception detection performance. For this analysis, we focus on the LLaMA and GPT model families to assess whether larger architectures are better able to capture stylistic cues indicative of deception. We evaluate the prompts \textit{zs\_detailed}, \textit{fs\_random}, and \textit{fs\_sem\_sim}, along with their CoT variants, as these configurations showed improved performance across datasets and models, albeit not consistently.

From Figure \ref{fig:modelcomparisonplotlarge}, we observe that increasing model size does not lead to a substantial improvement in overall \F scores. Interestingly, in larger models, the truth bias observed with the simple prompt re-emerges across all prompt types. The deceptive-label bias observed for CoT prompting in smaller models is largely absent. These results suggest that scaling up model size mitigates some of the prompt-specific biases observed in smaller architectures but does not yield notable gains in deception detection accuracy.

\paragraph{\textbf{(d) Cross-linguistic comparison on \corpusname datasets}}
To assess whether the observed bias patterns generalize across languages, we compare model performance on the English (\defabeleng) and German (\defabelone, \defabeltwo) variants of the \corpusname datasets. Overall, the results show very similar performance trends across languages, with differences primarily in magnitude rather than direction. In both English and German, the bias patterns observed for prompts and model sizes also hold. While the general behavioral patterns are comparable across languages, the German datasets—especially \defabelone—exhibit larger bias swings and greater performance variability across prompt types. CoT prompts, in particular, often produce much higher false-positive rates in German compared to the English dataset. However, these differences are relatively minor and may stem from the predominance of English in model pretraining and instruction-tuning data, as well as from structural and stylistic differences in how deceptive cues are expressed in German. Based on these findings, we argue that the observations made on belief-based deception are largely language-independent.

\subsection{How Do LLMs Justify Misclassifications?}\label{sec: llm_error}

Deception detection experiments with instruction-tuned models revealed that variations in prompts and model size do not lead to consistent improvements in overall performance. In the zero-shot simple prompt, the models were not provided with explicit cues about deception-detection strategies such as internal contradiction, strategic framing, emotional consistency, or contextual plausibility. This setup offers a baseline for examining how models justify their predictions. 

In this section, we examine the internal reasoning patterns that underlie incorrect predictions made by large language models in the context of deception detection. The analysis focuses on the explanations that LLMs generate when they misclassify inputs, specifically, when deceptive content is labeled as truthful or vice versa. By systematically analyzing these justifications, we aim to uncover recurring biases in model reasoning and assess whether they correspond to linguistic or pragmatic cues commonly associated with deceptive language.

\subsubsection{Experimental Setup}

As discussed in Section~\ref{sec:llm}, our LLM-based classification experiments involve prompting instruction-tuned models to perform deception detection. Given an input text, each model is asked to output (1) a binary label indicating whether the author is lying or telling the truth, (2) a short justification summarizing the rationale for its decision (referred to as the \textit{reason title}), and (3) a longer, more detailed explanation of the decision process. 

\begin{figure}
	\centering
	\includegraphics[width=\linewidth]{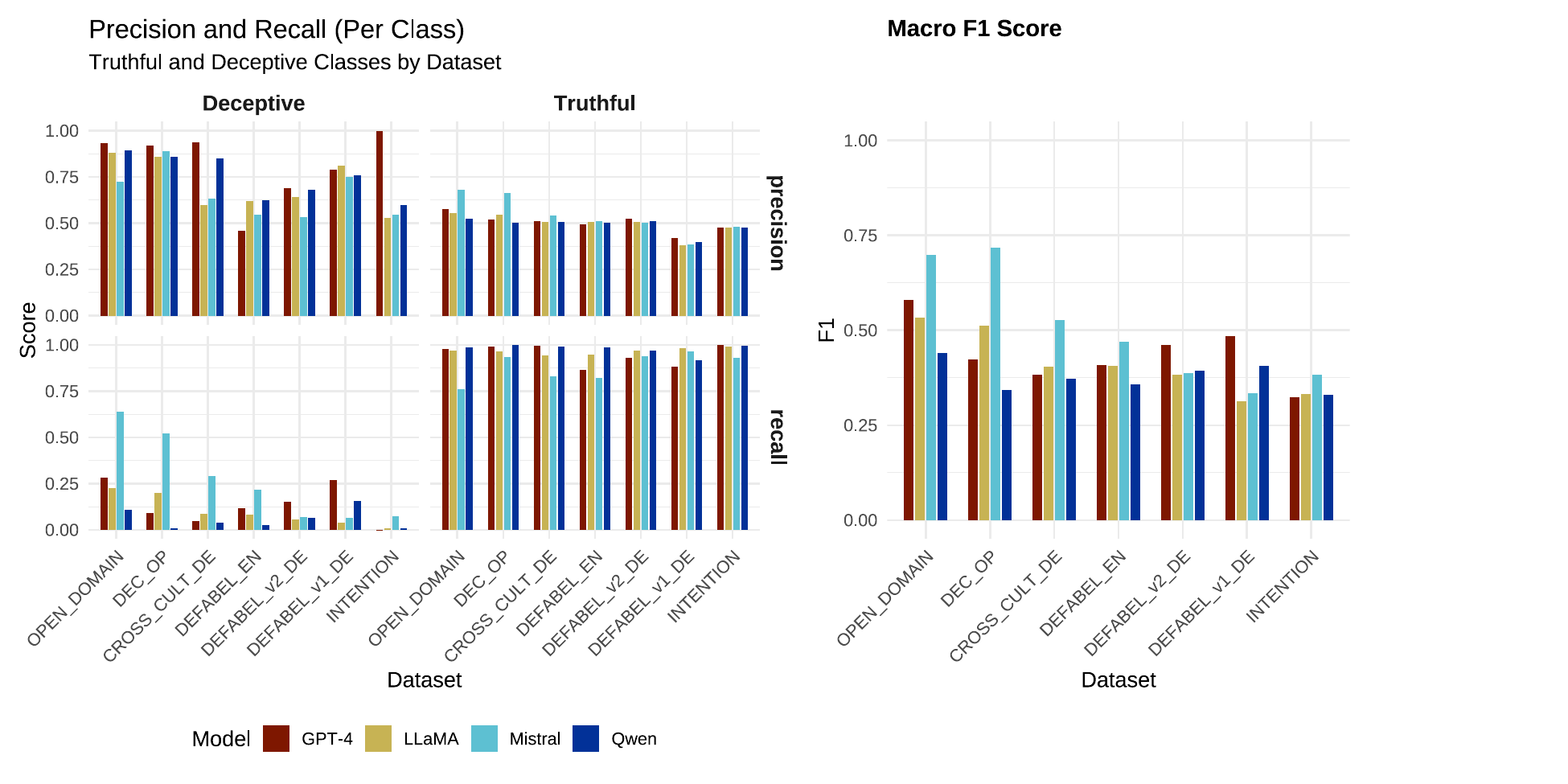}
	\caption{(Right) Precision vs. Recall comparison for Truthful and Deceptive labels across four LLMs, Mistral-7B, Llama-8B, Qwen-14B and GPT-4o-mini (Left) macro \(F_1\) scores for all LLMs.}
	\label{fig:error}
\end{figure}
Figure~\ref{fig:error} shows the precision, recall and macro \(F_1\) for each LLM evaluated on the full dataset. On reviewing the results, we observe a consistent pattern across all LLMs: a strong bias toward predicting inputs as truthful. We see that recall for the truthful class is consistently high, indicating that the models correctly identify most truthful instances. However, this comes at the cost of lower precision for the same class, suggesting that many deceptive inputs are incorrectly labeled as truthful. In contrast, for the deceptive class, precision is relatively high---when the model does predict deception, it is often correct---but recall is substantially lower, meaning the majority of deceptive instances are not detected. This imbalance reveals a systematic tendency to favor the truthful label unless there are clear and strong cues suggesting deception. To better understand the models' internal reasoning processes and sources of error, we conduct a detailed analysis of the misclassified instances. 

Although our experiments include multiple large language models, we center this error analysis on the outputs of the Mistral model as it consistently achieves stronger performance across datasets, both in terms of macro-averaged \(F_1\) score and recall for deceptive instances. These properties make it the most informative model for analyzing misclassification behavior, particularly in examining how deceptive content is interpreted and justified. We focus our analysis on the \textit{reason titles}, treating them as concise, high-level representations of the models' decision rationales. 

Each model prediction is compared to the corresponding ground truth label, and the outcomes are categorized into four standard evaluation classes: true positives (TP), true negatives (TN), false positives (FP), and false negatives (FN). In our evaluation, as the task is deception detection, we treat the \textit{deceptive} class as the positive class. In this context, a false positive denotes a truthful input incorrectly classified as deceptive, while a false negative refers to a deceptive input wrongly labeled as truthful.

To investigate the reasoning patterns underlying model misclassifications, we focus on the \textit{reason titles} associated with false positives and false negatives, as these errors offer the most direct insight into model failures. Each reason title is embedded using the \texttt{all-MiniLM-L6-v2} model from the SentenceTransformers\footnote{\url{https://huggingface.co/sentence-transformers/all-MiniLM-L6-v2}} library, which maps the short justifications into dense semantic vectors in a high-dimensional space. We then apply KMeans clustering to group together semantically similar justifications within each error type. The number of clusters is determined dynamically for each subset, with an upper bound of 20 and a lower bound of 2, depending on the number of available samples. This allows us to maintain a balance between interpretability and resolution, while adapting to the size of each group. To visualize the resulting clusters, we reduce the embedding dimensions to two using t-SNE. The perplexity parameter for t-SNE is set adaptively based on subset size, capped at 30 and floored at 5, to ensure stability across datasets. The resulting two-dimensional coordinates are used to analyze and interpret the distribution of reasoning patterns across error types.

After clustering, we manually reviewed each group of reason titles within the false positive and false negative subsets. For each cluster, we examined representative examples and assigned a descriptive label summarizing its dominant reasoning pattern. In cases where multiple clusters reflected closely related themes, we merged them to avoid redundancy and improve interpretability. This qualitative analysis allowed us to identify recurring justification strategies used by the model when it misclassified deceptive or truthful instances.

\begin{figure}
	\centering
	\includegraphics[width=1\linewidth]{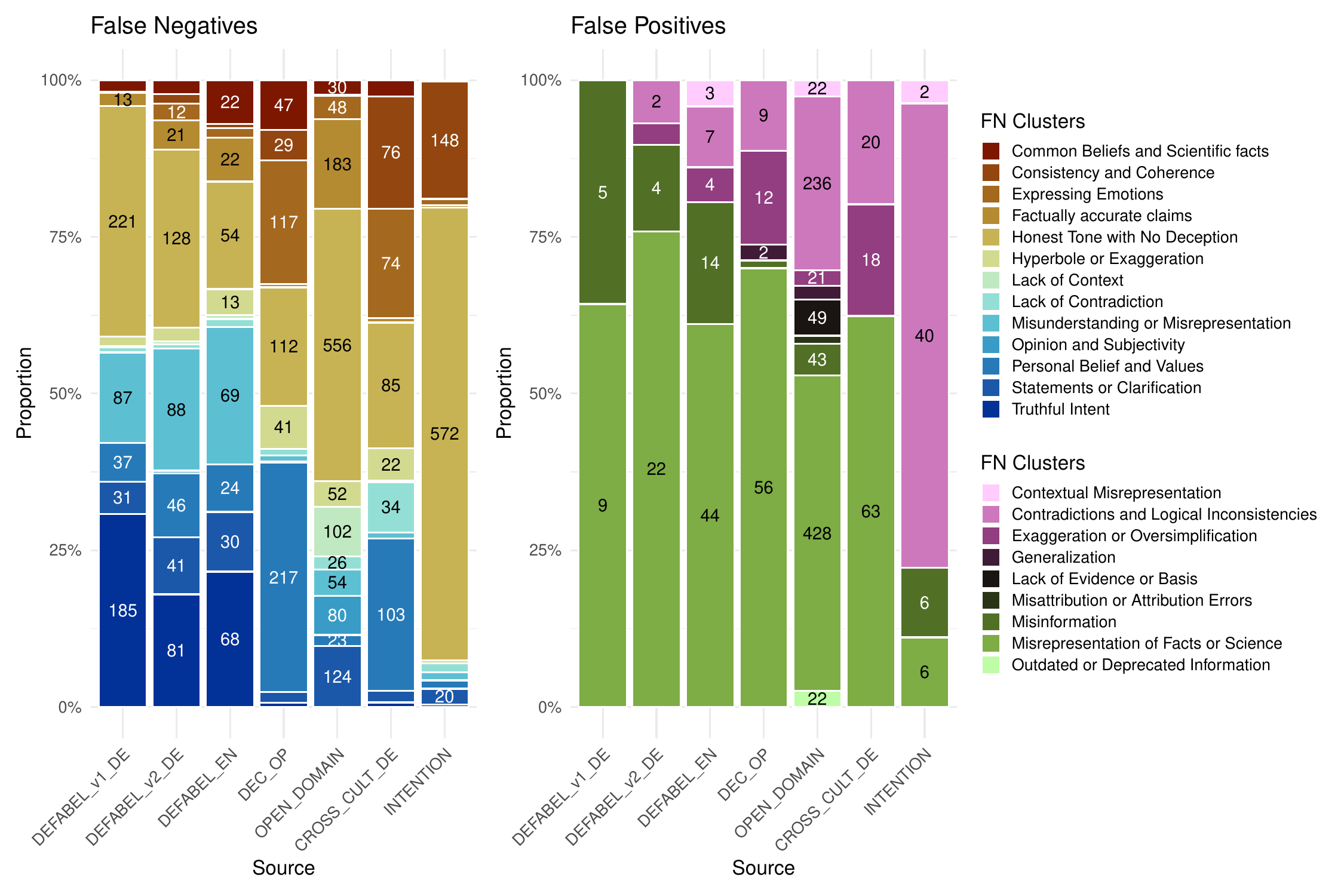}
	\caption{Distribution of explanation clusters among false positives and false negatives across datasets. Each bar shows the proportion of error instances attributed to different types of model reasoning.}
	\label{fig:clusterdistribution}
\end{figure}

\subsubsection{Results}

Figure~\ref{fig:clusterdistribution} shows the proportion of reasoning clusters associated with false positives and false negatives, as produced by the Mistral model across datasets. Each bar represents the relative share of explanation types within a given error category, allowing comparison of dominant model rationales across different sources (See Table \ref{tab:fp_fn_by_dataset} in the Appendix for absolute counts).

Among false negatives---deceptive instances wrongly predicted as truthful---the most common reasoning clusters include \textit{Honest Tone with No Deception}, \textit{Expressing Emotions}, and \textit{Consistency and Coherence}. These features have been associated with truthful communication in prior work, where positive tone and emotion, and coherent narratives are often correlated with honesty \cite{newman2003, vrij2008detecting}. As a result, deceptive messages that exhibit these characteristics are more likely to be misclassified as truthful by the model. 

Consistent with our hypothesis that LLMs with large-scale pretraining are biased toward factual accuracy, we observe a strong reliance on factual plausibility as a cue for truthfulness. This tendency is particularly evident in the \corpusname datasets and in \opendomain, where many deceptive instances are misclassified with justifications such as \textit{Factually Accurate Claims} or \textit{Misunderstandings or Misrepresentations}.
 These explanations indicate that the model focuses on factual correctness over the speaker's underlying intent, often interpreting statements as truthful when they are logically sound or factually aligned. 
 
In opinion- and belief-based datasets such as \decop, \crosscult, and \corpusname, statements expressing subjective viewpoints or personal convictions are frequently judged as truthful, with reason titles commonly assigned to the \textit{Personal Belief and Values} cluster. This suggests that the model treats the expression of belief itself as evidence of sincerity, independent of whether the belief is intended to mislead.
In the personal narrative dataset \intn, deceptive inputs are often justified as truthful based on the presence of a sincere tone, with many misclassifications falling under \textit{Honest Tone with No Deception}. This highlights the model's difficulty in detecting deception in subjective or experiential content, where factual verification is not possible. Interestingly, in \corpusname datasets, a notable number of false negatives are explained using the \textit{Truthful Intent} cluster, suggesting that the model attempts to assess communicative intent, as instructed in the prompts, but lacks a reliable basis for inferring it. 

Taken together, these findings suggest that the model equates emotional tone, factual plausibility, and personal expression with truthfulness. In doing so, it overgeneralizes from linguistic cues and fails to capture the deeper pragmatics of deceptive communication.

In the case of false positives---truthful instances incorrectly labeled as deceptive---the most common reasoning cluster is \textit{Misrepresentation of Facts or Science}. This supports our hypothesis that LLMs tend to treat factual accuracy as a proxy for honesty, using it as a primary basis for deception judgments. In \opendomain, a large proportion of false positives are attributed to \textit{Contradictions and Logical Inconsistencies}, with explanations citing inconsistencies in narrative, timeline, or detail. Such cues are often associated with deception in psychological literature, particularly in high-stakes scenarios, where the cognitive load of lying can make it difficult to maintain a coherent story. However, in this context, the model appears to misapply these heuristics, likely due to its exposure to generalized patterns of deceptive behavior during training. As a result, even truthful content that deviates from strict factual or narrative coherence may be incorrectly flagged as deceptive.

When comparing the German and English \corpusname datasets, we find that the distribution of reasoning for false negatives is quite similar in the belief-based deception datasets compared to other datasets. In the case of false positives, \defabeltwo and \defabeleng, which were collected using the same setup, also show comparable patterns. This suggests that the truth-bias behavior does not vary considerably across languages, but rather across datasets.

While the above patterns capture the main quantitative trends, a closer manual inspection of model explanations reveals additional qualitative insights. We find that Mistral rarely engages in explicit reasoning about deceptive intent and instead relies predominantly on judgments of factual accuracy. Even when a statement is factually incorrect, the model often classifies it as truthful if it appears plausible or confidently stated by the author. Table~\ref{tab:camel_fp_examples} presents a few illustrative examples from the \defabeleng dataset, all of which were labeled as truthful by the model. In the first two instances, the factual content aligns with the author's belief, whereas in the latter two it does not. Nevertheless, the model predicts all of them as truthful, providing rationales that emphasize factual accuracy without critical evaluation. These cases suggest that the model tends to echo or ``parrot'' the authors' claims rather than verify them, highlighting one of the many inconsistencies in its reasoning process.

Overall, the model appears to conflate factual correctness with honesty and often attempts to justify its predictions by referencing linguistic correlates of deceptive communication, even in cases where these features are not contextually warranted.

\begin{table}
	\centering
	\scriptsize
	\setlength{\tabcolsep}{3pt}
	\renewcommand{\arraystretch}{0.9}
       \begin{tabularx}{\textwidth}{X c m{2.5cm} X}
	
		\toprule
		\textbf{Argument} & \textbf{True Label} & \textbf{Reason Title} & \textbf{Rationale} \\
		\midrule
		{\sethlcolor{green!30}\hl{Camels store fat in their hump(s)}} as a source of energy. Where they live means that they may go without food for long periods and long distances. It is not possible for a camel to store water and it is a common misconception that humps contain water. &
		{\sethlcolor{green!30}\hl{truthful}} &
		No deception detected &
		The author is providing \hl{accurate information about camels}, stating that they store fat in their humps as a source of energy and clarifying that it is a common misconception that humps contain water. There is no indication that the author is intentionally stating something they know to be untrue or deliberately misrepresenting their thoughts, feelings, intentions, or actions. \\

\midrule
		{\sethlcolor{red!30}\hl{Camels have to store water in their humps}} because they travel long distances in the desert in very high temperatures and would die of thirst if they were unable to do this. Many people think that the humps contain fat but this is a common misconception! &
		{\sethlcolor{red!30}\hl{deceptive}} &
		No deception detected &
		The author is \hl{stating a fact about camels and their humps}, and there is no evidence to suggest that they are intentionally misrepresenting this information. While it is true that some people may hold the misconception that camel humps contain fat, the author is not one of them and is instead correcting this misconception. \\
\midrule

		{\sethlcolor{red!30}\hl{Camels store water in their humps}} to survive in the desert. While people often think their humps only store fat, the truth is the humps also play a key role in holding water. Camels are known for being able to survive without water for a long time, and this is possible because their humps store water to keep them hydrated. 
		Without the ability to store water in their humps, camels wouldn't be able to go for days without drinking, especially in harsh desert environments. The humps are a special adaptation that helps camels survive in places with little water. This shows that the humps are important for storing water, not just fat, and are crucial for camels living in the desert. &
		{\sethlcolor{green!30}\hl{truthful}} &
		The text accurately describes the function of camel humps &
		The author provides a \hl{clear and accurate explanation of the role of camel humps} in storing water, debunking the common misconception that they only store fat. The text also supports this claim with evidence, such as the camels ability to survive without water for extended periods and the importance of humps in desert survival. \\
\midrule

		{\sethlcolor{green!30}\hl{Camels don't store water in their humps}}. Their humps are full of fat, not water. They use this fat for energy when they can't find food.
		Camels are good at saving water, but they keep it in their blood, not their humps. They can drink a lot of water at once and stay hydrated for a long time.
		If humps stored water, they would get smaller when the camel got thirsty. But humps only shrink when the camel uses up its fat.
		 &
		{\sethlcolor{red!30}\hl{deceptive}} &
		The author is stating accurate information &
		The author is \hl{providing correct information about camels and their humps}. They explain that camel humps are not storage for water, but rather contain fat for energy, and that camels store water in their blood. The author also clarifies common misconceptions about camels storing water in their humps and how their humps shrink when they use up their fat, not when they get thirsty.\\
		\bottomrule
	\end{tabularx}
\caption{Representative examples from the \defabeleng dataset showing author arguments, ground-truth labels, and the model's explanatory rationales.}
	\label{tab:camel_fp_examples}
\end{table}

\section{Conclusion \& Future Work}

Can deception be detected purely from linguistic cues in text? This question has long guided research in computational linguistics, showing promising results across a variety of datasets. Previous work often relies on the assumption that deception creates cognitive strain, which in turn reveals itself through linguistic markers, suggesting that surface-level textual features can reliably signal deceptive intent. In this paper, we critically reassess these assumptions through the belief-based deception framework.

By introducing the \corpusname corpus---a multilingual dataset grounded in belief-based deception annotations---we present an alternative framework that isolates deceptive intent from factual correctness. This shift in operationalization allows us to examine deception as a misalignment between belief and expression, irrespective of factual accuracy. Across all variants of \corpusname, we find that linguistic cues traditionally associated with deception show no significant correlation with deception labels. In similar datasets, these cues exhibit weak and inconsistent correlations. Models that perform well on existing deception datasets, including feature-based classifiers, transformer models, and instruction-tuned large language models, consistently fail under our belief-based paradigm. Instruction-tuned models in particular display a pronounced truth bias, consistently classifying most statements as truthful regardless of the author's intent. A similar observation
has been made by \citet{chuang-etal-2024-simulating} showing that LLMs have a bias towards producing factually accurate information in line with scientific reality, despite being prompted to support an opposing stand. Although chain-of-thought prompting can partially reverse this tendency by increasing deceptive classifications, neither prompt variation nor model scaling effectively mitigates the bias. These findings suggest that prior successes in deception detection may stem more from dataset-specific artifacts than from any stable, generalizable signals of deception, which also invite broader reflection on the nature of deception in language and discourse.

\paragraph{\textbf{Open Question}} Why do deception cues vanish in the \textsc{DEFABEL} setting? We hypothesize that the argumentative nature of the task with explicitly tasks the participants to convince another person plays a critical role. Deceptive arguments, especially in low-stakes experimental settings, are not spontaneous lies but carefully constructed rhetorical acts. They do not necessarily impose heightened cognitive load, nor do they evoke emotional discomfort; instead, they may rely on the same linguistic strategies used in persuasive writing. As a result, deception in this context may not disrupt language, instead may refine it. This blurs the line between deception and persuasion: the liar and the debater may be linguistically indistinguishable, not because deception leaves no trace, but because argumentation offers tools to mask it. 

This interpretation also helps explain why our findings may not directly extend to higher-stakes or emotionally charged forms of deception. In such contexts, the psychological and social pressures surrounding truth and consequence could yield different linguistic dynamics, ones less constrained by rhetorical form and more influenced by affective or cognitive strain. 

Our findings thus call for a broader rethinking of deception detection \textit{solely from textual data}, in computational linguistics. Rather than treating deception as a fixed linguistic pattern, we propose viewing it as a context-sensitive, pragmatically constructed phenomenon, rooted in speaker intent and communicative setting. In doing so, we shift the analytical focus from linguistic features to epistemic alignment and communicative goals. A closely related perspective comes from \citet{li-santos-2012-argument}, one of the few studies to explicitly model deception in structured, argumentative settings. Their work simulates social interactions between agents, capturing both the internal reasoning behind deceptive arguments and the communicative strategies used to present them. Deception, in their model, emerges through the intentional construction of arguments designed to align with the expectations of others. Crucially, they find that deception leaves behind detectable patterns in belief manipulation and argument structure. Our findings present a contrast: in real-world rhetorical contexts, such as those represented in our corpus, these patterns may be masked rather than revealed. When deception is embedded in persuasive, fluent argumentation, it may not disrupt linguistic or structural norms at all. This divergence points to a key question for future work: under what conditions do belief-based signals of deception survive, and when are they erased by the demands of genre and discourse?

\paragraph{\textbf{What if deception cannot be detected?}} This question, posed in the title of our paper, serves as a final reflection. 
If deception, specifically in low-stakes contexts, cannot be reliably detected through linguistic cues alone, then the widespread deployment of automated deception detection systems must be approached with extreme caution. Our findings challenge the foundational assumption that textual features can capture the intent to deceive, especially when separated from factual accuracy. This casts doubt on the reliability and ethical defensibility of systems that claim to detect deception in online-communication.

More broadly, our results urge the community to reconsider the framing of deception detection as a solvable classification task. If deception is inherently contextual, strategic, and shaped by interpersonal and cultural dynamics, then purely text-based models, no matter how advanced, may always fall short. Future work must grapple with these limitations, exploring richer representations of belief, context, and communicative intent, or perhaps acknowledging the limits of what such systems can responsibly infer. 

\paragraph{\textbf{Future work}} Our findings point toward a deeper reconceptualization of deception in structured discourse. In argumentative settings, which reward coherence, fluency, and persuasiveness, deception may not manifest through disruption or leakage, but through conformity. Rather than breaking with linguistic norms, deceptive statements may align closely with the rhetorical expectations of the genre, effectively camouflaging intent behind structured language. This echoes Goffman's notion of performance in social interaction, where the speaker's presentation of self is always strategically constructed to fit the frame of the situation \citep{goffman1959presentation}. Similarly, in the context of mediated communication, Turkle \citep{turkle1995life} noted that text-based environments allow for the construction of idealized selves, where messages can be carefully ''\textit{edited}, \textit{deleted}, and \textit{retouched}``. Walther's hyperpersonal model \citep{walther1996computer} likewise emphasizes how low-cue, asynchronous settings enable users to curate their personas in ways that obscure intent. In these contexts, deception becomes less a linguistic anomaly and more a stylized act, indistinguishable from persuasive stance-taking. Taken together, these insights suggest that the absence of detectable cues in our study may not reflect a shortcoming of computational models, but rather the success of deceivers in adapting to the communicative affordances and rhetorical conventions of mediated, structured discourse.

This shift opens up several directions for future research. One is to investigate how deception behaves across different communicative genres---for instance, comparing structured argumentation with more spontaneous, less rhetorically constrained settings. Another is to examine whether belief-expression misalignment becomes more salient under higher-stakes or emotionally charged conditions. Finally, while our results suggest that linguistic cues are weak signals in rhetorical contexts, future work may explore whether deeper models of reasoning and audience design can recover latent traces of deceptive intent. Understanding when, where, and how deception becomes visible remains an open challenge that lies at the intersection of language, intent, and context.

\section*{Limitations}

This study investigates linguistic cues of deception in low-stakes scenarios, across German and English. While we analyze how previously reported cues behave on selected datasets and evaluate the performance of existing state-of-the-art models, our primary focus is not on optimizing model accuracy. Our aim is to understand how deception manifests linguistically rather than to build the most effective classifier. We acknowledge that tailored models could yield improved performance for specific usecases. However, our study intentionally prioritizes the linguistic exploration of deception over model refinement, leaving further optimization for future work. 

The null findings in our linguistic correlation experiments on the \corpusname datasets should be interpreted with appropriate caution. The absence of statistically significant correlations between linguistic features and deceptive intent may be due to the specific operationalization of the data collection process or the feature sets examined in this study, rather than the total absence of linguistic cues of deception. Further, deceptive language in argumentative contexts may differ from other forms of deception. However, we do not explore or contrast linguistic patterns of deception in argumentative contexts with those in other low-stakes scenarios, as this is beyond the scope of the current study. This is an area that warrants further exploration, particularly in understanding how deception patterns vary across different discourse genres and communicative goals.

Beyond methodological scope, deception detection research carries societal limitations. Linguistic cues are not universal, they vary with language, culture, and communication style. Models trained on specific datasets may therefore generalize poorly and risk biased or unfair outcomes, particularly for non-native speakers. These limitations highlight the need for fairness evaluation and transparency in future cross-linguistic work.
\section*{Ethical Considerations}
\label{sec:ethical}

This study involves the elicitation and analysis of deceptive arguments in a controlled, low-stakes setting. Participants were asked to argue against their own beliefs, with all prompts carefully curated to avoid sensitive or emotionally charged content. These prompts reflect common misconceptions rather than deeply held values, minimizing potential discomfort. 

All participants were informed of the nature and goals of the study in advance and gave explicit consent prior to taking part. They were also reminded of their right to withdraw at any time, without consequence. While we collected demographic information to support analysis, we did not collect or publish any personally identifying information. All data used in our analyses remains fully anonymized.

We are aware of the broader ethical implications of deception detection research. While our research aims to deepen the scientific understanding of deceptive intent through linguistic and computational analysis, such models could be misused, for example, to unfairly profile individuals or generate more convincing disinformation. 
 We also recognize that linguistic cues to deception can vary across languages, cultures, and proficiency levels. Automatic systems trained on one dataset may misinterpret the communication styles of non-native speakers or minority groups, leading to biased or unfair predictions. Mislabeling an author's intent could have reputational or social consequences, particularly in sensitive domains such as education, hiring, or law enforcement. We therefore caution against any deployment of such models in consequential settings, particularly where individual judgment or profiling is involved.

 At present, automatic deception detection models should not be relied upon for high-stakes decisions. The performance of such systems remains insufficient for practical deployment, and their use in real-world scenarios risks producing incorrect or unjust outcomes. We advocate for the responsible and transparent use of such models, and stress that any deployment should be accompanied by appropriate ethical oversight.

\section*{Acknowledgments}

We gratefully acknowledge the insightful feedback provided by researchers at the Leibniz Institute for Psychology (ZPID), Trier, particularly with regard to our study design from a psychology perspective. We also extend our sincere thanks to Amelie Wührl (IT University of Copenhagen) for her valuable contributions to our discussions on the complex interplay between factuality and deception.

\bibliographystyle{compling}
\bibliography{COLI_template, my_papers}

\appendix

\newpage

\section{Belief-based Deception Survey}\label{app:survey_design}

\subsection{\textbf{Corpus I (\defabelone)}}

The survey for Corpus I was conducted using Google Forms and consisted of two distinct parts: (1) Belief Assessment and (2) Argument Generation.

\begin{figure}[h]
	\centering
	\includegraphics[trim=2cm 13cm 2cm 5cm, clip, scale=0.7]{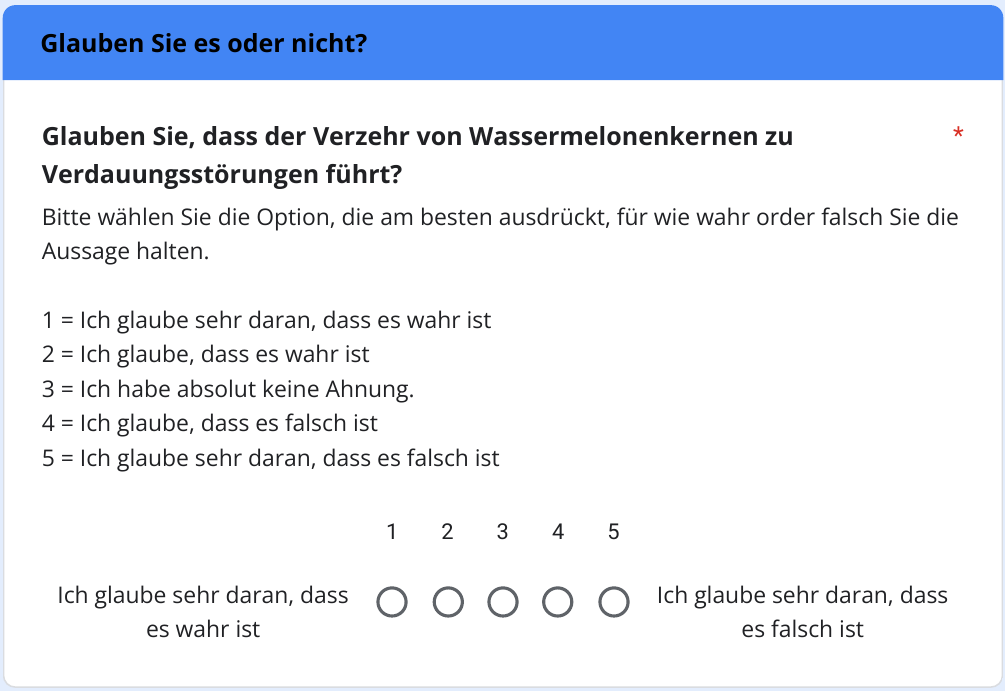}
	\caption{Belief assessment question as displayed in Google Forms}
	\label{fig:belielfqns}
\end{figure}

\paragraph{Belief Assessment}: In this step, participants were asked to indicate their level of agreement with a given statement using a 5-point Likert scale, as illustrated in Figure~\ref{fig:belielfqns}. To ensure attentiveness, five attention check questions were interspersed throughout the survey. 

\begin{figure}[h]
	\centering
	\includegraphics[trim=3.7cm 1.8cm 3.7cm 4.9cm, clip, scale=0.88]{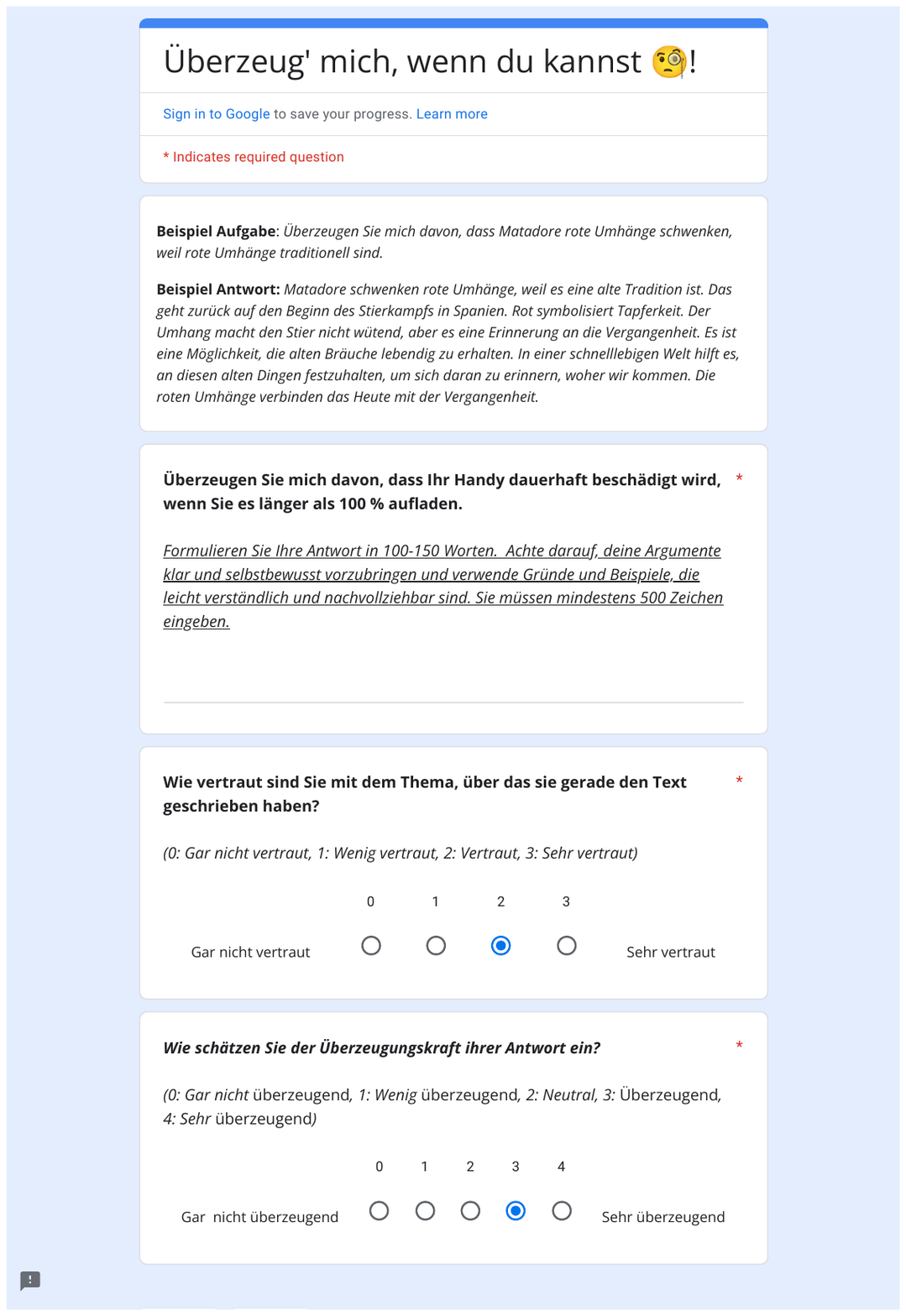}
	\caption{Argument generation task as displayed in Google Forms. }
	\label{fig:arggen}
\end{figure}

\paragraph{Argument Generation}: Here, participants were prompted to compose convincing arguments in support of a given statement. An example response was provided to set expectations for the task. After completing their written arguments, participants were asked to self-report their familiarity with the topic and their confidence in the persuasiveness of their argument, as shown in Figure~\ref{fig:arggen}. Originally, topic familiarity was recorded on a scale from 0 to 3, and confidence in argument convincingness on a scale from 0 to 4. For consistency in comparative analysis presented in this paper, these scales are normalized to 1–4 and 1–5, respectively.

\clearpage

\subsection{\textbf{Corpus II (\defabeltwo) \& Corpus III (\defabeleng)}}
\begin{figure}[h]
	\centering
	\includegraphics[width=0.75\linewidth, trim=80 60 80 55, clip]{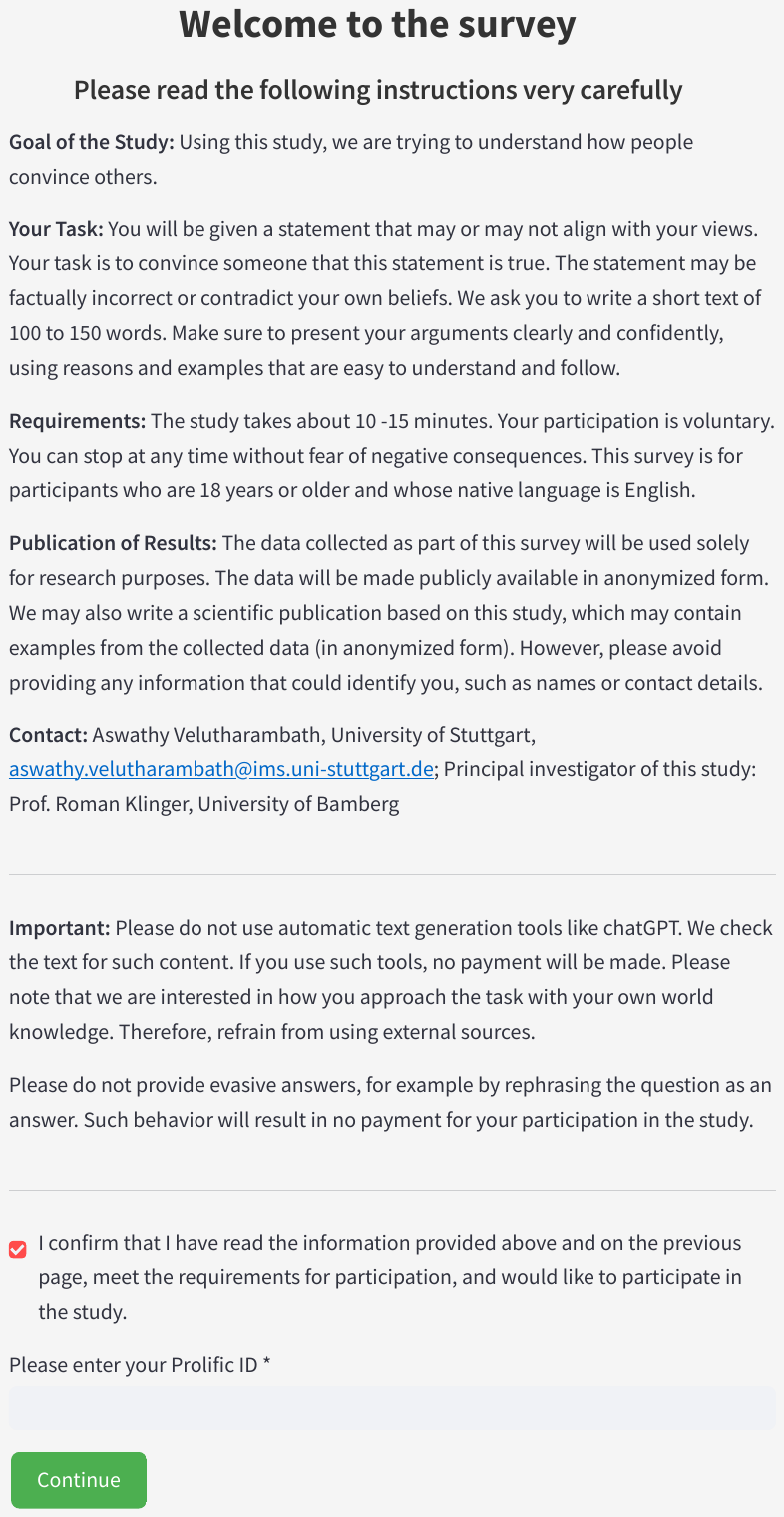} 
	\caption{Task introduction page}
	\label{fig:1survey}
\end{figure}

\begin{figure}[h]
	\centering
	\includegraphics[width=0.75\linewidth, trim=80 630 80 30, clip]{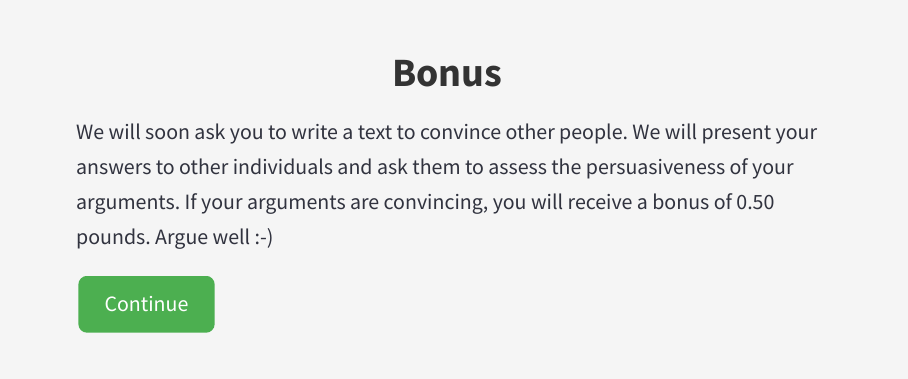}

	\includegraphics[width=0.75\linewidth, trim=80 550 80 30, clip]{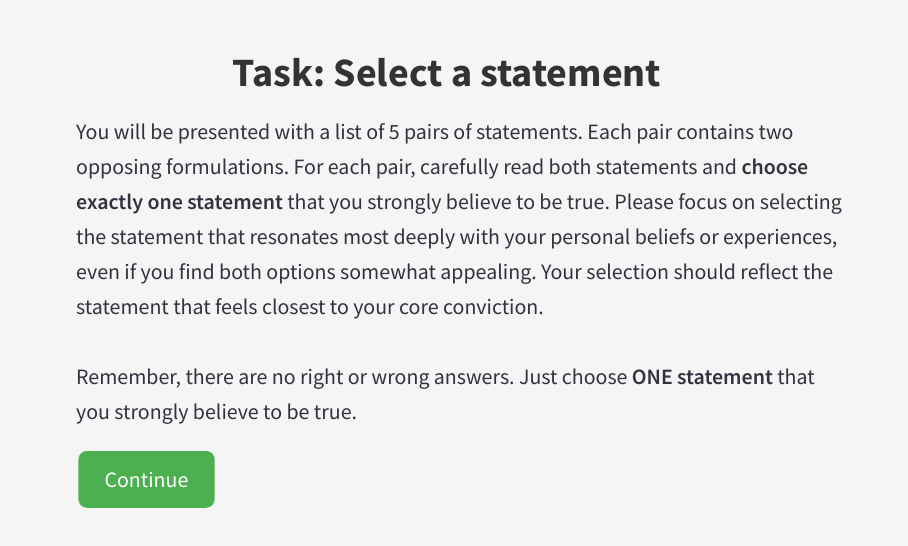}
	\caption{\textbf{Top}: Bonus announcement page. \textbf{Bottom}: Instruction for the statement selection task.}
	\label{fig:combined}
\end{figure}

\begin{figure}
	\centering
	\includegraphics[width=0.75\linewidth, trim=80 300 80 30, clip]{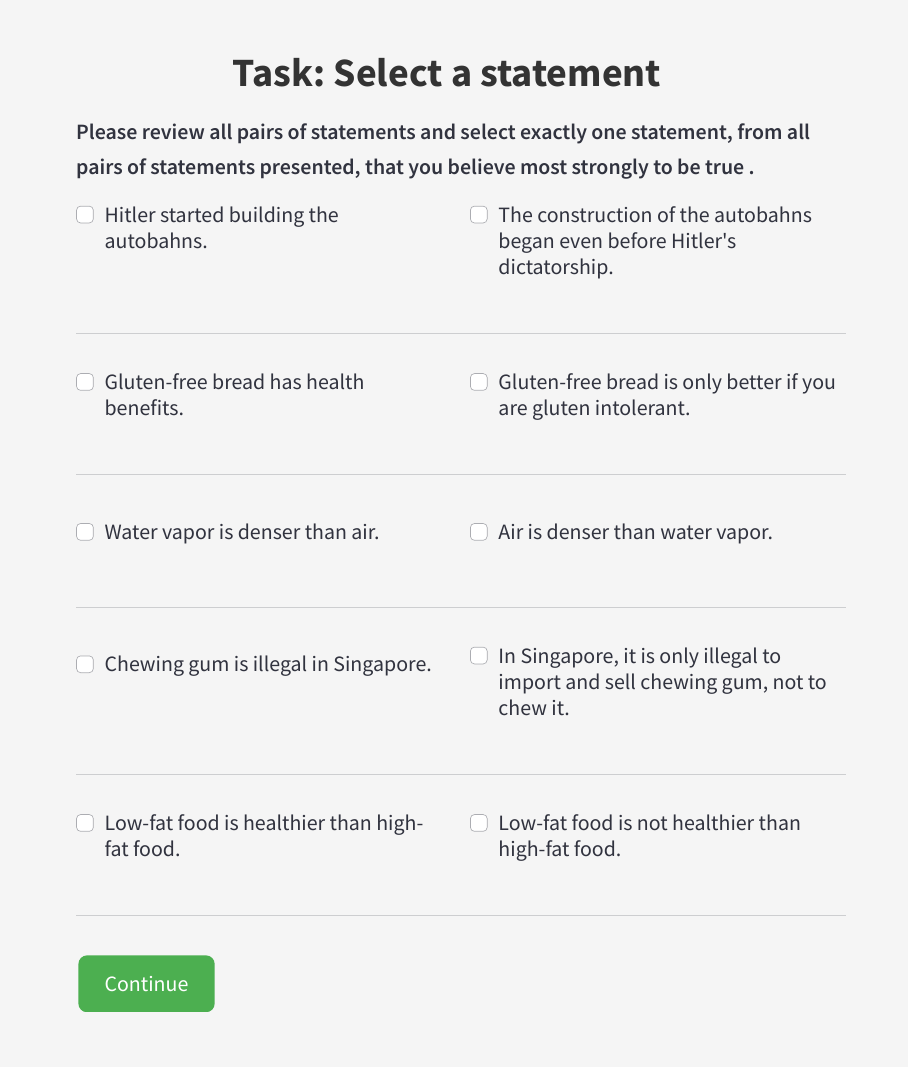}
	\caption{Statement Selection}
	\label{fig:4survey}
\end{figure}

\begin{figure}
	\centering
	\includegraphics[width=0.8\linewidth, trim=80 620 80 30, clip]{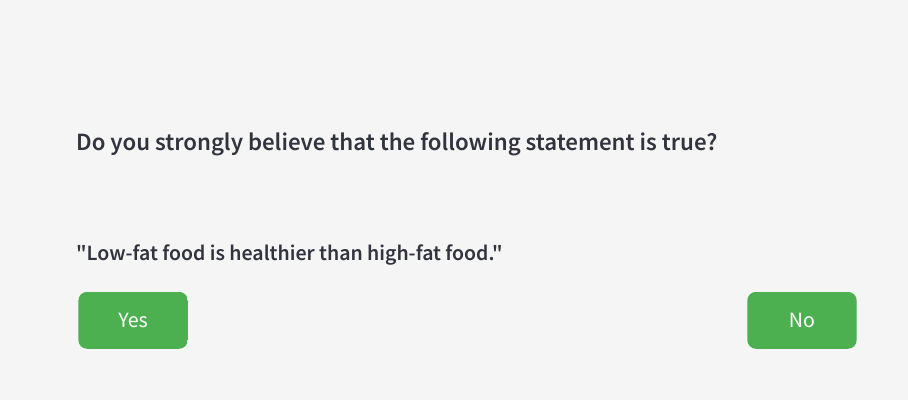}
	\caption{Belief Confirmation}
	\label{fig:5survey}
\end{figure}

\begin{figure}
	\centering
	\includegraphics[width=0.8\linewidth, trim=80 385 80 30, clip]{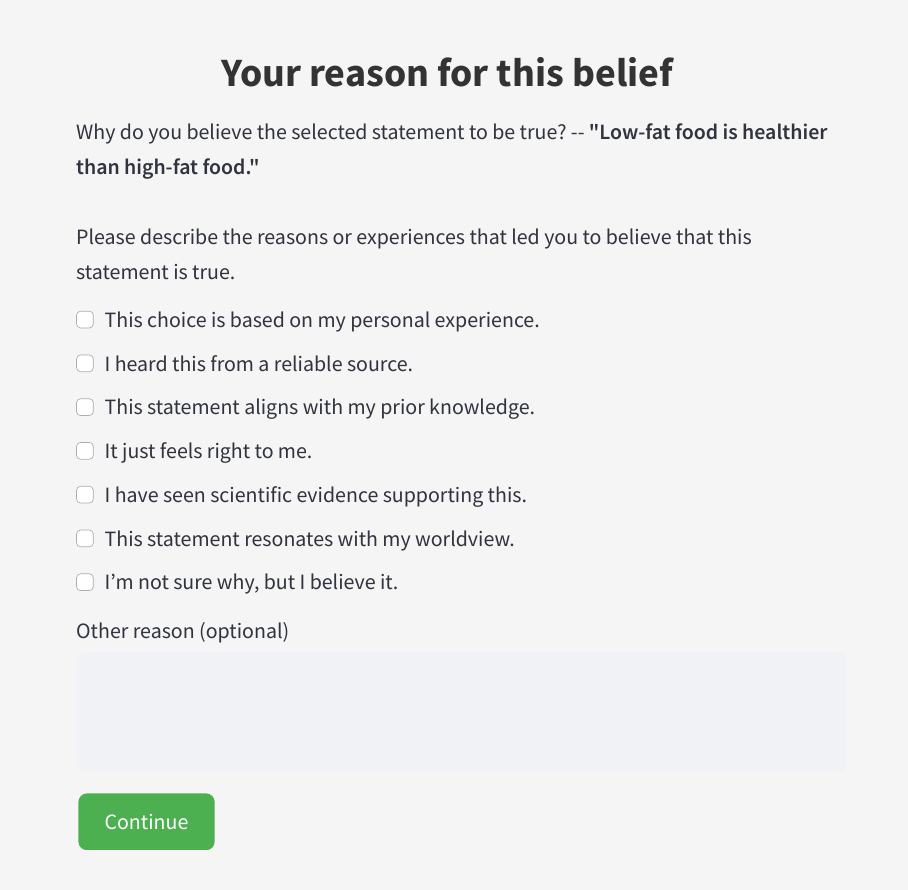}
	\caption{Belief Commitment}
	\label{fig:6survey}
\end{figure}

\begin{figure}
	\centering
	\includegraphics[width=0.8\linewidth, trim=80 600 80 30, clip]{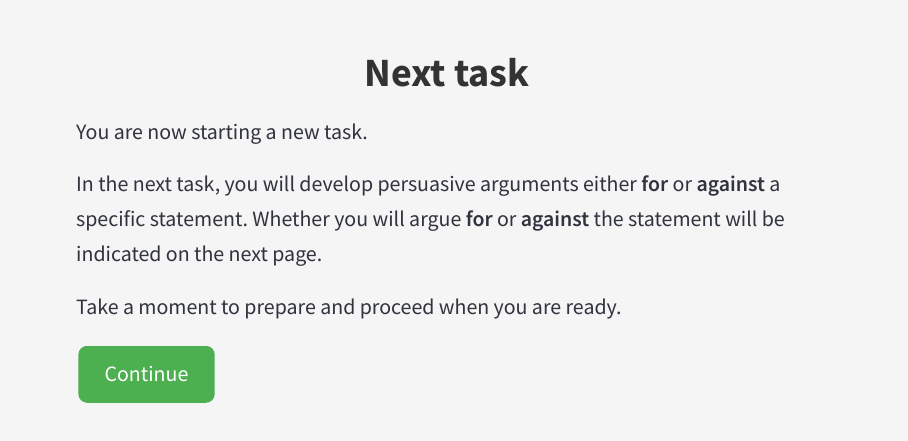}
	\caption{Task 1 - Introduction}
	\label{fig:7survey}
\end{figure}

\begin{figure}
	\centering
	\includegraphics[width=0.8\linewidth, trim=80 400 80 30, clip]{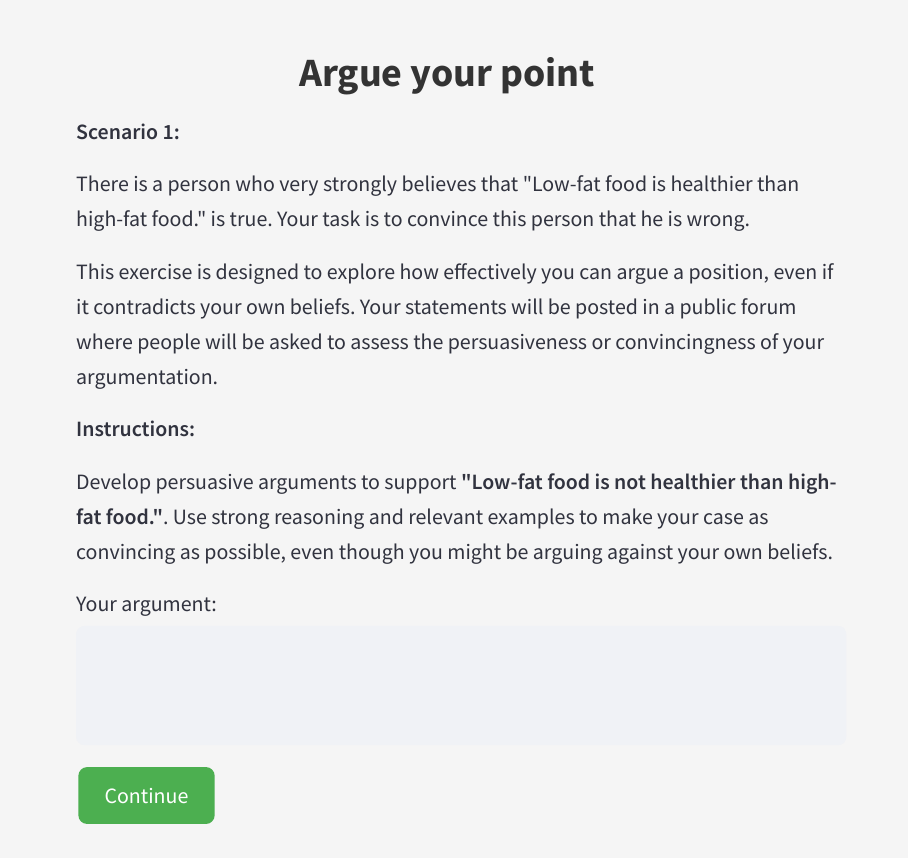}
	\caption{Task 1}
	\label{fig:8survey}
\end{figure}

\begin{figure}
	\centering
	\includegraphics[width=0.8\linewidth, trim=80 600 80 30, clip]{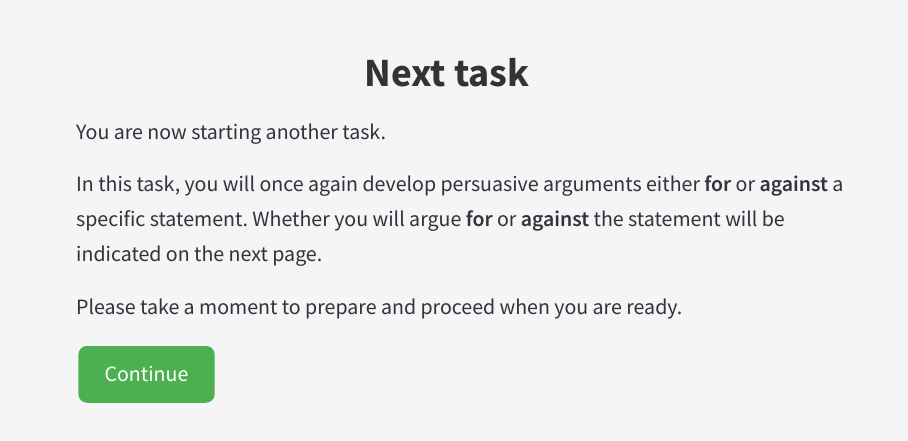}
	\caption{Task 2 - Introduction}
	\label{fig:9survey}
\end{figure}

\begin{figure}
	\centering
	\includegraphics[width=0.8\linewidth, trim=80 400 80 40, clip]{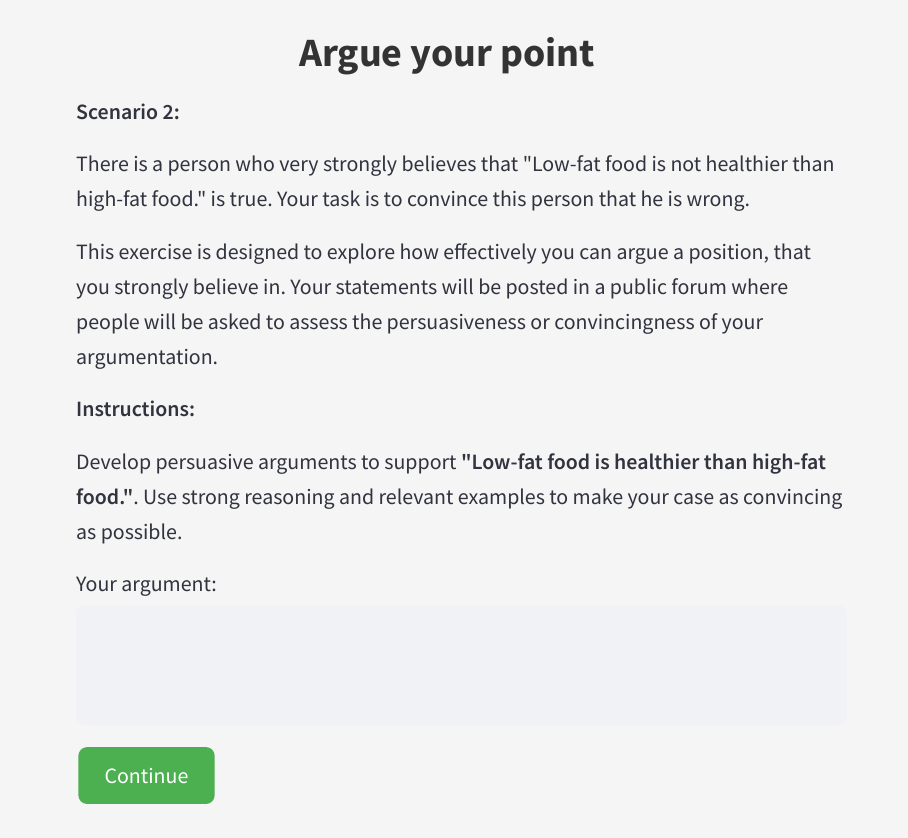}
	\caption{Task 2}
	\label{fig:10survey}
\end{figure}

\begin{figure}
	\centering
	\includegraphics[width=0.8\linewidth, trim=80 440 80 40, clip]{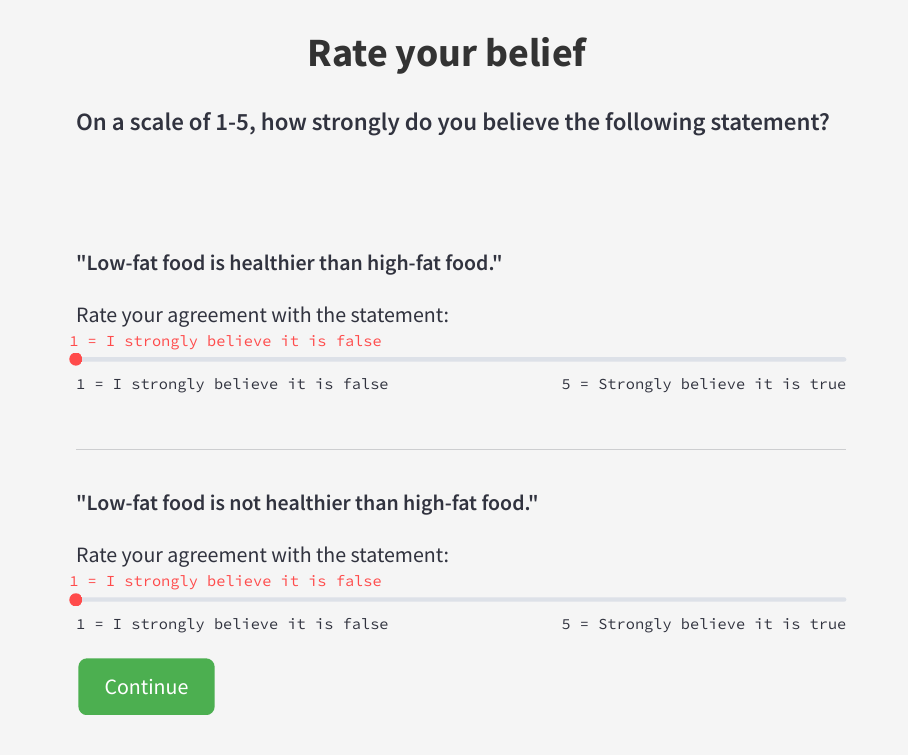}
	\caption{Belief Rating}
	\label{fig:11survey}
\end{figure}

For the data collection in \defabeltwo and \defabeleng, the survey was implemented using the \texttt{Streamlit} package in Python which is an interactive web-based application with a more dynamic design. Following are the screenshots (\ref{fig:1survey} to \ref{fig:11survey}) of the annotation interface as implemented in the Streamlit application, illustrating the layout and flow of the tasks.

\clearpage

\subsection{Stimuli Design}
The stimuli used to generate \defabelone and \defabeltwo in German differ slightly in their construction. For \defabelone, each item consists of either a factual or a non-factual variant of a given statement. In contrast, \defabeltwo includes both the factual and the non-factual variants for each statement. For the English version, \defabeleng, we translated both variants from \defabeltwo into English. Tables~\ref{tab: stimuli_set1}, ~\ref{tab: stimuli_set2} and~\ref{tab: stimuli_set3} present the full list of stimuli used for creating each of the respective corpora.
\begin{table}[ht]
	\centering
	\scriptsize
	\renewcommand{\arraystretch}{1.0}
	\begin{tabularx}{\textwidth}{p{0.25\textwidth}p{0.35\textwidth}p{0.35\textwidth}}
		\textbf{\defabelone} & \textbf{\defabeltwo} & \textbf{\defabeleng} \\
		\hline
		Der Verzehr von Wassermelonenkernen führt zu Verdauungsstörungen. & Wenn man Wassermelonenkernen isst, passieren sie das Verdauungssystem und werden ausgeschieden. \newline Der Verzehr von Wassermelonenkernen führt zu Verdauungsstörungen. & When watermelon seeds are eaten, they pass through the digestive system and are excreted. \newline The consumption of watermelon seeds leads to indigestion. \\
		\hline
		Glückskekse haben ihren Ursprung in den Vereinigten Staaten. & Glückskekse haben ihren Ursprung in China. \newline Glückskekse haben ihren Ursprung in den Vereinigten Staaten. & Fortune cookies have their origins in China. \newline Fortune cookies have their origins in the United States. \\
		\hline
		Der Mensch nutzt die meiste Zeit nur 10\% seines Gehirns. & Der Mensch nutzt die meiste Zeit nur 10\% seines Gehirns. \newline Das Gehirn des Menschen ist fast immer vollständig aktiv. & Humans use only 10\% of their brain most of the time. \newline The human brain is almost always fully active. \\
		\hline
		Die Sonne erscheint aus dem Weltraum gesehen weiß. & Die Sonne erscheint aus dem Weltraum gesehen weiß. \newline Die Sonne erscheint aus dem Weltraum gesehen gelb. & The sun appears white when viewed from space. \newline The sun appears yellow when viewed from space. \\
		\hline
		Wenn man einen Regenwurm durchschneidet, entstehen zwei Regenwürmer. & Wenn man einen Regenwurm durchschneidet, stirbt er sehr wahrscheinlich. \newline Wenn man einen Regenwurm durchschneidet, entstehen zwei Regenwürmer. & If you cut through an earthworm, you get two earthworms. \newline If you cut an earthworm, it is very likely to die. \\
		\hline
		In der Bibel wird erwähnt, welche verbotene Frucht Adam und Eva im Garten Eden aßen. & In der Bibel wird nicht erwähnt, welche verbotene Frucht Adam und Eva im Garten Eden aßen. \newline In der Bibel wird erwähnt, welche verbotene Frucht Adam und Eva im Garten Eden aßen. & The Bible mentions the forbidden fruit that Adam and Eve ate in the Garden of Eden. \newline The Bible does not mention which forbidden fruit Adam and Eve ate in the Garden of Eden. \\
		\hline
		In kleinen Dosen ist Glutamat (ein Geschmacksverstärker) schädlich für den Menschen. & In kleinen Dosen ist Glutamat (ein Geschmacksverstärker) schädlich für den Menschen. \newline Es gibt keine eindeutigen wissenschaftlichen Beweise dafür, dass Glutamat (ein Geschmacksverstärker) in kleinen Dosen für den Menschen schädlich ist. & There is no clear scientific evidence that glutamate (a flavor enhancer) is harmful to humans in small doses. \newline In small doses, glutamate (a flavor enhancer) is harmful to humans. \\
		\hline
		Man sollte nicht zu nah vor dem Fernseher sitzen, weil das zu einer vorübergehenden Ermüdung der Augen führen kann. & Man sollte nicht zu nah vor dem Fernseher sitzen, weil das zu einer vorübergehenden Ermüdung der Augen führen kann. \newline Man sollte nicht zu nah am Fernseher sitzen, da dies zur Erblindung führen kann. & You should not sit too close to the TV as this can cause temporary eye fatigue. \newline You should not sit too close to the television as this can lead to blindness. \\
		\hline
		Glutenfreies Brot hat gesundheitliche Vorteile. & Glutenfreies Brot hat gesundheitliche Vorteile. \newline Glutenfreies Brot ist nur besser, wenn man glutenintolerant ist. & Gluten-free bread is only better if you are gluten intolerant. \newline Gluten-free bread has health benefits. \\
		\hline
		Mäßiger Weinkonsum ist nachweislich gut für die Gesundheit. & Es gibt keine klaren wissenschaftlichen Beweise dafür, dass mäßiger Weinkonsum gut für die Gesundheit ist. \newline Mäßiger Weinkonsum ist nachweislich gut für die Gesundheit. & There is no clear scientific evidence that moderate wine consumption is good for your health. \newline Moderate wine consumption is proven to be good for your health. \\
		\hline
		Luft ist dichter als Wasserdampf. & Luft ist dichter als Wasserdampf. \newline Wasserdampf ist dichter als Luft. & Air is denser than water vapor. \newline Water vapor is denser than air. \\
	\hline	
	\end{tabularx}
	\caption{Stimuli used for each of the corpora (1-11 of 30)}
	\label{tab: stimuli_set1}
\end{table}

\begin{table}[ht]
	\centering
	\scriptsize
	\renewcommand{\arraystretch}{1.0}
	\begin{tabularx}{\textwidth}{p{0.25\textwidth}p{0.35\textwidth}p{0.35\textwidth}}
		\textbf{\defabelone} & \textbf{\defabeltwo} & \textbf{\defabeleng} \\
		\hline
		Bio-Lebensmittel sind gesünder als konventionelle Lebensmittel. & Bio-Lebensmittel sind gesünder als konventionelle Lebensmittel. \newline Konventionelle Lebensmittel sind genauso gesund wie Bio-Lebensmittel. & Organic food is healthier than conventional food. \newline Conventional food is just as healthy as organic food. \\
		\hline
		In der ökologischen Landwirtschaft werden Pestizide verwendet. & In der ökologischen Landwirtschaft werden Pestizide verwendet. \newline In der ökologischen Landwirtschaft sind Pestizide ausschließlich verboten. & Pesticides are strictly prohibited in organic farming. \newline Pesticides are used in organic farming. \\
		\hline
		Das Kauen von Kaugummi ist in Singapur illegal. & Das Kauen von Kaugummi ist in Singapur illegal. \newline In Singapur ist es nur illegal, Kaugummi zu importieren und zu verkaufen, nicht es zu kauen. & Chewing gum is illegal in Singapore. \newline In Singapore, it is only illegal to import and sell chewing gum, not to chew it. \\
		\hline
		
		Mehr Amerikaner besuchen eine Hochschule oder Universität als Europäer. & Es besuchen mehr Amerikaner als Europäer eine Hochschule oder Universität. \newline Es besuchen mehr Europäer als Amerikaner eine Hochschule oder Universität. & More Europeans attend college or university than Americans. \newline More Americans than Europeans attend college or university. \\
		\hline
				Atheisten schreiben häufiger 'Xmas' als Christen, die eher 'Weihnachten' schreiben. & Die Verwendung von 'Xmas' oder 'Weihnachten' hat nichts mit Glauben zu tun, sondern ist lediglich eine Abkürzung. \newline Atheisten schreiben häufiger 'Xmas' als Christen, die eher 'Weihnachten' schreiben. & The use of 'Xmas' or 'Christmas' has nothing to do with faith, but is merely an abbreviation. \newline Atheists are more likely to write 'Xmas' than Christians, who are more likely to write 'Christmas'. \\
		\hline
		Der Verzehr von Brokkoli kann Ihre Lebenserwartung um viele Jahrzehnte verlängern. & Es wurde wissenschaftlich nicht nachgewiesen, dass der Verzehr von Gemüse Ihre Lebenserwartung um viele Jahrzehnte verlängert. \newline Der Verzehr von Brokkoli kann Ihre Lebenserwartung um viele Jahrzehnte verlängern. & It has not been scientifically proven that eating vegetables extends your life expectancy by many decades. \newline Eating broccoli can extend your life expectancy by many decades. \\
		\hline
		Es gibt auf der Erde einen Ort, an dem immer Nacht ist. & Es gibt auf der Erde einen Ort, an dem immer Nacht ist. \newline Die Sonne scheint nie am Südpol. & There is a place on earth where it is always night. \newline The sun never shines at the South Pole. \\
		\hline
		Manche Menschen können im Schlaf lernen. & Manche Menschen können im Schlaf lernen. \newline Es gibt keine wissenschaftlichen Beweise dafür, dass Menschen im Schlaf lernen können. & Some people can learn in their sleep. \newline There is no scientific evidence that people can learn in their sleep. \\
		\hline
		Es gibt Impfstoffe, die dauerhaft deine DNA verändern können. & Impfstoffe haben keine dauerhaften Auswirkungen auf die DNA. \newline Es gibt Impfstoffe, die dauerhaft die DNA verändern können. & Vaccines have no lasting effects on DNA. \newline There are vaccines that can permanently alter the DNA. \\
		\hline
		Hitler hat mit dem Bau der Autobahnen begonnen. & Der Bau der Autobahnen begann bereits vor Hitlers Diktatur. \newline Hitler hat mit dem Bau der Autobahnen begonnen. & Hitler started building the autobahns. \newline The construction of the autobahns began even before Hitler's dictatorship. \\
		\hline
		Haie können Krebs bekommen. & Haie können Krebs bekommen. \newline Haie bekommen niemals Krebs, weil ihr Knorpel als Krebsschutz wirkt. & Sharks can get cancer. \newline Sharks never get cancer because their cartilage acts as cancer protection. \\
		\hline

		Kamele speichern Fett in ihren Höckern. & Kamele speichern Wasser in ihren Höckern. \newline Kamele speichern Fett in ihren Höckern. & Camels store water in their humps. \newline Camels store fat in their humps. \\
		\hline
		Die Erde ist im Januar der Sonne am nächsten. & Die Erde ist im Juli der Sonne am nächsten. \newline Die Erde ist im Januar der Sonne am nächsten. & The earth is closest to the sun in July. \newline The earth is closest to the sun in January. \\
		\hline
	\end{tabularx}
	\caption{Stimuli used for each of the corpora (12-24 of 30)}
	\label{tab: stimuli_set2}
\end{table}

\begin{table}
	\centering
	\scriptsize
	\renewcommand{\arraystretch}{1.0}
	\begin{tabularx}{\textwidth}{p{0.25\textwidth}p{0.35\textwidth}p{0.35\textwidth}}
		
		\textbf{\defabelone} & \textbf{\defabeltwo} & \textbf{\defabeleng} \\

		\hline
		Die Nutzung von Handys an Tankstellen kann eine Explosion verursachen. & Die Nutzung von Handys an Tankstellen kann eine Explosion verursachen. \newline Die Nutzung von Handys an Tankstellen ist untersagt, um mögliche Sicherheitsrisiken durch Ablenkung zu minimieren. & The use of cell phones at filling stations is prohibited in order to minimize possible safety risks due to distraction. \newline The use of cell phones at gas stations can cause an explosion. \\
		\hline
		Ihr Handy wird dauerhaft beschädigt, wenn Sie länger als 100 \% aufladen. & Ihr Handy wird dauerhaft beschädigt, wenn Sie es länger als 100 \% aufladen. \newline Ihr Handy kann problemlos länger als bis zu 100 \% aufgeladen werden. & Your cell phone can easily be charged for longer than 100\%. \newline Your cell phone will be permanently damaged if you charge it for longer than 100\%. \\
		\hline
		Affen können genauso gut addieren wie Menschen. & Kein Tier kann genauso gut wie der Mensch Zahlen addieren. \newline Affen können genauso gut addieren wie Menschen. & No animal can add numbers as well as humans. \newline Monkeys can add just as well as humans. \\
		\hline

		Es ist in Deutschland illegal, barfuß Auto zu fahren. & Es ist in Deutschland illegal, barfuß Auto zu fahren. \newline Es gibt keine gesetzliche Einschränkung in Deutschland, barfuß Auto zu fahren. & It is illegal to drive barefoot in Germany. \newline There is no legal restriction on driving barefoot in Germany. \\
		\hline
		Fettarmes Essen ist gesünder als fettreiches Essen. & Fettarmes Essen ist gesünder als fettreiches Essen. \newline Fettarmes Essen ist nicht gesünder als fettreiche Essen. & Low-fat food is healthier than high-fat food. \newline Low-fat food is not healthier than high-fat food. \\
		\hline
		Tigerhaie untergehen, wenn sie aufhören zu schwimmen. & Tigerhaie gehen unter, wenn sie aufhören zu schwimmen. \newline Tigerhaie treiben auf dem Wasser, wenn sie aufhören zu schwimmen. & Tiger sharks float on the water when they stop swimming. \newline Tiger sharks sink when they stop swimming. \\
		\hline
	\end{tabularx}
	\caption{Stimuli used for each of the corpora (24-30 of 30)}
	\label{tab: stimuli_set3}
\end{table}

\clearpage

\section{Experiment Details}

\subsection{Linguistic Cues as Features for Deception Detection}
To analyze deception in language, we extract a rich set of linguistic cues grounded in prior psycholinguistic and computational studies. These features span multiple dimensions, including affective tone, complexity, specificity, and syntactic structure, and are designed to capture both subtle and explicit markers of deceptive behavior. Table~\ref{tab:linguistic_cues} provides an overview of the feature categories, the specific cues used, and how each is operationalized.

\begin{table}[htbp]
	\scriptsize
	\centering
	\begin{tabularx}{0.85\textwidth}{lll} 
		\textbf{Linguistic cues} & \textbf{Features} & \textbf{Operationalization} \\
		\cmidrule{1-3}
		Authenticity & Authentic & Authenticity score (LIWC) \\
		\cmidrule{1-3}
		Affect & Tone, Affect, tone\_pos, tone\_neg & Sentiment score \\
		& emotion, emo\_pos, emo\_neg & Emotion scores \\
		& emo\_anx, emo\_anger, emo\_sad & Anxiety, anger, sadness scores \\
		& positive, negative, neutral & Positive, negative, neutral scores \\
		\cmidrule{1-3}
		Quantity & WPS, word\_count, sentence\_count & Words per sentence, word and sentence counts \\
		& syllable\_count, avg\_syllables\_per\_word & Syllable count, average syllables per word \\
		& verb, VERB & Verb frequency \\
		\cmidrule{1-3}
		Distancing & ppron, ipron, negate & Pronouns and negations \\
		& i, we, you, shehe, they & Personal pronouns \\
		& self\_reference, other\_reference & Self and other references \\
		\cmidrule{1-3}
		Activation & adverb & Frequency of adverbs \\
		\cmidrule{1-3}
		Complexity & adj & Frequency of adjectives \\
		& differ & Exclusive words \\
		& flesch\_kincaid\_grade & Flesch--Kincaid readability \\
		& gunning\_fog & Gunning--Fog readability \\
		\cmidrule{1-3}
		Specificity & quantity & Frequency of quantifiers \\
		& Cognition, cogproc & Cognitive processing words \\
		& insight, cause, memory & Insight, causation, memory words \\
		& Perception, attention & Perception and attention words \\
		& motion, space, time & Motion, space, and time references \\
		& visual, auditory, feeling & Visual, auditory, feeling references \\
		& PROPN & Proper noun frequency \\
		& people\_entities, temporal\_entities & Entity recognition: people, temporal \\
		& spatial\_entities, quantity\_entities & Entity recognition: spatial, quantity \\
		& perceptual\_details & Perceptual detail markers \\
		& contextual\_embedding & Contextual embedding score \\
		& reality\_monitoring & Reality monitoring features \\
		\cmidrule{1-3}
		Uncertainty & discrep, tentat, certitude & Discrepancy, tentative, certitude markers \\
		\cmidrule{1-3}
		Diversity & lexical\_diversity & Lexical diversity (unique tokens / tokens) \\
		& difficult\_words & Difficult word count \\
	\end{tabularx}
	\caption{Linguistic cues and their operationalization}
	\label{tab:linguistic_cues}
\end{table}

\subsection{Linguistic Correlation Analysis}
Table \ref{tab:feature_by_source_de} and Table \ref{tab:feature_by_source_en} shows the Point-biserial correlation values  and the lower bound and upper bound of the 95\% confidence interval, of the categories with correlations considered significant (p-value < 0.05) are marked with a * symbol

{\renewcommand{\arraystretch}{0.7}
	\setlength{\tabcolsep}{3pt}
	\begin{table}[!h]
		\centering \scriptsize
		\resizebox{\ifdim\width>\linewidth\linewidth\else\width\fi}{!}{
			\begin{tabular}[t]{lcc}
				\toprule
				& \defabelone & \defabeltwo \\
				\cmidrule{2-3}
				Categories & correlation[CI] & correlation[CI]\\
				\cmidrule{1-3}
				Authentic & -0.01 [-0.08, \phantom{-}0.05] & -0.05 [-0.11, \phantom{-}0.01]\\
				Tone & -0.00 [-0.06, \phantom{-}0.06] & -0.03 [-0.09, \phantom{-}0.03]\\
				affect & \phantom{-}0.05 [-0.01, \phantom{-}0.12] & \phantom{-}0.05 [-0.01, \phantom{-}0.11]\\
				positive & \phantom{-}0.03 [-0.03, \phantom{-}0.09] & -0.02 [-0.08, \phantom{-}0.05]\\
				negative & \phantom{-}0.05 [-0.01, \phantom{-}0.11] & \phantom{-}0.04 [-0.03, \phantom{-}0.10]\\
				\addlinespace
				neutral & -0.06 [-0.12, \phantom{-}0.01] & -0.02 [-0.08, \phantom{-}0.04]\\
				posemo & \phantom{-}0.02 [-0.04, \phantom{-}0.08] & \phantom{-}0.02 [-0.05, \phantom{-}0.08]\\
				negemo & \phantom{-}0.06 [-0.00, \phantom{-}0.12] & \phantom{-}0.05 [-0.02, \phantom{-}0.11]\\
				anx & \phantom{-}0.07 [\phantom{-}0.01, \phantom{-}0.13] & \phantom{-}0.08 [\phantom{-}0.02, \phantom{-}0.14]\\
				anger & \phantom{-}0.07 [\phantom{-}0.01, \phantom{-}0.13] & \phantom{-}0.04 [-0.03, \phantom{-}0.10]\\
				\addlinespace
				sad & -0.03 [-0.09, \phantom{-}0.03] & -0.00 [-0.07, \phantom{-}0.06]\\
				WPS & -0.02 [-0.08, \phantom{-}0.04] & -0.00 [-0.07, \phantom{-}0.06]\\
				verb & \phantom{-}0.01 [-0.05, \phantom{-}0.07] & \phantom{-}0.01 [-0.05, \phantom{-}0.08]\\
				word\_count & -0.05 [-0.11, \phantom{-}0.01] & -0.08 [-0.14, -0.02]\\
				sentence\_count & -0.02 [-0.08, \phantom{-}0.04] & -0.09 [-0.15, -0.02]\\
				\addlinespace
				syllable\_count & -0.05 [-0.11, \phantom{-}0.01] & -0.08 [-0.14, -0.02]\\
				avg\_syllables\_per\_word & \phantom{-}0.01 [-0.05, \phantom{-}0.07] & -0.01 [-0.07, \phantom{-}0.05]\\
				VERB & \phantom{-}0.01 [-0.05, \phantom{-}0.07] & \phantom{-}0.00 [-0.06, \phantom{-}0.06]\\
				ppron & -0.05 [-0.11, \phantom{-}0.01] & \phantom{-}0.03 [-0.03, \phantom{-}0.09]\\
				i & -0.01 [-0.07, \phantom{-}0.05] & -0.00 [-0.07, \phantom{-}0.06]\\
				\addlinespace
				we & -0.02 [-0.08, \phantom{-}0.05] & -0.00 [-0.07, \phantom{-}0.06]\\
				you\_sing & -0.02 [-0.08, \phantom{-}0.05] & -0.01 [-0.07, \phantom{-}0.05]\\
				you\_plur & -0.02 [-0.08, \phantom{-}0.04] & \phantom{-}0.01 [-0.05, \phantom{-}0.08]\\
				shehe & -0.03 [-0.09, \phantom{-}0.03] & \phantom{-}0.05 [-0.02, \phantom{-}0.11]\\
				they & -0.00 [-0.07, \phantom{-}0.06] & \phantom{-}0.03 [-0.03, \phantom{-}0.09]\\
				\addlinespace
				ipron & \phantom{-}0.02 [-0.04, \phantom{-}0.08] & -0.04 [-0.10, \phantom{-}0.03]\\
				negate & -0.02 [-0.09, \phantom{-}0.04] & \phantom{-}0.05 [-0.01, \phantom{-}0.11]\\
				self\_reference & -0.02 [-0.08, \phantom{-}0.04] & -0.00 [-0.07, \phantom{-}0.06]\\
				other\_reference & -0.02 [-0.08, \phantom{-}0.04] & \phantom{-}0.04 [-0.02, \phantom{-}0.10]\\
				adverb & -0.06 [-0.12, \phantom{-}0.00] & -0.04 [-0.10, \phantom{-}0.03]\\
				\addlinespace
				adj & \phantom{-}0.06 [-0.01, \phantom{-}0.12] & \phantom{-}0.00 [-0.06, \phantom{-}0.06]\\
				differ & \phantom{-}0.01 [-0.05, \phantom{-}0.07] & -0.03 [-0.10, \phantom{-}0.03]\\
				flesch\_kincaid\_grade & -0.02 [-0.08, \phantom{-}0.04] & \phantom{-}0.00 [-0.06, \phantom{-}0.07]\\
				gunning\_fog & -0.02 [-0.08, \phantom{-}0.04] & \phantom{-}0.01 [-0.05, \phantom{-}0.07]\\
				cogproc & \phantom{-}0.01 [-0.05, \phantom{-}0.07] & -0.00 [-0.06, \phantom{-}0.06]\\
				\addlinespace
				insight & -0.00 [-0.06, \phantom{-}0.06] & \phantom{-}0.01 [-0.05, \phantom{-}0.07]\\
				cause & -0.03 [-0.09, \phantom{-}0.03] & \phantom{-}0.06 [-0.00, \phantom{-}0.12]\\
				motion & -0.05 [-0.11, \phantom{-}0.02] & \phantom{-}0.02 [-0.04, \phantom{-}0.08]\\
				space & -0.01 [-0.07, \phantom{-}0.05] & -0.02 [-0.08, \phantom{-}0.04]\\
				see & \phantom{-}0.09 [\phantom{-}0.03, \phantom{-}0.15] & -0.02 [-0.09, \phantom{-}0.04]\\
				\addlinespace
				hear & -0.04 [-0.10, \phantom{-}0.02] & -0.02 [-0.08, \phantom{-}0.04]\\
				feel & -0.00 [-0.06, \phantom{-}0.06] & \phantom{-}0.02 [-0.04, \phantom{-}0.09]\\
				percept & \phantom{-}0.08 [\phantom{-}0.02, \phantom{-}0.14] & -0.00 [-0.07, \phantom{-}0.06]\\
				time & -0.02 [-0.08, \phantom{-}0.04] & -0.05 [-0.12, \phantom{-}0.01]\\
				PROPN & -0.01 [-0.07, \phantom{-}0.05] & -0.03 [-0.10, \phantom{-}0.03]\\
				\addlinespace
				people\_entities & \phantom{-}0.07 [\phantom{-}0.00, \phantom{-}0.13] & -0.04 [-0.10, \phantom{-}0.03]\\
				spatial\_entities & \phantom{-}0.02 [-0.05, \phantom{-}0.08] & -0.05 [-0.11, \phantom{-}0.01]\\
				perceptual\_details & \phantom{-}0.07 [\phantom{-}0.01, \phantom{-}0.13] & -0.01 [-0.08, \phantom{-}0.05]\\
				contextual\_embedding & -0.03 [-0.09, \phantom{-}0.03] & -0.04 [-0.10, \phantom{-}0.02]\\
				reality\_monitoring & \phantom{-}0.03 [-0.03, \phantom{-}0.09] & -0.01 [-0.08, \phantom{-}0.05]\\
				\addlinespace
				discrep & -0.02 [-0.08, \phantom{-}0.04] & -0.02 [-0.09, \phantom{-}0.04]\\
				tentat & \phantom{-}0.05 [-0.01, \phantom{-}0.11] & -0.05 [-0.11, \phantom{-}0.01]\\
				certain & -0.03 [-0.09, \phantom{-}0.03] & \phantom{-}0.03 [-0.04, \phantom{-}0.09]\\
				lexical\_diversity & \phantom{-}0.03 [-0.03, \phantom{-}0.09] & \phantom{-}0.05 [-0.02, \phantom{-}0.11]\\
				difficult\_words & -0.06 [-0.12, \phantom{-}0.00] & -0.08 [-0.14, -0.02]\\
				\bottomrule
		\end{tabular}}
		\caption{\label{tab:feature_by_source_de}Point-biserial Correlations by dataset (German)}
	\end{table}
}

{\renewcommand{\arraystretch}{0.8}
	\setlength{\tabcolsep}{3pt}
	\begin{table}
		\centering
		\resizebox{\ifdim\width>\linewidth\linewidth\else\width\fi}{!}{
			\begin{tabular}[t]{lccccc}
				\toprule
				& DEFABEL\_EN & CROSS\_CULT\_DE & DEC\_OP & INTENTION & OPEN\_DOMAIN \\
				\cmidrule{2-6}
				Categories & correlation[CI] & correlation[CI] & correlation[CI] & correlation[CI]& correlation[CI]\\
				\midrule
				Authentic & -0.05 [-0.12, \phantom{-}0.02] & \phantom{-}0.28 [\phantom{-}0.22, \phantom{-}0.33]* & \phantom{-}0.22 [\phantom{-}0.18, \phantom{-}0.26]* & -0.04 [-0.08, \phantom{-}0.01] & -0.04 [-0.07, -0.02]*\\
				Tone & -0.01 [-0.08, \phantom{-}0.06] & -0.01 [-0.07, \phantom{-}0.04] & \phantom{-}0.14 [\phantom{-}0.10, \phantom{-}0.18]* & \phantom{-}0.07 [\phantom{-}0.03, \phantom{-}0.12] & \phantom{-}0.01 [-0.02, \phantom{-}0.03]\\
				Affect & \phantom{-}0.02 [-0.05, \phantom{-}0.09] & -0.09 [-0.15, -0.03] & -0.05 [-0.09, -0.01] & \phantom{-}0.05 [\phantom{-}0.00, \phantom{-}0.10] & -0.00 [-0.03, \phantom{-}0.02]\\
				tone\_pos & -0.00 [-0.07, \phantom{-}0.07] & -0.05 [-0.11, \phantom{-}0.01] & \phantom{-}0.13 [\phantom{-}0.09, \phantom{-}0.17]* & \phantom{-}0.05 [\phantom{-}0.00, \phantom{-}0.10] & \phantom{-}0.02 [-0.00, \phantom{-}0.04]\\
				tone\_neg & \phantom{-}0.03 [-0.04, \phantom{-}0.10] & -0.04 [-0.10, \phantom{-}0.02] & -0.14 [-0.18, -0.10]* & \phantom{-}0.01 [-0.04, \phantom{-}0.05] & -0.03 [-0.06, -0.01]\\
				\addlinespace
				emotion & -0.02 [-0.09, \phantom{-}0.05] & -0.08 [-0.14, -0.03] & \phantom{-}0.08 [\phantom{-}0.04, \phantom{-}0.12]* & \phantom{-}0.07 [\phantom{-}0.02, \phantom{-}0.12] & \phantom{-}0.02 [-0.00, \phantom{-}0.04]\\
				emo\_pos & -0.03 [-0.10, \phantom{-}0.04] & -0.09 [-0.15, -0.04] & \phantom{-}0.06 [\phantom{-}0.02, \phantom{-}0.10] & \phantom{-}0.06 [\phantom{-}0.01, \phantom{-}0.11] & \phantom{-}0.04 [\phantom{-}0.02, \phantom{-}0.06]*\\
				emo\_neg & -0.00 [-0.07, \phantom{-}0.07] & -0.00 [-0.06, \phantom{-}0.05] & \phantom{-}0.03 [-0.01, \phantom{-}0.07] & \phantom{-}0.04 [-0.01, \phantom{-}0.09] & -0.02 [-0.04, \phantom{-}0.00]\\
				emo\_anx & \phantom{-}0.04 [-0.03, \phantom{-}0.11] & \phantom{-}0.06 [-0.00, \phantom{-}0.11] & \phantom{-}0.02 [-0.02, \phantom{-}0.06] & \phantom{-}0.04 [-0.01, \phantom{-}0.09] & -0.01 [-0.04, \phantom{-}0.01]\\
				emo\_anger & -0.02 [-0.09, \phantom{-}0.04] & -0.05 [-0.11, \phantom{-}0.01] & \phantom{-}0.02 [-0.02, \phantom{-}0.06] & -0.01 [-0.05, \phantom{-}0.04] & -0.03 [-0.05, -0.00]\\
				\addlinespace
				emo\_sad & \phantom{-}0.02 [-0.05, \phantom{-}0.09] & \phantom{-}0.02 [-0.04, \phantom{-}0.07] & \phantom{-}0.04 [\phantom{-}0.01, \phantom{-}0.08] & \phantom{-}0.01 [-0.04, \phantom{-}0.06] & -0.01 [-0.04, \phantom{-}0.01]\\
				positive & \phantom{-}0.00 [-0.07, \phantom{-}0.07] & -0.03 [-0.08, \phantom{-}0.03] & \phantom{-}0.16 [\phantom{-}0.12, \phantom{-}0.20]* & \phantom{-}0.10 [\phantom{-}0.05, \phantom{-}0.15]* & \phantom{-}0.02 [-0.00, \phantom{-}0.05]\\
				negative & \phantom{-}0.04 [-0.03, \phantom{-}0.11] & -0.01 [-0.07, \phantom{-}0.04] & -0.22 [-0.26, -0.18]* & \phantom{-}0.01 [-0.04, \phantom{-}0.06] & -0.04 [-0.07, -0.02]*\\
				neutral & -0.03 [-0.10, \phantom{-}0.04] & \phantom{-}0.04 [-0.02, \phantom{-}0.10] & \phantom{-}0.09 [\phantom{-}0.05, \phantom{-}0.13]* & -0.10 [-0.15, -0.05]* & \phantom{-}0.01 [-0.02, \phantom{-}0.03]\\
				WPS & -0.00 [-0.07, \phantom{-}0.07] & \phantom{-}0.07 [\phantom{-}0.01, \phantom{-}0.12] & \phantom{-}0.22 [\phantom{-}0.18, \phantom{-}0.26]* & \phantom{-}0.07 [\phantom{-}0.02, \phantom{-}0.12] & -0.02 [-0.04, \phantom{-}0.00]\\
				\addlinespace
				verb & -0.00 [-0.07, \phantom{-}0.06] & -0.06 [-0.12, -0.00] & -0.07 [-0.10, -0.03] & \phantom{-}0.17 [\phantom{-}0.13, \phantom{-}0.22]* & -0.01 [-0.03, \phantom{-}0.02]\\
				word\_count & -0.09 [-0.16, -0.02] & \phantom{-}0.22 [\phantom{-}0.16, \phantom{-}0.27]* & \phantom{-}0.25 [\phantom{-}0.21, \phantom{-}0.29]* & \phantom{-}0.11 [\phantom{-}0.06, \phantom{-}0.15]* & -0.02 [-0.04, \phantom{-}0.00]\\
				sentence\_count & -0.08 [-0.15, -0.01] & \phantom{-}0.14 [\phantom{-}0.09, \phantom{-}0.20]* & \phantom{-}0.08 [\phantom{-}0.04, \phantom{-}0.12]* & \phantom{-}0.05 [-0.00, \phantom{-}0.09] & -0.00 [-0.02, \phantom{-}0.02]\\
				syllable\_count & -0.08 [-0.15, -0.01] & \phantom{-}0.20 [\phantom{-}0.15, \phantom{-}0.26]* & \phantom{-}0.26 [\phantom{-}0.23, \phantom{-}0.30]* & \phantom{-}0.10 [\phantom{-}0.05, \phantom{-}0.15]* & -0.00 [-0.03, \phantom{-}0.02]\\
				avg\_syllables\_per\_word & \phantom{-}0.01 [-0.05, \phantom{-}0.08] & -0.03 [-0.08, \phantom{-}0.03] & \phantom{-}0.09 [\phantom{-}0.05, \phantom{-}0.12]* & -0.02 [-0.06, \phantom{-}0.03] & \phantom{-}0.04 [\phantom{-}0.02, \phantom{-}0.07]*\\
				\addlinespace
				VERB & \phantom{-}0.07 [-0.00, \phantom{-}0.14] & -0.06 [-0.11, -0.00] & -0.01 [-0.05, \phantom{-}0.03] & \phantom{-}0.01 [-0.03, \phantom{-}0.06] & -0.04 [-0.06, -0.02]*\\
				ppron & \phantom{-}0.02 [-0.05, \phantom{-}0.09] & -0.00 [-0.06, \phantom{-}0.06] & \phantom{-}0.04 [\phantom{-}0.00, \phantom{-}0.08] & \phantom{-}0.12 [\phantom{-}0.08, \phantom{-}0.17]* & -0.06 [-0.08, -0.04]*\\
				i & -0.04 [-0.11, \phantom{-}0.02] & \phantom{-}0.13 [\phantom{-}0.07, \phantom{-}0.19]* & \phantom{-}0.30 [\phantom{-}0.27, \phantom{-}0.34]* & \phantom{-}0.12 [\phantom{-}0.07, \phantom{-}0.17]* & -0.04 [-0.06, -0.01]\\
				we & -0.01 [-0.08, \phantom{-}0.06] & \phantom{-}0.13 [\phantom{-}0.07, \phantom{-}0.18]* & -0.15 [-0.19, -0.12]* & -0.05 [-0.10, -0.00] & \phantom{-}0.00 [-0.02, \phantom{-}0.03]\\
				you & \phantom{-}0.03 [-0.04, \phantom{-}0.10] & -0.05 [-0.11, \phantom{-}0.01] & -0.07 [-0.11, -0.03]* & \phantom{-}0.01 [-0.04, \phantom{-}0.06] & -0.05 [-0.07, -0.03]*\\
				\addlinespace
				shehe & -0.02 [-0.09, \phantom{-}0.05] & -0.15 [-0.21, -0.10]* & -0.00 [-0.04, \phantom{-}0.04] & \phantom{-}0.06 [\phantom{-}0.01, \phantom{-}0.11] & -0.04 [-0.06, -0.01]\\
				they & \phantom{-}0.04 [-0.03, \phantom{-}0.11] & -0.08 [-0.14, -0.02] & -0.08 [-0.11, -0.04]* & \phantom{-}0.04 [-0.01, \phantom{-}0.09] & -0.02 [-0.04, \phantom{-}0.00]\\
				ipron & -0.05 [-0.11, \phantom{-}0.02] & \phantom{-}0.01 [-0.05, \phantom{-}0.06] & \phantom{-}0.04 [-0.00, \phantom{-}0.08] & \phantom{-}0.14 [\phantom{-}0.09, \phantom{-}0.19]* & -0.04 [-0.06, -0.02]\\
				negate & -0.03 [-0.10, \phantom{-}0.04] & -0.13 [-0.18, -0.07]* & -0.10 [-0.14, -0.06]* & \phantom{-}0.11 [\phantom{-}0.06, \phantom{-}0.16]* & -0.09 [-0.12, -0.07]*\\
				self\_reference & -0.03 [-0.10, \phantom{-}0.04] & \phantom{-}0.17 [\phantom{-}0.12, \phantom{-}0.23]* & \phantom{-}0.15 [\phantom{-}0.11, \phantom{-}0.19]* & \phantom{-}0.09 [\phantom{-}0.04, \phantom{-}0.14]* & -0.04 [-0.06, -0.01]\\
				\addlinespace
				other\_reference & \phantom{-}0.05 [-0.02, \phantom{-}0.12] & -0.20 [-0.26, -0.15]* & -0.10 [-0.14, -0.07]* & \phantom{-}0.07 [\phantom{-}0.03, \phantom{-}0.12] & -0.06 [-0.09, -0.04]*\\
				adverb & \phantom{-}0.02 [-0.05, \phantom{-}0.09] & -0.03 [-0.09, \phantom{-}0.02] & \phantom{-}0.03 [-0.01, \phantom{-}0.07] & \phantom{-}0.13 [\phantom{-}0.09, \phantom{-}0.18]* & -0.07 [-0.09, -0.05]*\\
				adj & \phantom{-}0.03 [-0.04, \phantom{-}0.10] & \phantom{-}0.09 [\phantom{-}0.03, \phantom{-}0.15] & -0.00 [-0.04, \phantom{-}0.04] & \phantom{-}0.06 [\phantom{-}0.01, \phantom{-}0.11] & \phantom{-}0.03 [\phantom{-}0.00, \phantom{-}0.05]\\
				differ & -0.02 [-0.09, \phantom{-}0.04] & \phantom{-}0.11 [\phantom{-}0.05, \phantom{-}0.17]* & \phantom{-}0.10 [\phantom{-}0.06, \phantom{-}0.14]* & \phantom{-}0.02 [-0.03, \phantom{-}0.07] & -0.01 [-0.03, \phantom{-}0.01]\\
				flesch\_kincaid\_grade & \phantom{-}0.01 [-0.06, \phantom{-}0.07] & \phantom{-}0.09 [\phantom{-}0.03, \phantom{-}0.14] & \phantom{-}0.23 [\phantom{-}0.20, \phantom{-}0.27]* & \phantom{-}0.07 [\phantom{-}0.02, \phantom{-}0.11] & \phantom{-}0.02 [-0.00, \phantom{-}0.05]\\
				\addlinespace
				gunning\_fog & \phantom{-}0.01 [-0.06, \phantom{-}0.08] & \phantom{-}0.09 [\phantom{-}0.03, \phantom{-}0.14] & \phantom{-}0.21 [\phantom{-}0.18, \phantom{-}0.25]* & \phantom{-}0.05 [\phantom{-}0.00, \phantom{-}0.10] & \phantom{-}0.02 [-0.00, \phantom{-}0.04]\\
				quantity & \phantom{-}0.01 [-0.06, \phantom{-}0.08] & \phantom{-}0.08 [\phantom{-}0.02, \phantom{-}0.13] & \phantom{-}0.06 [\phantom{-}0.02, \phantom{-}0.10] & \phantom{-}0.03 [-0.01, \phantom{-}0.08] & -0.00 [-0.02, \phantom{-}0.02]\\
				Cognition & \phantom{-}0.02 [-0.05, \phantom{-}0.09] & \phantom{-}0.02 [-0.04, \phantom{-}0.07] & \phantom{-}0.07 [\phantom{-}0.03, \phantom{-}0.11]* & \phantom{-}0.14 [\phantom{-}0.09, \phantom{-}0.19]* & -0.09 [-0.12, -0.07]*\\
				cogproc & \phantom{-}0.04 [-0.03, \phantom{-}0.11] & \phantom{-}0.05 [-0.01, \phantom{-}0.10] & \phantom{-}0.12 [\phantom{-}0.08, \phantom{-}0.16]* & \phantom{-}0.12 [\phantom{-}0.08, \phantom{-}0.17]* & -0.02 [-0.04, \phantom{-}0.00]\\
				insight & \phantom{-}0.00 [-0.07, \phantom{-}0.07] & \phantom{-}0.10 [\phantom{-}0.05, \phantom{-}0.16]* & \phantom{-}0.25 [\phantom{-}0.21, \phantom{-}0.29]* & \phantom{-}0.10 [\phantom{-}0.05, \phantom{-}0.15]* & -0.03 [-0.05, -0.00]\\
				\addlinespace
				cause & \phantom{-}0.07 [-0.00, \phantom{-}0.14] & -0.03 [-0.09, \phantom{-}0.02] & -0.06 [-0.10, -0.02] & \phantom{-}0.05 [\phantom{-}0.00, \phantom{-}0.10] & -0.01 [-0.04, \phantom{-}0.01]\\
				memory & -0.09 [-0.16, -0.02] & \phantom{-}0.06 [-0.00, \phantom{-}0.11] & -0.01 [-0.05, \phantom{-}0.03] & \phantom{-}0.03 [-0.02, \phantom{-}0.07] & -0.01 [-0.03, \phantom{-}0.01]\\
				Perception & \phantom{-}0.01 [-0.06, \phantom{-}0.08] & \phantom{-}0.07 [\phantom{-}0.01, \phantom{-}0.12] & -0.01 [-0.05, \phantom{-}0.03] & -0.10 [-0.14, -0.05]* & \phantom{-}0.01 [-0.01, \phantom{-}0.03]\\
				attention & \phantom{-}0.00 [-0.07, \phantom{-}0.07] & \phantom{-}0.03 [-0.03, \phantom{-}0.08] & \phantom{-}0.08 [\phantom{-}0.04, \phantom{-}0.11]* & -0.02 [-0.07, \phantom{-}0.03] & -0.00 [-0.03, \phantom{-}0.02]\\
				motion & -0.01 [-0.08, \phantom{-}0.06] & \phantom{-}0.02 [-0.03, \phantom{-}0.08] & -0.07 [-0.11, -0.04]* & -0.09 [-0.14, -0.04]* & -0.02 [-0.05, \phantom{-}0.00]\\
				\addlinespace
				space & \phantom{-}0.01 [-0.06, \phantom{-}0.08] & \phantom{-}0.07 [\phantom{-}0.01, \phantom{-}0.13] & -0.05 [-0.09, -0.01] & -0.10 [-0.15, -0.05]* & \phantom{-}0.01 [-0.01, \phantom{-}0.04]\\
				visual & \phantom{-}0.01 [-0.06, \phantom{-}0.08] & -0.04 [-0.09, \phantom{-}0.02] & \phantom{-}0.05 [\phantom{-}0.01, \phantom{-}0.09] & \phantom{-}0.06 [\phantom{-}0.01, \phantom{-}0.11] & \phantom{-}0.01 [-0.02, \phantom{-}0.03]\\
				auditory & -0.03 [-0.10, \phantom{-}0.04] & \phantom{-}0.00 [-0.06, \phantom{-}0.06] & \phantom{-}0.02 [-0.02, \phantom{-}0.06] & \phantom{-}0.01 [-0.04, \phantom{-}0.06] & \phantom{-}0.01 [-0.02, \phantom{-}0.03]\\
				feeling & -0.01 [-0.08, \phantom{-}0.06] & \phantom{-}0.02 [-0.04, \phantom{-}0.07] & \phantom{-}0.12 [\phantom{-}0.08, \phantom{-}0.16]* & \phantom{-}0.02 [-0.02, \phantom{-}0.07] & \phantom{-}0.04 [\phantom{-}0.01, \phantom{-}0.06]\\
				time & -0.02 [-0.09, \phantom{-}0.05] & \phantom{-}0.02 [-0.03, \phantom{-}0.08] & -0.02 [-0.06, \phantom{-}0.02] & -0.02 [-0.07, \phantom{-}0.03] & -0.03 [-0.06, -0.01]\\
				\addlinespace
				PROPN & -0.00 [-0.07, \phantom{-}0.06] & -0.01 [-0.07, \phantom{-}0.05] & -0.15 [-0.19, -0.11]* & \phantom{-}0.05 [-0.00, \phantom{-}0.10] & \phantom{-}0.04 [\phantom{-}0.02, \phantom{-}0.07]*\\
				people\_entities & -0.03 [-0.10, \phantom{-}0.03] & \phantom{-}0.01 [-0.05, \phantom{-}0.07] & -0.08 [-0.12, -0.04]* & \phantom{-}0.13 [\phantom{-}0.08, \phantom{-}0.18]* & \phantom{-}0.02 [\phantom{-}0.00, \phantom{-}0.05]\\
				temporal\_entities & -0.02 [-0.08, \phantom{-}0.05] & \phantom{-}0.15 [\phantom{-}0.09, \phantom{-}0.20]* & \phantom{-}0.02 [-0.02, \phantom{-}0.06] & \phantom{-}0.02 [-0.03, \phantom{-}0.07] & \phantom{-}0.01 [-0.01, \phantom{-}0.03]\\
				spatial\_entities & -0.00 [-0.07, \phantom{-}0.07] & \phantom{-}0.09 [\phantom{-}0.03, \phantom{-}0.14] & \phantom{-}0.04 [-0.00, \phantom{-}0.08] & \phantom{-}0.07 [\phantom{-}0.02, \phantom{-}0.12] & \phantom{-}0.02 [-0.01, \phantom{-}0.04]\\
				quantity\_entities & -0.05 [-0.12, \phantom{-}0.02] & \phantom{-}0.08 [\phantom{-}0.03, \phantom{-}0.14] & \phantom{-}0.03 [-0.01, \phantom{-}0.07] & -0.06 [-0.10, -0.01] & -0.01 [-0.03, \phantom{-}0.02]\\
				\addlinespace
				perceptual\_details & \phantom{-}0.00 [-0.07, \phantom{-}0.07] & \phantom{-}0.00 [-0.05, \phantom{-}0.06] & \phantom{-}0.14 [\phantom{-}0.11, \phantom{-}0.18]* & \phantom{-}0.04 [-0.01, \phantom{-}0.09] & \phantom{-}0.02 [-0.00, \phantom{-}0.04]\\
				contextual\_embedding & -0.01 [-0.08, \phantom{-}0.06] & \phantom{-}0.06 [\phantom{-}0.01, \phantom{-}0.12] & -0.07 [-0.11, -0.03]* & -0.11 [-0.15, -0.06]* & -0.02 [-0.04, \phantom{-}0.01]\\
				reality\_monitoring & -0.01 [-0.08, \phantom{-}0.06] & -0.02 [-0.07, \phantom{-}0.04] & -0.08 [-0.12, -0.05]* & -0.13 [-0.17, -0.08]* & \phantom{-}0.04 [\phantom{-}0.02, \phantom{-}0.06]*\\
				discrep & \phantom{-}0.09 [\phantom{-}0.02, \phantom{-}0.16] & -0.06 [-0.12, -0.01] & -0.09 [-0.13, -0.05]* & \phantom{-}0.12 [\phantom{-}0.07, \phantom{-}0.16]* & \phantom{-}0.01 [-0.01, \phantom{-}0.03]\\
				tentat & \phantom{-}0.03 [-0.04, \phantom{-}0.09] & -0.00 [-0.06, \phantom{-}0.06] & \phantom{-}0.06 [\phantom{-}0.02, \phantom{-}0.10] & -0.01 [-0.06, \phantom{-}0.04] & \phantom{-}0.01 [-0.02, \phantom{-}0.03]\\
				\addlinespace
				certitude & -0.06 [-0.13, \phantom{-}0.00] & -0.01 [-0.06, \phantom{-}0.05] & \phantom{-}0.05 [\phantom{-}0.01, \phantom{-}0.09] & \phantom{-}0.10 [\phantom{-}0.05, \phantom{-}0.14]* & -0.04 [-0.06, -0.02]*\\
				lexical\_diversity & \phantom{-}0.09 [\phantom{-}0.02, \phantom{-}0.16] & -0.12 [-0.17, -0.06]* & -0.21 [-0.25, -0.17]* & -0.05 [-0.10, -0.00] & -0.02 [-0.05, \phantom{-}0.00]\\
				difficult\_words & -0.04 [-0.11, \phantom{-}0.03] & \phantom{-}0.12 [\phantom{-}0.06, \phantom{-}0.17]* & \phantom{-}0.25 [\phantom{-}0.21, \phantom{-}0.28]* & \phantom{-}0.06 [\phantom{-}0.01, \phantom{-}0.11] & \phantom{-}0.02 [-0.00, \phantom{-}0.04]\\
				\bottomrule
		\end{tabular}}
		\caption{\label{tab:feature_by_source_en}Point-biserial Correlations by dataset (English)}
	\end{table}
}

\subsection{Prompt Design}\label{sec:prompt-design}
The detailed version of the zeroshot prompt included additional instructions about cues
of deception. For fewshot prompts, the \texttt{<fewshot\_examples>} tag is replaced with two examples each for deceptive and truthful instances, selected based on the sampling logic.
\begin{quote}
	\textbf{System:} \textit{You are an expert deception detection analyst. Your task is to determine whether the author of a text is being intentionally deceptive. A statement is deceptive when the author deliberately expresses something they believe to be false or misrepresents their real thoughts, feelings, intentions, or actions. Focus on the author's intent rather than factual correctness. A person can be honest but mistaken if they sincerely believe what they say, and they can be deceptive while stating something true if they do not genuinely believe it. Use linguistic, emotional, and contextual indicators to infer intent.}
	\vspace{.2cm}
	
	\noindent\textbf{User:} \hspace{.3cm}    \textit{Evaluate whether the author of the provided text is being intentionally deceptive. Consider the author deceptive
		if they knowingly misrepresent their beliefs, emotions, or intentions.}
		
		\noindent\textit{Analyze the following dimensions:}
			
		\noindent\textit{\textbf{Intention over accuracy}: decide whether the author actually believes their statements. } \textit{\textbf{Internal contradictions}: look for inconsistencies within the text or between claims and context.} \textit{\textbf{Strategic framing}: detect signs of manipulation, concealment, persuasion, or image management.}\textit{\textbf{Language indicators}: identify vagueness, distancing, over-justification, rehearsed tone, or unnatural certainty.}
		\textit{\textbf{Emotional consistency}: check if tone and emotional content match the claimed feelings or stakes.}\textit{\textbf{Contextual probability}: judge whether the described situation makes sense if the author is honest.}\textit{\textbf{Ambiguity handling}: being wrong or exaggerating is not lying unless knowingly false or strategically misleading.}
		
		\noindent\textit{If honesty and deception are both plausible, select the more likely interpretation and explain why. Possible scenarios: honest but mistaken about facts means not lying; factually true but not believed means lying; selective omission or framing to mislead means lying; pretending emotions or intentions means lying.}
		
		\noindent\texttt{<fewshot\_examples>}
		
		\noindent\textit{Provide your output strictly in the following JSON format use quotes consistently. Strictly use single quotes for the keys and values in the JSON as shown below:}
		
		\noindent\textit{{\{
				`is\_lying': true/false, `reason\_title': Very brief summary (as a short phrase) of why you believe the author is or isn't lying, `explanation': Detailed explanation supporting your judgment	\}}}
		
		\noindent\textit{Text to evaluate:} \texttt{<input\_instance>}
\end{quote}

For chain-of-thought reasoning based prompts. The section after the \texttt{<fewshot\_examples>} is replace with the following section to encourage reasoning before predicting the label.

\begin{quote}
    \textit{Think step by step: first analyse the text for deception cues and present your reasoning on whether the text is deceptive or not. Provide your output strictly in the following JSON format use quotes consistently. Strictly use single quotes for the keys and values in the JSON as shown below:}

		\noindent\textit{{\{
				'explanation': detailed reasoning including cues, contradictions, and inferred intent, 'reason\_title': 'very brief phrase explaining the main reason'
		`is\_lying': true/false, 	\}}}

\noindent\textit{Text to evaluate:} \texttt{<input\_instance>}
\end{quote}

\subsection{Deception Detection Results}
Table \ref{tab:model_results_de_train} and Table \ref{tab:model_results_en_train} shows the performance of classifiers across all datasets on the training data and Table \ref{tab:model_results_de_train} and Table \ref{tab:model_results_en_holdout} shows validation results on holdout data. We observe that for the comparison datasets \textbf{Bert} model performs consistently well on all three datasets, which \textbf{Mistral-7B} outperforming \textbf{Bert} in \opendomain dataset. Large language models (LLMs) \texttt{Mistral-7b}, \texttt{Llama3-8B}, and \texttt{Qwen-14B} exhibit variable results, often favoring high recall on truthful instances but struggling with deceptive cases.
Table \ref{tab:dataset_composition} shows the label distribution in the data used for training and evaluating the models.

Tables \ref{tab:stage2-results-gpt-4o}, \ref{tab:stage2-results-gpt-4o-mini}, \ref{tab:stage2-results-llama3-70b}, \ref{tab:stage2-results-llama3-8b} and \ref{tab:stage2-results-mistral-7b} show the results for prompt and model size variation
experiments.

\begin{table}
	\centering
	\footnotesize
\begin{tabular}{lrrrrrr}
		\toprule
		\multirow{2}{*}{\textbf{Dataset}} & \multicolumn{3}{c}{\textbf{Train Split}} & \multicolumn{3}{c}{\textbf{Holdout Split}} \\
		\cmidrule(lr){2-4} \cmidrule(lr){5-7}
		& \textbf{Total} & \textbf{Truthful} & \textbf{Deceptive} & \textbf{Total} & \textbf{Truthful} & \textbf{Deceptive} \\
		\midrule
		\defabelone   & 828  & 301 & 527 & 203 & 87 & 116 \\
		\defabeltwo   & 776  & 388 & 388 & 192 & 96 & 96 \\
		\defabeleng       & 654  & 327 & 327 & 150 & 75 & 75 \\
		\crosscult   & 800  & 400 & 400 & 400  & 200 & 200 \\
		\opendomain      & 5734 & 2851 & 2883 & 1434 & 733 & 701 \\
		\intn         & 1312 & 604 & 708 & 328 & 179 & 149 \\
		\decop           & 2000 & 1000 & 1000 & 500 & 250 & 250 \\
		
		\bottomrule
	\end{tabular}
	\caption{Dataset composition across training and holdout splits. Each row represents a dataset, with counts of total, truthful, and deceptive instances per split.}
	\label{tab:dataset_composition}
\end{table}

\begin{table}
	\centering
	\scriptsize
	\setlength{\tabcolsep}{5pt}
	\renewcommand{\arraystretch}{1.1}
	\resizebox{\textwidth}{!}{%
		\begin{tabular}{llllllllll}
			\toprule
			\textbf{Source} & \textbf{Classifier} & \textbf{Accuracy} &
			\multicolumn{3}{c|}{\textbf{Deceptive}} &
			\multicolumn{3}{c|}{\textbf{Truthful}} &
			\textbf{F1 Overall} \\
			\cmidrule(lr){4-6} \cmidrule(lr){7-9}
			& & & \textbf{Prec.} & \textbf{Recall} & \textbf{F1} & \textbf{Prec.} & \textbf{Recall} & \textbf{F1} & \\
			\midrule
			\multirow{11}{*}{DEFABEL\_EN} & MajorityBaseline & 0.50$\,\pm\,0.01$ & 0.50 & 1.00 & 0.67$\,\pm\,0.01$ & 0.00 & 0.00 & 0.00 & 0.33$\,\pm\,0.01$ \\
			& RandomBaseline & 0.49$\,\pm\,0.06$ & 0.49 & 0.51 & 0.50$\,\pm\,0.05$ & 0.49 & 0.47 & 0.48$\,\pm\,0.06$ & 0.49$\,\pm\,0.06$ \\
			& LogisticRegression & 0.51$\,\pm\,0.05$ & 0.51 & 0.51 & 0.51$\,\pm\,0.05$ & 0.51 & 0.51 & 0.51$\,\pm\,0.06$ & 0.51$\,\pm\,0.05$ \\
			& SVC & 0.50$\,\pm\,0.06$ & 0.50 & 0.50 & 0.50$\,\pm\,0.05$ & 0.50 & 0.50 & 0.50$\,\pm\,0.09$ & 0.50$\,\pm\,0.06$ \\
			& RandomForest & 0.52$\,\pm\,0.05$ & 0.52 & 0.53 & 0.52$\,\pm\,0.05$ & 0.52 & 0.50 & 0.51$\,\pm\,0.06$ & 0.52$\,\pm\,0.05$ \\
			& GradientBoosting & 0.51$\,\pm\,0.06$ & 0.51 & 0.51 & 0.51$\,\pm\,0.07$ & 0.51 & 0.51 & 0.51$\,\pm\,0.07$ & 0.51$\,\pm\,0.06$ \\
\bottomrule
		\end{tabular}%
	}
	\caption{Performance across \corpusname datasets and classifiers on \textbf{train} split. Precision, recall, and F1 scores with standard errors are grouped by class label.}
	\label{tab:model_results_de_train}
\end{table}

\begin{table}
	\centering
	\scriptsize
	\setlength{\tabcolsep}{5pt}
	\renewcommand{\arraystretch}{1.1}
	\resizebox{\textwidth}{!}{%
		\begin{tabular}{llllllllll}
			\toprule
			\textbf{Source} & \textbf{Classifier} & \textbf{Accuracy} &
			\multicolumn{3}{c|}{\textbf{Deceptive}} &
			\multicolumn{3}{c|}{\textbf{Truthful}} &
			\textbf{F1 Overall} \\
			\cmidrule(lr){4-6} \cmidrule(lr){7-9}
			& & & \textbf{Prec.} & \textbf{Recall} & \textbf{F1} & \textbf{Prec.} & \textbf{Recall} & \textbf{F1} & \\
			\midrule
			\multirow{11}{*}{CROSS\_CULT\_DE} & MajorityBaseline & 0.50 & 0.50 & 1.00 & 0.67 & 0.00 & 0.00 & 0.00 & 0.33 \\
			& RandomBaseline & 0.50$\,\pm\,0.05$ & 0.50 & 0.48 & 0.49$\,\pm\,0.05$ & 0.50 & 0.53 & 0.51$\,\pm\,0.05$ & 0.50$\,\pm\,0.05$ \\
			& LogisticRegression & 0.65$\,\pm\,0.08$ & 0.64 & 0.66 & 0.65$\,\pm\,0.08$ & 0.65 & 0.64 & 0.64$\,\pm\,0.08$ & 0.65$\,\pm\,0.08$ \\
			& SVC & 0.64$\,\pm\,0.07$ & 0.63 & 0.68 & 0.65$\,\pm\,0.07$ & 0.65 & 0.61 & 0.63$\,\pm\,0.08$ & 0.64$\,\pm\,0.07$ \\
			& RandomForest & 0.66$\,\pm\,0.07$ & 0.65 & 0.69 & 0.67$\,\pm\,0.07$ & 0.67 & 0.62 & 0.64$\,\pm\,0.08$ & 0.66$\,\pm\,0.07$ \\
			& GradientBoosting & 0.66$\,\pm\,0.06$ & 0.66 & 0.68 & 0.67$\,\pm\,0.06$ & 0.67 & 0.65 & 0.65$\,\pm\,0.07$ & 0.66$\,\pm\,0.06$ \\
\bottomrule
		\end{tabular}%
	}
	\caption{Performance across English datasets and classifiers on \textbf{train} split. Precision, recall, and F1 scores with standard errors are grouped by class label.}
	\label{tab:model_results_en_train}
\end{table}

\begin{table}
	\centering
	\footnotesize
	\setlength{\tabcolsep}{5pt}
	\renewcommand{\arraystretch}{1.1}
	\resizebox{\textwidth}{!}{%
		\begin{tabular}{llrccc|ccc|c}
			\toprule
			\textbf{Source} & \textbf{Classifier} & \textbf{Accuracy} &
			\multicolumn{3}{c|}{\textbf{Deceptive}} &
			\multicolumn{3}{c|}{\textbf{Truthful}} &
			\textbf{F1 Overall} \\
			\cmidrule(lr){4-6} \cmidrule(lr){7-9}
			& & & \textbf{Prec.} & \textbf{Recall} & \textbf{F1} & \textbf{Prec.} & \textbf{Recall} & \textbf{F1} & \\
			\midrule
		
\multirow{11}{*}{DEFABEL\_EN} & MajorityBaseline & 0.50 & 0.50 & 1.00 & 0.67 & 0.00 & 0.00 & 0.00 & 0.33 \\
& RandomBaseline & 0.51 & 0.51 & 0.47 & 0.49 & 0.51 & 0.56 & 0.54 & 0.51 \\
& LogisticRegression & 0.50 & 0.50 & 0.64 & 0.56 & 0.50 & 0.36 & 0.42 & 0.49 \\
& SVC & 0.53 & 0.55 & 0.31 & 0.39 & 0.52 & 0.75 & 0.61 & 0.50 \\
& RandomForest & 0.47 & 0.48 & 0.56 & 0.52 & 0.47 & 0.39 & 0.42 & 0.47 \\
& GradientBoosting & 0.51 & 0.51 & 0.55 & 0.53 & 0.51 & 0.48 & 0.50 & 0.51 \\
\bottomrule
		\end{tabular}%
	}
	\caption{Performance across \corpusname datasets and classifiers on \textbf{holdout} split. Precision, recall, and F1 scores with standard errors are grouped by class label.}
	\label{tab:model_results_de_holdout}
\end{table}

\begin{table}
	\centering
	\scriptsize
	\setlength{\tabcolsep}{5pt}
	\renewcommand{\arraystretch}{1.1}
	\resizebox{\textwidth}{!}{%
		\begin{tabular}{llrccc|ccc|c}
			\toprule
			\textbf{Source} & \textbf{Classifier} & \textbf{Accuracy} &
			\multicolumn{3}{c|}{\textbf{Deceptive}} &
			\multicolumn{3}{c|}{\textbf{Truthful}} &
			\textbf{F1 Overall} \\
			\cmidrule(lr){4-6} \cmidrule(lr){7-9}
			& & & \textbf{Prec.} & \textbf{Recall} & \textbf{F1} & \textbf{Prec.} & \textbf{Recall} & \textbf{F1} & \\
			
			\midrule
			\multirow{11}{*}{CROSS\_CULT\_DE} & MajorityBaseline & 0.50 & 0.50 & 1.00 & 0.67 & 0.00 & 0.00 & 0.00 & 0.33 \\
			& RandomBaseline & 0.48 & 0.48 & 0.47 & 0.48 & 0.48 & 0.48 & 0.48 & 0.48 \\
			& LogisticRegression & 0.59 & 0.58 & 0.68 & 0.62 & 0.61 & 0.50 & 0.55 & 0.59 \\
			& SVC & 0.57 & 0.56 & 0.69 & 0.61 & 0.59 & 0.46 & 0.51 & 0.56 \\
			& RandomForest & 0.65 & 0.62 & 0.79 & 0.69 & 0.71 & 0.52 & 0.60 & 0.65 \\
			& GradientBoosting & 0.66 & 0.64 & 0.70 & 0.67 & 0.67 & 0.61 & 0.64 & 0.65 \\
\bottomrule
		\end{tabular}%
	}
	\caption{Performance across English comparison datasets and classifiers on \textbf{holdout} split. Precision, recall, and F1 scores with standard errors are grouped by class label.}
	\label{tab:model_results_en_holdout}
\end{table}

\begin{table}
	\centering
	\scriptsize
	\setlength{\tabcolsep}{5pt}
	\renewcommand{\arraystretch}{.9}
	\resizebox{\textwidth}{!}{%
		\begin{tabular}{llrccc|ccc|c}
			\toprule
			\textbf{Source} & \textbf{Prompt type} & \textbf{Accuracy} & \multicolumn{3}{c|}{\textbf{Deceptive}} & \multicolumn{3}{c|}{\textbf{Truthful}} & \textbf{F1 Overall} \\
			\cmidrule(lr){4-6} \cmidrule(lr){7-9}
			&  &  & \textbf{Prec.} & \textbf{Recall} & \textbf{F1} & \textbf{Prec.} & \textbf{Recall} & \textbf{F1} & \\
			\midrule
			\multirow{9}{*}{CROSS\_CULT\_DE} & zero\_shot\_simple & $.507$ & $\mathbf{1.000}$ & $.015$ & $.030$ & $.504$ & $\mathbf{1.000}$ & $\mathbf{.670}$ & $.350$ \\
			& zero\_shot\_detailed & $.522$ & $.558$ & $.215$ & $.310$ & $.514$ & $.830$ & $.635$ & $.473$ \\
			& fewshot\_random & $\mathbf{.542}$ & $.655$ & $.180$ & $.282$ & $\mathbf{.525}$ & $.905$ & $.664$ & $.473$ \\
			& fewshot\_feature\_similarity & $.515$ & $.537$ & $.220$ & $.312$ & $.509$ & $.810$ & $.625$ & $.469$ \\
			& fewshot\_semantic\_similarity & $.520$ & $.561$ & $.185$ & $.278$ & $.512$ & $.855$ & $.640$ & $.459$ \\
			& zero\_shot\_detailed\_COT & $.435$ & $.432$ & $.415$ & $.423$ & $.438$ & $.455$ & $.446$ & $.435$ \\
\bottomrule
		\end{tabular}
	}
	\caption{Evaluation metrics for \textbf{GPT-4o-mini}: accuracy, class-wise precision/recall/F1, and overall F1. Bold indicates the best value within each source across prompt types.}
	\label{tab:stage2-results-gpt-4o-mini}
\end{table}


\begin{table}
	\centering
	\scriptsize
	\setlength{\tabcolsep}{5pt}
	\renewcommand{\arraystretch}{.9}
	\resizebox{\textwidth}{!}{%
		\begin{tabular}{llrccc|ccc|c}
			\toprule
			\textbf{Source} & \textbf{Prompt type} & \textbf{Accuracy} & \multicolumn{3}{c|}{\textbf{Deceptive}} & \multicolumn{3}{c|}{\textbf{Truthful}} & \textbf{F1 Overall} \\
			\cmidrule(lr){4-6} \cmidrule(lr){7-9}
			&  &  & \textbf{Prec.} & \textbf{Recall} & \textbf{F1} & \textbf{Prec.} & \textbf{Recall} & \textbf{F1} & \\
			\midrule
			\multirow{9}{*}{CROSS\_CULT\_DE} & zero\_shot\_simple & $.490$ & $.250$ & $.010$ & $.019$ & $.495$ & $\mathbf{.970}$ & $\mathbf{.655}$ & $.337$ \\
			& zero\_shot\_detailed & $.468$ & $.397$ & $.125$ & $.190$ & $.481$ & $.810$ & $.603$ & $.397$ \\
			& fewshot\_random & $.512$ & $.514$ & $.465$ & $.488$ & $.511$ & $.560$ & $.535$ & $.511$ \\
			& fewshot\_feature\_similarity & $.468$ & $.397$ & $.125$ & $.190$ & $.481$ & $.810$ & $.603$ & $.397$ \\
			& fewshot\_semantic\_similarity & $\mathbf{.527}$ & $\mathbf{.536}$ & $.410$ & $.465$ & $\mathbf{.522}$ & $.645$ & $.577$ & $\mathbf{.521}$ \\
			& zero\_shot\_detailed\_COT & $.475$ & $.483$ & $.700$ & $.571$ & $.455$ & $.250$ & $.323$ & $.447$ \\
\bottomrule
		\end{tabular}
	}
	\caption{Evaluation metrics for \textbf{LLaMA3-8B}: accuracy, class-wise precision/recall/F1, and overall F1. Bold indicates the best value within each source across prompt types.}
	\label{tab:stage2-results-llama3-8b}
\end{table}


\begin{table}
	\centering
	\scriptsize
	\setlength{\tabcolsep}{5pt}
	\renewcommand{\arraystretch}{.9}
	\resizebox{\textwidth}{!}{%
		\begin{tabular}{llrccc|ccc|c}
			\toprule
			\textbf{Source} & \textbf{Prompt type} & \textbf{Accuracy} & \multicolumn{3}{c|}{\textbf{Deceptive}} & \multicolumn{3}{c|}{\textbf{Truthful}} & \textbf{F1 Overall} \\
			\cmidrule(lr){4-6} \cmidrule(lr){7-9}
			&  &  & \textbf{Prec.} & \textbf{Recall} & \textbf{F1} & \textbf{Prec.} & \textbf{Recall} & \textbf{F1} & \\
			\midrule
			\multirow{9}{*}{CROSS\_CULT\_DE} & zero\_shot\_simple & $.492$ & $.483$ & $.215$ & $.298$ & $.495$ & $.770$ & $.603$ & $.450$ \\
			& zero\_shot\_detailed & $.504$ & $.508$ & $.348$ & $.413$ & $.502$ & $.661$ & $.571$ & $\mathbf{.492}$ \\
			& fewshot\_random & $.519$ & $.530$ & $.256$ & $.345$ & $.515$ & $.777$ & $.620$ & $.482$ \\
			& fewshot\_feature\_similarity & $.504$ & $.508$ & $.348$ & $.413$ & $.502$ & $.661$ & $.571$ & $\mathbf{.492}$ \\
			& fewshot\_semantic\_similarity & $\mathbf{.531}$ & $\mathbf{.549}$ & $.224$ & $.318$ & $.526$ & $\mathbf{.824}$ & $\mathbf{.642}$ & $.480$ \\
			& zero\_shot\_detailed\_COT & $.492$ & $.497$ & $\mathbf{.932}$ & $.648$ & $.409$ & $.048$ & $.085$ & $.367$ \\
\bottomrule
		\end{tabular}
	}
	\caption{Evaluation metrics for \textbf{Mistral-7B}: accuracy, class-wise precision/recall/F1, and overall F1. Bold indicates the best value within each source across prompt types.}
	\label{tab:stage2-results-mistral-7b}
\end{table}


\begin{table}
	\centering
	\scriptsize
	\setlength{\tabcolsep}{5pt}
	\renewcommand{\arraystretch}{1}
	\resizebox{\textwidth}{!}{%
		\begin{tabular}{llrccc|ccc|c}
			\toprule
			\textbf{Source} & \textbf{Prompt type} & \textbf{Accuracy} & \multicolumn{3}{c|}{\textbf{Deceptive}} & \multicolumn{3}{c|}{\textbf{Truthful}} & \textbf{F1 Overall} \\
			\cmidrule(lr){4-6} \cmidrule(lr){7-9}
			&  &  & \textbf{Prec.} & \textbf{Recall} & \textbf{F1} & \textbf{Prec.} & \textbf{Recall} & \textbf{F1} & \\
			\midrule
			\multirow{9}{*}{CROSS\_CULT\_DE} 
			& zero\_shot\_detailed & $.497$ & $.467$ & $.035$ & $.065$ & $.499$ & $.960$ & $.656$ & $.361$ \\
			& fewshot\_random & $.505$ & $.529$ & $\mathbf{.090}$ & $\mathbf{.154}$ & $.503$ & $.920$ & $.650$ & $\mathbf{.402}$ \\
			
			& fewshot\_semantic\_similarity & $\mathbf{.517}$ & $\mathbf{.684}$ & $.065$ & $.119$ & $\mathbf{.509}$ & $\mathbf{.970}$ & $\mathbf{.668}$ & $.393$ \\
			& zero\_shot\_detailed\_COT & $.512$ & $.609$ & $.070$ & $.126$ & $.507$ & $.955$ & $.662$ & $.394$ \\
			& fewshot\_random\_COT & $.505$ & $.538$ & $.070$ & $.124$ & $.503$ & $.940$ & $.655$ & $.389$ \\
			
			& fewshot\_semantic\_similarity\_COT & $.510$ & $.625$ & $.050$ & $.093$ & $.505$ & $\mathbf{.970}$ & $.664$ & $.378$ \\
\bottomrule
		\end{tabular}
	}
	\caption{Evaluation metrics for GPT-4o: accuracy, class-wise precision/recall/F1, and overall F1. Bold indicates the best value within each source across prompt types.}
	\label{tab:stage2-results-gpt-4o}
\end{table}


\begin{table}
	\centering
	\scriptsize
	\setlength{\tabcolsep}{5pt}
	\renewcommand{\arraystretch}{1}
	\resizebox{\textwidth}{!}{%
		\begin{tabular}{llrccc|ccc|c}
			\toprule
			\textbf{Source} & \textbf{Prompt type} & \textbf{Accuracy} & \multicolumn{3}{c|}{\textbf{Deceptive}} & \multicolumn{3}{c|}{\textbf{Truthful}} & \textbf{F1 Overall} \\
			\cmidrule(lr){4-6} \cmidrule(lr){7-9}
			&  &  & \textbf{Prec.} & \textbf{Recall} & \textbf{F1} & \textbf{Prec.} & \textbf{Recall} & \textbf{F1} & \\
			\midrule
			\multirow{9}{*}{CROSS\_CULT\_DE} 
			& zero\_shot\_detailed & $.507$ & $\mathbf{.714}$ & $.025$ & $.048$ & $.504$ & $\mathbf{.990}$ & $\mathbf{.668}$ & $.358$ \\
			& fewshot\_random & $\mathbf{.530}$ & $.630$ & $.145$ & $.236$ & $\mathbf{.517}$ & $.915$ & $.661$ & $\mathbf{.448}$ \\
			
			& fewshot\_semantic\_similarity & $.522$ & $.645$ & $.100$ & $.173$ & $.512$ & $.945$ & $.664$ & $.419$ \\
			& zero\_shot\_detailed\_COT & $.507$ & $.565$ & $.065$ & $.117$ & $.504$ & $.950$ & $.659$ & $.388$ \\
			& fewshot\_random\_COT & $.490$ & $.476$ & $\mathbf{.195}$ & $\mathbf{.277}$ & $.494$ & $.785$ & $.606$ & $.441$ \\
			
			& fewshot\_semantic\_similarity\_COT & $.519$ & $.566$ & $.152$ & $.240$ & $.512$ & $.884$ & $.648$ & $.444$ \\
\bottomrule
		\end{tabular}
	}
	\caption{Evaluation metrics for LLaMA3-70B: accuracy, class-wise precision/recall/F1, and overall F1. Bold indicates the best value within each source across prompt types.}
	\label{tab:stage2-results-llama3-70b}
\end{table}

\subsection{Misclassification Analysis}

Table \ref{tab:fp_fn_by_dataset} presents the false positive (FP) and false negative (FN) counts for each rationale category (cluster) generated from the reasoning provided by the Mistral model, shown across the different datasets.

\begin{table}[htbp]
	\centering
	 \scriptsize
	\setlength{\tabcolsep}{3pt}   
	\renewcommand{\arraystretch}{0.9} 
	\resizebox{\textwidth}{!}{
		\begin{tabular}{ll rrrrrrr}
			\toprule
			& Dataset & \roth{DEFABEL\_v1\_DE} & \roth{DEFABEL\_v2\_DE} & \roth{DEFABEL\_EN} & \roth{DEC\_OP} & \roth{OPEN\_DOMAIN} & \roth{CROSS\_CULT\_DE} & \roth{INTENTION} \\
			\midrule
			Error Type & Reason Cluster Name &  &  &  &  &  &  &  \\
			\cmidrule{1-9}
			\multirow[t]{9}{*}{FN} & Contextual Misrepresentation & 0 & 0 & 3 & 0 & 22 & 0 & 2 \\
			& Contradictions and Logical Inconsistencies & 0 & 2 & 7 & 9 & 236 & 20 & 40 \\
			& Exaggeration or Oversimplification & 0 & 1 & 4 & 12 & 21 & 18 & 0 \\
			& Generalization & 0 & 0 & 0 & 2 & 19 & 0 & 0 \\
			& Lack of Evidence or Basis & 0 & 0 & 0 & 0 & 49 & 0 & 0 \\
			& Misattribution or Attribution Errors & 0 & 0 & 0 & 0 & 11 & 0 & 0 \\
			& Misinformation & 5 & 4 & 14 & 1 & 43 & 0 & 6 \\
			& Misrepresentation of Facts or Science & 9 & 22 & 44 & 56 & 428 & 63 & 6 \\
			& Outdated or Deprecated Information & 0 & 0 & 0 & 0 & 22 & 0 & 0 \\
			\cmidrule{1-9}
			\multirow[t]{13}{*}{FP} & Common Beliefs and Scientific facts & 11 & 10 & 22 & 47 & 30 & 11 & 2 \\
			& Consistency and Coherence & 1 & 7 & 2 & 29 & 2 & 76 & 148 \\
			& Expressing Emotions & 0 & 12 & 5 & 117 & 48 & 74 & 8 \\
			& Factually accurate claims & 13 & 21 & 22 & 3 & 183 & 3 & 3 \\
			& Honest Tone with No Deception & 221 & 128 & 54 & 112 & 556 & 85 & 572 \\
			& Hyperbole or Exaggeration & 9 & 10 & 13 & 41 & 52 & 22 & 4 \\
			& Lack of Context & 1 & 2 & 2 & 0 & 102 & 1 & 0 \\
			& Lack of Contradiction & 5 & 3 & 4 & 6 & 26 & 34 & 11 \\
			& Misunderstanding or Misrepresentation & 87 & 88 & 69 & 6 & 54 & 4 & 10 \\
			& Opinion and Subjectivity & 0 & 2 & 0 & 1 & 80 & 0 & 1 \\
			& Personal Belief and Values & 37 & 46 & 24 & 217 & 23 & 103 & 10 \\
			& Statements or Clarification & 31 & 41 & 30 & 10 & 124 & 8 & 20 \\
			& Truthful Intent & 185 & 81 & 68 & 4 & 0 & 3 & 3 \\
			\bottomrule
	\end{tabular}}
	\caption{Counts by error type (FP/FN), reasoning cluster, and dataset.}
	\label{tab:fp_fn_by_dataset}
\end{table}

\clearpage
\subsection{Cross-corpus analysis}
\begin{figure}
	\centering
	\includegraphics[width=0.8\linewidth]{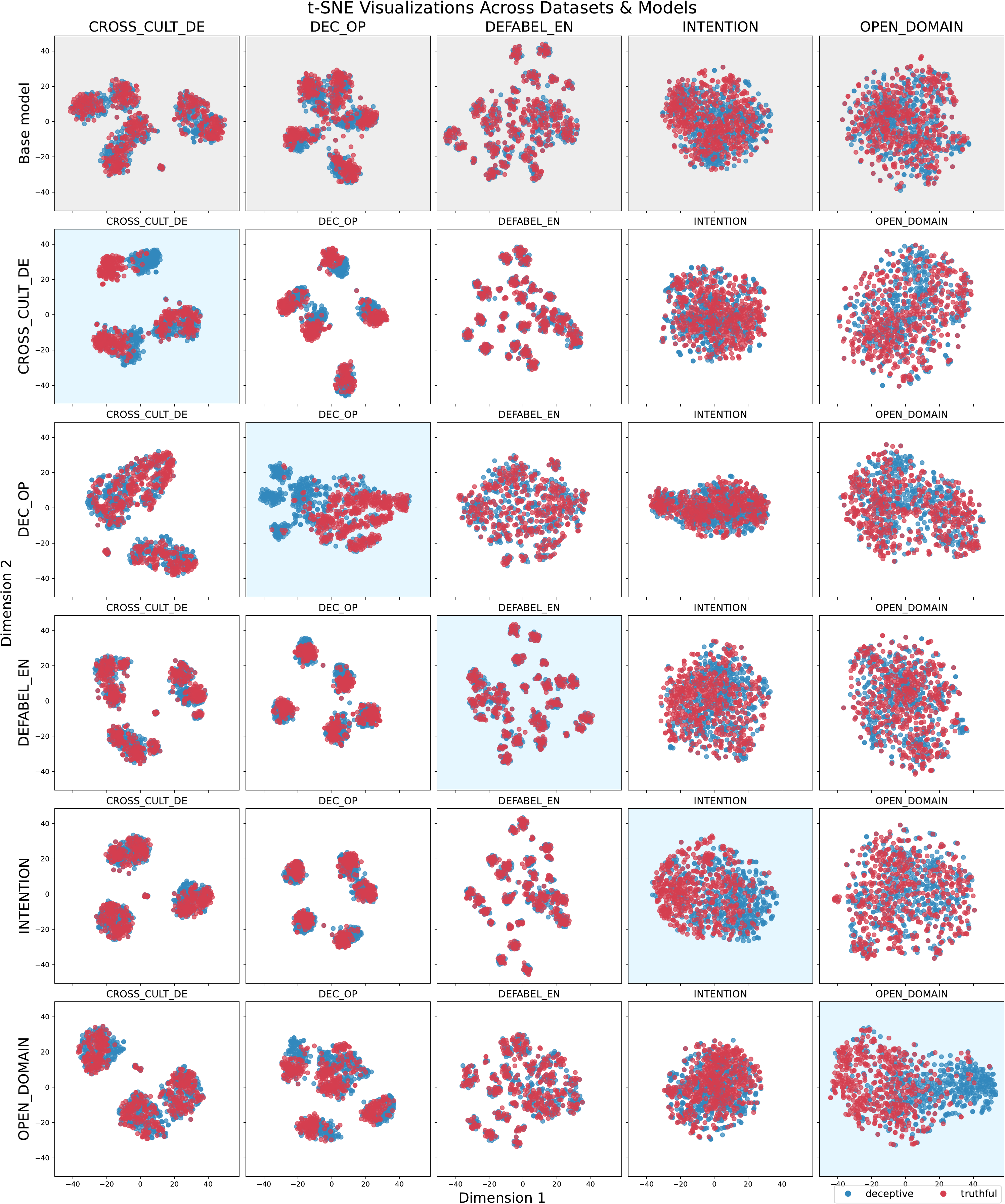}
	\caption{TSNE}
	\label{fig:tsnebiggrid}
\end{figure}

In Figure~\ref{fig:tsnebiggrid}, we visualize a cross-corpus comparison based on bert-based deception classifiers. It shows the representations learned by the base model and multiple fine-tuned RoBERTa models using t-SNE projections of sentence embeddings. Each subplot shows the separation between deceptive and truthful texts across the English datasets, enabling comparison of how well different models capture deception-related distinctions. Rows correspond to models (base and fine-tuned), while columns represent the datasets being evaluated.

We observe that, along the diagonal, where each dataset is evaluated on the model which is fine-tuned with the same dataset, the separation between deceptive and truthful texts is generally more distinct in the case of most datasets except for \defabeleng. This visualized the model's difficulty to discern truthful and deceptive instances in belief-based dataset. 

\end{document}